\documentclass{article}
\usepackage{arxiv_preprint}
\usepackage{cmap}
\usepackage[T1]{fontenc}

\usepackage{amsmath,amsfonts,bm}

\def\eqref#1{equation~\ref{#1}}
\def\Eqref#1{Equation~\ref{#1}}

\def\1{\bm{1}}

\DeclareMathAlphabet{\mathsfit}{\encodingdefault}{\sfdefault}{m}{sl}
\SetMathAlphabet{\mathsfit}{bold}{\encodingdefault}{\sfdefault}{bx}{n}

\usepackage{graphicx}
\usepackage{amssymb,amsthm}
\usepackage{xcolor}
\usepackage[hidelinks]{hyperref}
\usepackage{url}
\usepackage{placeins}
\usepackage{booktabs}
\usepackage{array}

\graphicspath{{figures/}}

\newcommand{\tra}{TRA}
\newcommand{\trap}{TRAP}
\newcommand{\nauc}{\operatorname{nAUC}}
\usepackage{accsupp}
\newcommand{\trapemoji}{%
  \BeginAccSupp{ActualText={},method=escape}%
  \raisebox{-0.18em}{\includegraphics[height=1.05em]{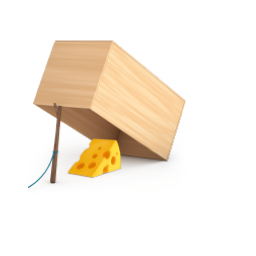}}%
  \EndAccSupp{}}

\theoremstyle{definition}

\theoremstyle{remark}

\theoremstyle{plain}

\newcommand{\essays}{\textit{essays}}
\newcommand{\Essays}{\textit{Essays}}
\newcommand{\apex}{\textit{APEX}}
\newcommand{\medcase}{\textit{MedCase}}
\newcommand{\medex}{\textit{MEDEX}}

\newcommand{\Dtgt}{\mathcal D_{\mathrm{tgt}}}
\newcommand{\Dref}{\mathcal D_{\mathrm{ref}}}
\newcommand{\qref}{q_{\mathrm{ref}}}

\title{TRAP \trapemoji{}: Understanding and Mitigating Privacy Memorization in
Language Models}

\author{%
\mbox{Muhammed Ustaomeroglu\textsuperscript{1,2,}\thanks{Corresponding author:
\href{mailto:mustaome@andrew.cmu.edu}{\texttt{mustaome@andrew.cmu.edu}}.}}%
\And
\mbox{Ziyue Xu\textsuperscript{2}}%
\And
\mbox{Hanshen Xiao\textsuperscript{2,3}}%
\And
\mbox{Peter Cnudde\textsuperscript{2}}%
\And
\mbox{Guannan Qu\textsuperscript{1}}%
\And
\mbox{Holger R. Roth\textsuperscript{2}}%
}
\affiliations{%
\textsuperscript{1}\,Carnegie Mellon University
\quad
\textsuperscript{2}\,NVIDIA
\quad
\textsuperscript{3}\,Purdue University
}

\begin{document}
\maketitle

\begin{abstract}
Fine-tuning a language model on sensitive records can leave it able to reproduce them. We ask when this memorization arises and how to prevent it without knowing in advance which spans are sensitive. Our starting point is that most memorization scores and attacks share one statistical core: whether the model assigns a token more probability than some reference would. Taking as the reference a model trained on the complementary half of the same corpus gives the Target Reference Advantage (TRA), a per-token signal that separates what a model fit to a particular record from what it learned across records, and is cheap and differentiable. We then study what drives memorization during fine-tuning: it keeps growing well past the validation minimum, is larger on small datasets and at higher learning rates, and higher when the underlying task is harder. Early stopping removes much of it, but because it is chosen by aggregate validation loss it helps least for rare, hard-to-predict spans embedded in otherwise learnable text, which is exactly what sensitive information tends to be. We therefore introduce TRAP, a one-sided penalty on tokenwise TRA that acts only where the target model pulls ahead of its reference. On student essays with annotated personal information and clinical cases with patient identifiers, TRAP brings memorization near the level of an untrained model at little utility cost, where generic regularizers barely move and differential privacy gives up most of what fine-tuning bought.
\end{abstract}

\section{Introduction}
\label{sec:intro}

The data that makes a fine-tuned language model valuable is often the data
that cannot leave the building: patient records, hiring files, student
essays, support tickets. Post-training on such data is now routine, and it
carries a specific risk. A model does not only learn how a diagnosis follows
from a case description or how an address is formatted; it can also store the
particular diagnosis of a particular patient, or the phone number of one
applicant, and hand it to whoever asks in the right way. Verbatim training
data, contact details among it, has been extracted from GPT-2 and from a
production chat model with nothing but
prompts~\citep{carlini2021extracting,nasr2023scalable}, fine-tuned models
leak personal information at higher rates
still~\citep{lukas2023analyzing,zeng2024exploring}, and the risk is uneven: a
patient-level audit of medical imaging models found membership inferable
almost perfectly for some patients, disproportionately from under-represented
subgroups, while the aggregate attack barely beat
chance~\citep{knolle2026disparate}. So organizations leave valuable data
unused, and sharing parameters instead of data, as in federated training,
does not help, since the parameters carry the memorized values.

Two questions follow: which training practices cause memorization, so that
the avoidable part can be avoided, and how can the rest be removed without
losing what the fine-tuning bought? Both need a measure of memorization that
is cheap enough to compute for every run and clear about what it measures.
Our starting point is that most memorization scores and attacks ask the same
question: does the trained model assign a sensitive token more probability
than some reference would? The reference decides what counts as
memorization. We take it to be a model trained with the same recipe on the
other half of the data and call the excess log-likelihood the target
reference advantage (\tra{}): what both halves support cancels, what only
the target's records support remains. \tra{} needs two models rather than
the hundreds that counterfactual estimates
require~\citep{feldman2020memorize,zhang2023counterfactual}, it is
differentiable, and in expectation over the split it equals the
generalization gap.

We then use \tra{} and the scores from the literature to map when
memorization of sensitive spans appears during fine-tuning, on student
essays with annotated personal information, synthetic applicant dossiers,
and clinical case reports with a patient identifier. Memorization is the fit
of what records do not share, and it grows with everything that puts
training pressure on that part: training past the validation minimum,
smaller training sets, larger learning rates, tasks whose secrets cannot be
derived from the rest of the record, and rare minorities inside an easy
majority; model size shows no consistent effect once training stops at the
validation minimum. Early stopping removes much of the leak but leaves a
residue the scores still see.
Finally, because \tra{} is differentiable, it can be trained against.
\trap{} (target reference advantage penalty) adds a one-sided penalty on
every token whose advantage over the reference exceeds a threshold, and
needs no annotation of which tokens are sensitive. From 70M to 8B parameters
and across the testbeds, it moves the memorization--utility frontier further
than any baseline we compare against. In addition, Appendix~\ref{sec:related}
reviews the literature on memorization measures, what drives memorization,
and existing defenses.

\paragraph{Contributions.} (i) A reference-comparison view that unifies
memorization scores, and \tra{}, a cheap and differentiable signal whose
expectation is the generalization gap (Section~\ref{sec:method}). (ii) A
controlled study of when sensitive spans are memorized during fine-tuning,
across training duration, data size, learning rate, task hardness, data
mixtures and model size (Section~\ref{sec:when-memorization}). (iii)
\trap{}, a tokenwise penalty on \tra{}, applied during training, that needs
no annotation of which tokens are sensitive (Section~\ref{sec:method}).
(iv) An evaluation on three testbeds and models up to 8B parameters, under
four memorization scores, showing that \trap{} lowers worst-case leakage at
little utility cost and outperforms a broad set of existing defenses
(Section~\ref{sec:mitigation-results}), with a comparison of
reference-model choices in Appendix~\ref{app:reference-schedules}.

\section{Target Reference Advantage and TRAP}
\label{sec:estimator}
\label{sec:method}

\subsection{Target Reference Advantage as a memorization signal}
\label{sec:tra}
\label{sec:sibling-gap}

The memorization measures and attacks reviewed in Appendix~\ref{sec:related}
specify different operations, but most of them ask the same question: does
the target model assign the sensitive tokens more probability than some
reference would? A measure becomes an attack by thresholding or ranking its
score, and an attack defines a measure through what it can distinguish or
recover, so the two families share this core.\footnote{Appendix~\ref{app:metric-forms}
writes each of the individual definitions and attacks as a comparison with an
appropriate reference.} We keep the core and drop the rest. For a tokenized
text $x=(x_1,\ldots,x_T)$, let $\mathcal J_x\subseteq\{1,\ldots,T\}$ be the
positions being evaluated, for instance those of the sensitive tokens, and let
$\qref(\cdot\mid x_{<t})$ be a reference conditional distribution. The average
excess log-likelihood of the target model $p_\theta$ over the reference is
\begin{equation}
\frac{1}{|\mathcal J_x|}
\sum_{t\in\mathcal J_x}
\left[
\log p_\theta(x_t\mid x_{<t})
-
\log \qref(x_t\mid x_{<t})
\right].
\label{eq:general-reference-gap}
\end{equation}

The reference decides which predictive advantage counts as memorization. If
$\qref$ is uniform over the vocabulary, its log-likelihood is a constant and
\Eqref{eq:general-reference-gap} is the target's own average log-likelihood
up to that constant, so any structure learned beyond random prediction
counts. If $\qref$ is a general language model, ordinary linguistic structure
is discounted and task- or domain-specific learning becomes the evidence,
the choice made by membership and extraction work on pretraining data
\citep{carlini2021extracting,mattern2023neighborhood,fu2024selfprompt}.

\textbf{Target Reference Advantage.} We use a reference matched to the task.
A fixed corpus is divided at random into equal complementary halves $\Dtgt$
and $\Dref$, so that each record's sensitive information lies in one half
only, and two models
$\theta$ and $\theta_{\mathrm{ref}}$ are trained independently on them with
the same architecture and learning procedure. Setting
$\qref(\cdot\mid x_{<t})=p_{\theta_{\mathrm{ref}}}(\cdot\mid x_{<t})$ in
\Eqref{eq:general-reference-gap} defines the Target Reference Advantage
(\tra{}) of $\theta$ on a text of $\Dtgt$,
\begin{equation}
\begin{aligned}
\mathrm{TRA}_t(x)
&=
\log p_\theta(x_t\mid x_{<t})
-
\log p_{\theta_{\mathrm{ref}}}(x_t\mid x_{<t}),\\
\mathrm{TRA}(x)
&=
\frac{1}{|\mathcal J_x|}
\sum_{t\in\mathcal J_x}
\mathrm{TRA}_t(x).
\end{aligned}
\label{eq:mia-score}
\end{equation}
A positive \tra{} means that $\theta$ assigns the evaluated tokens more
likelihood than $\theta_{\mathrm{ref}}$. Patterns supported by both halves can
be learned by either model, so they raise both likelihoods and cancel; a
pattern fitted from $\Dtgt$ alone raises the likelihood under $\theta$ with no
matching rise under $\theta_{\mathrm{ref}}$, and remains. \tra{} thus
separates what is specific to the target's records from what generalizes
across the task, and restricting $\mathcal J_x$ to the sensitive positions
measures this advantage where it matters.

The split symmetry also gives \tra{} an aggregate meaning. When
$\mathcal J_x$ covers every token the fine-tuning loss trains on, the
expected \tra{} of a random training example equals the expected
generalization gap, the test loss minus the training loss
(Appendix~\ref{app:tra-gap}). Memorization of a span, measured this way, is
the local form of a quantity every practitioner already tracks.

\subsection{Target Reference Advantage Penalty}
\label{sec:training}

Counterfactual memorization compares a model trained with a sample to one
trained without it, which is more direct than \tra{}, but \tra{} has two
practical advantages. It needs one target and one reference model rather than
many models trained on many subsets of the data, and tokenwise \tra{} is
differentiable in the target's parameters, so it can be a training signal and
not only a post-hoc score.\footnote{Appendix~\ref{app:counterfactual-sibling}
relates the target-and-reference construction to counterfactual memorization
and discusses its limitations.}

The \emph{Target Reference Advantage Penalty} (\trap{}) penalizes a token only
when its \tra{} exceeds a threshold $\epsilon$, that is, when the target
assigns it markedly more probability than a model trained on complementary
data would. The reference model is trained on $\Dref$ and the target on
$\Dtgt$. For a target minibatch $\mathcal B$, let $\mathcal J_x$ contain the
tokens of each $x\in\mathcal B$ that the fine-tuning loss covers, so the
penalty falls on exactly the tokens the cross-entropy trains on, and let
$N=\sum_{x\in\mathcal B}|\mathcal J_x|$. For a weight $w\geq0$ and a
threshold $\epsilon\in\mathbb R$, the target minimizes
\begin{equation}
\resizebox{0.98\linewidth}{!}{$\displaystyle
\mathcal L_{\mathrm{tgt}}
=
\frac{1}{N}
\sum_{x\in\mathcal B}
\sum_{t\in\mathcal J_x}
\left[
-\log p_\theta(x_t\mid x_{<t})
+
w\,\operatorname{ReLU}\!\left(
\underbrace{
\log p_\theta(x_t\mid x_{<t})
-
\log p_{\theta_{\mathrm{ref}}}(x_t\mid x_{<t})
}_{\mathrm{TRA}_t(x)}
-\epsilon
\right)
\right].
$}
\label{eq:loss}
\end{equation}
The first term is the ordinary next-token cross-entropy and the second is
\trap{}. Because the hinge is taken separately at each token, a token with
$\mathrm{TRA}_t(x)\leq\epsilon$ adds nothing and one above the threshold adds
its excess advantage. No annotation of which tokens are sensitive is needed:
the penalty is applied to every trained token, and the reference itself picks
out the ones whose likelihood under the target has risen beyond what the
complementary data can explain.

Two consequences follow directly. First, away from the threshold the
penalty gates the gradient: a token whose advantage is below $\epsilon$
contributes its ordinary cross-entropy gradient, and a token above it
contributes that gradient scaled by $1-w$, attenuated for $0<w<1$, removed
at $w=1$, reversed for $w>1$. \trap{} therefore changes the training signal
only where the target pulls ahead of its reference; whether this buys a
better memorization--utility trade-off is the question of
Section~\ref{sec:mitigation-results}. Second, the tokenwise penalty controls
whole spans: if every token of a span of length $L$ stays below $\epsilon$,
the target assigns that span at most $e^{L\epsilon}$ times the probability
the reference does, so a sensitive value the reference would not produce
cannot become much more likely under the target, and no single token far
above the threshold can hide behind the others.
Appendix~\ref{app:tra-properties} derives both. In addition, once the target
is trained with the penalty and the reference is not, the two are no longer
trained alike, so the identity of Section~\ref{sec:tra} between expected
\tra{} and the generalization gap no longer holds, and during \trap{}
training \tra{} is a training signal only.

\paragraph{The reference.} Our primary variant, the \emph{frozen reference},
trains the reference on $\Dref$ with the ordinary objective, freezes it, and
uses its token probabilities while the target trains on $\Dtgt$ with
\Eqref{eq:loss}; the reference receives no updates. Run alongside the target
it adds one forward pass per step, at most about a third more compute, and
its probabilities on $\Dtgt$ can instead be computed once before training. A
second variant, \emph{iterated rounds}, reuses the best checkpoint of one
\trap{} run as the frozen reference of the next, with the roles of the two
halves swapped. Appendix~\ref{app:reference-schedules} describes both and
compares them with other choices of reference, including the pretrained
base itself. Unless stated otherwise, \trap{} below uses the frozen reference.

\section{When Do LMs Memorize Sensitive Information?}
\label{sec:when-memorization}
This section asks when memorization of sensitive information appears
during fine-tuning and which factors move it, and uses the answer to pick
the training regime that the mitigation study of
Section~\ref{sec:mitigation-results} should target. We first describe the
testbeds briefly; construction details, splits and recipes are in
Appendix~\ref{app:datasets}, and Figure~\ref{fig:testbed-anatomy-wide} there
shows one training example of each, marking which part the loss covers.

\paragraph{\Essays{}.}
We use a public corpus of student essays with naturally occurring, annotated
sensitive spans such as names, email addresses, phone numbers, IDs and
addresses \citep{pii-detection-removal-from-educational-data}. We fine-tune
with the standard next-token objective on each essay independently, as in
continued pretraining on a collection of sensitive documents, so every
token, the sensitive spans included, is a prediction target (see
Figure~\ref{fig:testbed-anatomy-wide} for one sample).

\paragraph{\apex{} (Applicant Extraction).} \apex{} is a fully synthetic
corpus of job-applicant dossiers. Each is a few hundred words of messy hiring
paperwork (resume notes, emails, recruiter chat, an interview transcript)
about one applicant, whose details are spread across its sections, with one
to three other people mentioned in passing as distractors, with their own
names, emails and phone numbers. The prompt holds the dossier, an extraction
instruction and a JSON prefix that already fills in the applicant's name and
application id; the model is trained to complete the remaining fields,
email, phone and address, and the loss covers only this completion (see
Figure~\ref{fig:testbed-anatomy-wide} for one sample). All three appear
verbatim in the dossier, so with the dossier present the task is reading,
not recall. We evaluate memorization closed book: the dossier is dropped and
only the instruction and the prefix remain, so a field can be completed
correctly only from memory.

\paragraph{\medcase{}.} \medcase{} builds on a public corpus of
clinician-authored diagnostic cases distilled from open-access case reports
\citep{wu2025medcase}, to each of which we attach a synthetic patient identifier unique to it (e.g.,
\texttt{ZYRA-4927}). The prompt presents the case and asks for the diagnosis
of the patient with that identifier; the model is trained to produce the
diagnosis, again with the loss on the completion only (see
Figure~\ref{fig:testbed-anatomy-wide} for one sample). The text is real and
the task a genuine clinical skill, but the secret is a short label bound to
a key that occurs in exactly one training example. At evaluation we drop the
case and ask for the diagnosis from the identifier alone, so a correct
closed-book answer is possible only if the model memorized which patient has
which diagnosis, while open-book accuracy still tells whether it learned the
task.

\paragraph{Measuring memorization.} We measure memorization on the sensitive
span itself, an annotated span in an essay, the completed field in \apex{},
or the diagnosis in \medcase{}, not on the whole document. \apex{} and
\medcase{} spans are scored on the closed-book prompt; \essays{} have no
closed-book form, so a span's context starts right after the last earlier
occurrence of the same value in the document. Either way a high score
cannot come from copying the value out of the context. Each testbed is
split into the halves $\Dtgt$ and $\Dref$ of Section~\ref{sec:tra}, one
model is fine-tuned on each with the same recipe, and each is the other's
reference. Members are the spans of the half a
model trained on, non-members the spans of a held-out test split, and every
span is scored exactly once.

Alongside \tra{}, we score each span with three measures that need only the
released model: the probability the model assigns to the exact span, per
token (probability; the LOSS score of the membership-inference literature,
read as a probability), the same quantity standardized token by token
against the model's own predictive distribution (loss++), and the span's
rank among decoy values of the same type (exposure). Like \tra{}, each is a score $S(x)$ computed on
the span positions of $x$; Appendix~\ref{app:metric-forms} gives their
exact forms. Thresholding a score yields a membership test, which we
summarize by its normalized AUC,
\begin{equation}
  \nauc(S)=2\left|\operatorname{AUC}(S)-\tfrac12\right|,
  \qquad
  \operatorname{AUC}(S)=\Pr\bigl(S(X_{\mathrm{train}})>S(X_{\mathrm{test}})\bigr),
  \label{eq:normalized-advantage}
\end{equation}
where $X_{\mathrm{train}}$ is a random member span and $X_{\mathrm{test}}$ a
random non-member span, so $\nauc$ is zero at chance and one under perfect
separation. We normalize because a defended model can over-correct and rank
its own training spans below held-out ones, which an attacker exploits by
flipping the sign; normalizing also puts every score on one scale, so a
model's worst case is its largest $\nauc$. With finitely many spans $\nauc$
is not zero under the null, so every frontier also marks the untrained model
as the empirical floor; chance bands and directional AUCs are in
Appendix~\ref{app:raw-auc-frontiers}.

\subsection{What moves memorization}
\label{sec:factors}

All runs in this subsection are plain fine-tuning with no defense. We change
one factor at a time and report $\nauc$ on the sensitive spans for every
score the runs were evaluated with (a blank panel means that score was not
computed for those runs); the full sweeps are in
Appendix~\ref{app:understanding-sweeps}. The reading that organizes what
follows is that a model first learns what is shared across examples and only
then, if training continues, fits what is specific to each. Memorization is
the second part, and each factor changes either how much example-specific
residue there is or how much training pressure is put on it.

\begin{figure}[t]
  \centering
  \includegraphics[width=\linewidth]{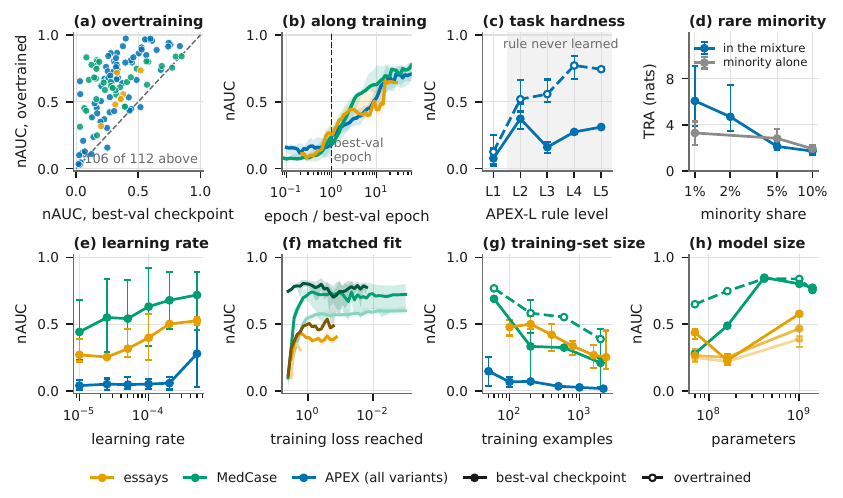}
  \caption{\textbf{What moves memorization.} All panels report $\nauc$ on the
  sensitive spans, averaged over scores, pooling Pythia-70M to 1B
  (\medcase{} to 1.4B) and seeds, except (d), which reports raw \tra{}; the
  less averaged versions are in Appendix~\ref{app:understanding-sweeps}.
  (a) Each configuration trained twice, best-val (horizontal) against
  overtrained (vertical): a point above the diagonal memorized more when
  trained longer. (b) Memorization along training, one median curve per
  testbed, with the epoch axis rescaled so that $1$ is each run's best-val
  epoch. (c) \apex{}-L: against the rule level that generates the secret;
  shading marks levels never learned. (d) A rare minority that can only be
  memorized: \apex{}-FavNum examples, whose answer is a random number absent
  from the text, at $1$ to $10\%$ of a $2000$-example \apex{} training set
  (blue), and the same minority trained alone (grey); \tra{} on the
  minority's secrets in nats per token at best-val, pooling Pythia-70M and
  160M. (e) Against learning rate at best-val. (f) \medcase{} (green) and
  \essays{} (orange) along training at three learning rates each (lighter
  colour is smaller), against the training loss reached, on a log axis from
  about four (left) to a hundredth (right). (g) Against the number of
  training examples. (h) Against model size with the recipe fixed:
  \medcase{} in both regimes, \essays{} at best-val for three learning rates
  (darker colour is larger). In all panels, bars and bands show the spread
  over what was pooled (interquartile range in (b), full range elsewhere).}
  \label{fig:understanding-regime}
\end{figure}

\paragraph{Training duration.}
Every configuration is compared in two regimes: \emph{best-val}, the
checkpoint with the lowest validation loss, which is what early stopping
selects, and \emph{overtrained}, the same configuration trained longer.
Figure~\ref{fig:understanding-regime}a compares the two for every
configuration we trained both ways: the large majority of pairs lie above
the diagonal (per score in Figure~\ref{fig:app-regime-pairs}).
Figure~\ref{fig:understanding-regime}b follows every run we scored along
training (Pythia-70M to 1B~\citep{biderman2023pythia} on each testbed) with
the epoch axis rescaled so that $1$ is each run's best-val epoch
(individual runs in Figure~\ref{fig:app-trajectory-details}). After the
best-val epoch, memorization keeps rising until roughly ten times that
epoch and then plateaus. Stopping at best-val therefore removes much of the
memorization, but not all of it, and the earlier the best-val epoch falls
on this rise, the more it removes. Consequently, we focus on the best-val
checkpoint from here on unless stated otherwise, since it is what a
practitioner who stops on validation loss would deploy; the figures keep
the overtrained arm alongside, and since it memorizes more everywhere it is
also the more demanding case for a defense.

\paragraph{Task hardness.}
\apex{}-L keeps the \apex{} dossiers and prompt, with the applicant's name
and application id prefilled, but the field to produce is the applicant's
portal number, computed from six visible ingredients of the dossier by a
rule the text never states (see Figure~\ref{fig:testbed-anatomy-wide} for
one sample). The rule comes in five levels, L1 to L5, from a small lookup
table to a six-term modular hash; the dossiers are identical across levels,
so the levels differ only in how hard the rule is to learn
(Appendix~\ref{app:datasets}, Table~\ref{tab:apex-ladder}).
Figure~\ref{fig:understanding-regime}c plots memorization against the
level, and Figure~\ref{fig:app-hardness-detail}b whether the rule was
learned, as open-book accuracy on unseen applicants. At the easy level the
model learns the rule within a few epochs, so the portal numbers are derived
rather than stored and the best-val checkpoint memorizes almost nothing. At
the harder levels the model cannot find the pattern, and the only way left
to lower the loss is to bind each portal number to the name and id in the
prompt; that binding is memorization, and every unlearnable level is
memorized more than the learnable one already at best-val. Among the
unlearnable levels the harder the rule, the higher memorization climbs once
training continues past best-val (the dashed curve of
Figure~\ref{fig:understanding-regime}c), whereas at best-val they are not
ordered, since validation loss bottoms out within a few epochs whatever the
rule, before most of the binding has formed. Appendix~\ref{app:understanding-hardness}
gives per-score curves, best-val epochs and closed-book reproduction of the
portal numbers.

\paragraph{Rare, hard subgroups.}
Real training sets mix easy patterns shared by many examples with hard ones
shared by few. We push this to the extreme with a mixture: most of the
training set is ordinary \apex{} examples, and a small share is
\apex{}-FavNum examples, the same dossiers but with the question asking for
the applicant's favorite number, a random four-digit number that appears
nowhere in the text, so these examples can only be memorized. For comparison
we also train on the minority alone, the same number of examples without any
majority.

Figure~\ref{fig:understanding-regime}d shows how strongly the minority is
memorized at best-val as its share varies, in raw \tra{}, the average
log-probability gap between the target and its reference on the
secret.\footnote{Here it is not $\nauc$ because with so few secrets it is
close to one in every run and can no longer separate the cases.} The rarer
the minority, the more it is memorized, and at the rarest share it is
memorized far more inside the mixture than when trained alone, whereas at
$5$ to $10\%$ the two are alike. Figure~\ref{fig:app-mixture-detail}b
explains why. Trained alone, the minority reaches its best-val checkpoint
after one epoch, since there is nothing to generalize. Inside the mixture
the majority keeps validation loss improving for several more epochs, each
another pass over the minority examples, which the model can only store,
and the rarer the minority the longer this goes on.
Figure~\ref{fig:app-mixture-detail}c repeats the experiment with an
\apex{}-L secret at $1\%$ share in place of an impossible one: the
learnable level stays near chance and the harder levels are memorized. In
every case the validation loss that decides when to stop is dominated by
the majority, so the examples it protects least are the rare, hard ones.

This situation is common, and the \essays{} are an instance of it. An email
address or a phone number cannot be predicted from the surrounding text, so
each sensitive span is a favorite number inside an otherwise learnable
essay, and the best-val epoch is set by the essay text, not by the spans.
Figure~\ref{fig:understanding-regime}b shows the result: on \apex{}, where
the whole target is the secret, memorization stays flat until the best-val
epoch and rises only afterwards, whereas on the \essays{} it rises
throughout.
\paragraph{Learning rate.}
Figure~\ref{fig:understanding-regime}e varies only the learning rate, with
every run stopped at its own best-val epoch. Memorization increases with
the rate on every testbed, and almost every sweep series behind the curve
rises too (Appendix~\ref{app:understanding-lr}, also one score at a time).
This is not just more training in disguise: compared at the same training
loss rather than the same epoch, the higher rate still memorizes more at
every level of fit on both \medcase{} and the \essays{}
(Figure~\ref{fig:understanding-regime}f), most clearly under the probability
score and by a smaller margin under \tra{} (Figure~\ref{fig:app-lr-fit}).
Our reading is that a larger step imprints more of each example per
presentation, while the shared structure that sets the best-val epoch is
learned either way. Large learning rates are often credited with an implicit
regularization that improves
generalization~\citep{liwei2019regularization,smith2021origin}; whatever it
does for the shared structure, it does not extend to the example-specific
part of the fit.

\paragraph{Training-set and model size.}
The fewer examples a model is trained on, the more of each it memorizes, on
every testbed, across more than an order of magnitude of size and in both
regimes (Figure~\ref{fig:understanding-regime}g;
Appendix~\ref{app:understanding-size}): each example receives a larger share
of the updates and there is less shared structure to learn instead, and
early stopping helps most at the smallest sizes. Model size is less clear.
When overtrained, larger models memorize
more~\citep{carlini2023quantifying,tirumala2022memorization,biderman2023emergent},
and \medcase{} shows this in both regimes
(Figure~\ref{fig:understanding-regime}h). At the best-val checkpoint,
however, we find no consistent pattern: on the \essays{} the ordering is not
monotone and depends on the learning rate, with the largest model memorizing
most when all sizes share a rate and least once its rate is lowered, and on
\apex{}-L the sizes are close (Appendix~\ref{app:understanding-scale}). What
survives is a caution: a claim about scale has to say which learning rate
and which stopping rule it was made under.

\paragraph{Lessons.} What a fine-tuned model memorizes is the part of each
example that the other examples do not explain, and it forms after the
shared structure has been learned. For a practitioner this gives a short
list of practices, each of which helped in our runs and none was enough. Stop on validation loss and do not train past it: the best-val
checkpoint removes most of what longer training would add. Use the smallest
learning rate that reaches the task performance needed, since at the same
training loss a larger rate leaves more of each example behind. Pool as much
data as possible, because each example's share of the updates is what
matters, and do not assume a larger model is safer: at the best-val
checkpoint we found no consistent effect of size, and where larger models
looked safer they had been trained at a lower learning rate. Above all, do not read a healthy validation curve as
evidence of privacy. Validation loss is an aggregate, set by the majority of
the data, and knows nothing about a particular record, so early stopping
protects least exactly the spans that matter most: rare, hard-to-predict
values inside otherwise learnable text, of which a hard minority inside an
easy majority is the extreme case (Figure~\ref{fig:understanding-regime}d)
and sensitive spans in ordinary documents the common one. Memorization
should therefore be measured on those spans directly, against held-out
spans, with more than one score, and with an untrained model as the floor.
All of these knobs act on every token alike and are bounded by the task and
the model one wants, and none brought the sensitive spans to chance in our
runs. What is missing is a per-token signal that separates the two kinds of
fitting. \tra{} aims to be that signal: shared structure, including a
rule the model has learned, should cancel and only example-specific fitting
remain, and on the mixture it is large on the minority's secrets. \trap{} penalizes that part while the standard objective
keeps learning the rest. The next section tests whether this improves the
trade-off between privacy and utility rather than moving along it, comparing
defenses at the best-val checkpoint, since that is what would be deployed,
and a defense that only removes what overtraining added has done nothing
early stopping would not.

\section{Mitigating Privacy Memorization}
\label{sec:mitigation-results}

A practitioner has a fixed fine-tuning set with sensitive spans in it and
wants the model that learns the task best while memorizing those spans
least.\footnote{We leave compute out of the comparison: all methods here cost
comparable total compute, and in this setting the memorization-utility
trade-off is what matters, not the training bill.} This section asks whether
\trap{} gives a better trade-off than the ordinary knobs and the existing
defenses, across datasets and model sizes.

\begin{figure}[t]
  \centering
  \includegraphics[width=\linewidth]{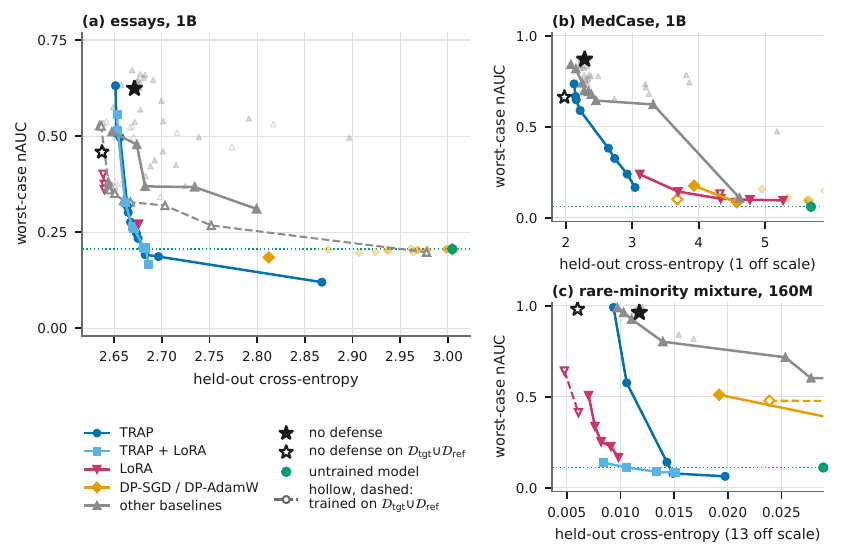}
  \caption{\textbf{Memorization against utility.} Held-out cross-entropy
  (horizontal) against the worst-case $\nauc$ over the scores each run was
  evaluated with (vertical); lower is better on both. Each line is a method
  family's lower-left frontier over its hyperparameter values (seed means
  where seeds exist): \trap{} pools its frozen and iterated references,
  \trap{} + LoRA is the same penalty through an adapter, DP the two
  optimizers, and the grey line pools the remaining baselines into one
  frontier (drawn one by one in Figures~\ref{fig:app-frontier-essays-1b},
  \ref{fig:app-frontier-medcase-1b} and~\ref{fig:app-frontier-mixture-160m});
  for the baselines and DP the remaining hyperparameter values are the faint
  dots. Hollow markers with dashed lines are the same families trained on
  $\Dtgt\cup\Dref$; the star is the undefended model, the hollow star its
  union-trained twin, the green dot and dotted line the untrained model.
  Points beyond $2.5\times$ the undefended cross-entropy are left out and
  counted in the axis label; the untrained model is drawn at the right edge
  when it lies beyond. (a) \Essays{} at Pythia-1B. (b) \medcase{} at
  Pythia-1B. (c) The rare-minority mixture of Section~\ref{sec:factors} at
  Pythia-160M.}
  \label{fig:mitigation}
\end{figure}

\paragraph{Setup.}
Every defended model is trained on $\Dtgt$ with the recipe of its undefended
counterpart and stopped at its own best-val epoch. Utility is the held-out
cross-entropy on the tokens the training loss covers: every token of an
essay, and the completion alone for \apex{} and \medcase{}, that is, the
diagnosis. Privacy is the worst case among the
four memorization scores of Section~\ref{sec:when-memorization}
(probability, loss++, exposure and \tra{}), that is, the largest $\nauc$ any
of them reaches on the sensitive spans. Of the four, \tra{} needs the
reference model, which an attacker with only the released model lacks, and
it is also what \trap{} trains against;
Appendix~\ref{app:mitigation-frontiers} therefore reports the frontiers one
score at a time, and \trap{}'s advantage holds under each. The
hyperparameters shared by all methods, such as batch size, learning rate and
the early-stopping rule, are held at the testbed's recipe (DP under its own
optimizer's); each method is then swept over its own hyperparameters (the
penalty weight for \trap{}, the rank for LoRA, the privacy budget for DP,
and so on), from a value with little effect up to the strongest value in its
published range, and drawn as its lower-left frontier through seed means
where seeds exist, so the comparison is between curves rather than chosen
points and a larger grid cannot buy an advantage from seed noise;
Appendix~\ref{app:frontier-settings} lists every frontier setting, and
\trap{}'s threshold $\epsilon$ is zero unless stated.
Figure~\ref{fig:mitigation} shows the \essays{} and \medcase{} at Pythia-1B
and the rare-minority mixture of Section~\ref{sec:factors}, \apex{}-FavNum
secrets at a $1\%$ share inside ordinary \apex{}, at Pythia-160M.
Appendix~\ref{app:mitigation-frontiers} repeats the comparison on the
\essays{} and the mixture at Pythia-70M and on \medcase{} at Llama-3.1-8B
and in the overtrained regime, and gives the number of seeds behind every
campaign.

\paragraph{Baselines.}
We compare against what a practitioner might reach for: LoRA
\citep{hu2022lora}, DP-SGD and DP-AdamW \citep{abadi2016deep,li2022large},
and a set of objectives and regularizers, Goldfish loss
\citep{hans2024goldfish}, RelaxLoss \citep{chen2022relaxloss}, flooding
\citep{ishida2020flooding}, the confidence penalty
\citep{pereyra2017regularizing}, loss truncation \citep{kang2020improved},
NEFTune \citep{jain2024neftune}, dropout, weight decay and gradient
clipping, plus two of our own design, a KL anchor toward the pretrained base or toward the
reference model, the latter the training-time form of bounding divergence
from a reference trained without the sensitive
content~\citep{vyas2023provable}, and an $n$-gram regularizer. At 70M we also
ran FedAvg \citep{mcmahan2017communication} and SCAFFOLD
\citep{karimireddy2020scaffold} with $\Dtgt$ and $\Dref$ as the two
clients. None of
the methods, \trap{} included, reads the annotations of which tokens are
sensitive, and DP points are what the scores measure on those models, not a
certified guarantee. \trap{} is run with its frozen and iterated references,
on the full model and through a LoRA adapter (\trap{} + LoRA);
Appendix~\ref{app:mitigation-frontiers} shows every method on its own.

\paragraph{Results.}
The picture in Figure~\ref{fig:mitigation} is the same in every panel. The
undefended model sits in the upper left, and almost every objective and
regularizer sits next to it whatever its hyperparameter: tools built against
overfitting change utility more than memorization. DP, and on \medcase{}
also LoRA, reach the floor only after giving up most of what fine-tuning had
bought. \trap{}'s frontier is the one that descends: near the undefended
utility it removes most of the memorization, and a larger weight trades
utility for the rest, down to the floor. LoRA competes on the \essays{},
where restricting capacity already limits what can be stored, and there
\trap{} + LoRA is the best of all; on \medcase{} and the mixture \trap{}
leads. The mixture panel is the uneven case of Section~\ref{sec:factors}, a
rare, hard minority inside an easy majority~\citep{knolle2026disparate}: the
undefended model memorizes the twenty secrets almost perfectly while the
aggregate loss looks healthy, \trap{} brings their worst case close to the
floor at a utility cost of a few thousandths of a nat, and \trap{} + LoRA
does so below the undefended cross-entropy, with no label to find the
minority, since the penalty is largest where the target pulls furthest ahead
of its reference. LoRA alone stays above the \trap{} + LoRA frontier at
every rank. DP enters the panel only at its loosest budget, where it leaks
as much as the widest adapter; at the budgets where it does approach the
floor it costs several times the undefended cross-entropy. Doubling the
data by training on $\Dtgt\cup\Dref$ lowers memorization, as
Section~\ref{sec:factors} predicts, but neither it nor the union-trained
baselines reach \trap{}'s frontier.

This is what Section~\ref{sec:factors} predicts, and it comes down to how
each method decides which part of the training signal to alter. DP-SGD
clips every example's gradient to the same norm and adds the same noise, so
it restrains most the examples with large gradients; gradient clipping and
loss truncation act on the size of the gradient or the loss, the confidence
penalty and RelaxLoss on how peaked the prediction is, Goldfish loss on a
pseudorandom mask, LoRA on how much can be stored at all. Each of these is
at most correlated with memorization: a token can have a large gradient, a
high loss or a confident prediction because it is hard, or because the
pattern behind it is real and shared by the rest of the data, and none of
these criteria can tell the two apart. \trap{}'s criterion is the advantage
over a reference trained on other data, which is meant to be zero for
anything that data supports, so the penalty falls on the example-specific
part of the fit and on little else. Consequently, the knobs move
memorization only as far as they move utility while \trap{} moves the
frontier, and the adapter composes with it, since one limits how much can
be stored and the other what is stored. With twenty member spans, on
\medcase{} and the mixture, differences among defended models near the
floor are directions rather than rankings.

\FloatBarrier

\section{Conclusion}

We studied when a fine-tuned language model memorizes the sensitive spans in
its training data and how to stop it. Most memorization scores compare the
model's likelihood with a reference, and a reference trained on the other
half of the data gives \tra{}, a score that is cheap, differentiable, and
equal in expectation to the generalization gap. Measured this way,
memorization is the fit of what the training examples do not share: it grows
with training past the best-val epoch, smaller training sets, larger
learning rates, harder tasks and rare subgroups, with no consistent effect of
model size at the best-val checkpoint, and early stopping removes much of it
but not all. \trap{} penalizes \tra{} token by
token, needs no annotation of which tokens are sensitive, and gives a better
memorization--utility trade-off than differential privacy, adapters and a
dozen regularizers on every testbed we ran. It is an empirical defense
rather than a formal guarantee, and our scores describe an attacker who does
not adapt to it; one who calibrates against shadow models of his
own~\citep{carlini2022lira} is untested. The shipped model has also learned
from only half the data and still beats the baselines trained on all of it;
a pair of models each defended against the other would leave no data
unused. Finally, a smaller learning rate memorizes less at the same task
performance, and we could not say why.

\bibliography{references}
\bibliographystyle{preprint}

\appendix

\section{Background and Related Work}
\label{sec:related}
\paragraph{Understanding and measuring memorization.}
Membership inference asks whether a text was in the training
set~\citep{shokri2017membership,yeom2018privacy,song2021systematic,carlini2022lira},
with scores that threshold the loss, average the least likely
tokens~\citep{shi2024detecting,zhang2024minkpp}, or calibrate against
another model or nearby texts~\citep{mattern2023neighborhood,fu2024selfprompt};
exposure ranks a canary among decoys~\citep{carlini2019secret}; extraction
measures what prompting
recovers~\citep{carlini2021extracting,carlini2023quantifying,lukas2023analyzing};
and counterfactual memorization compares models trained with and without an
example~\citep{feldman2020memorize,feldman2020longtail,zhang2023counterfactual},
the cleanest definition but one that needs many models, which cheaper
reference-subtracting proxies
approximate~\citep{tiwari2026prior,morris2025memorize}. Whether membership
scores work at all on large models is contested~\citep{duan2024membership}.
Section~\ref{sec:method} writes these measures as one comparison against a
reference, and our experiments report four of them side by side. As for
what drives memorization, it grows with model size, duplication and context
length~\citep{carlini2023quantifying}, appears early along training and
without overfitting~\citep{tirumala2022memorization}, is only partly
predictable from smaller models or earlier
checkpoints~\citep{biderman2023emergent}, fills a capacity of a few
bits per parameter~\citep{morris2025memorize}, and is reduced by
deduplication~\citep{kandpal2022deduplicating,lee2022deduplicating};
fine-tuning on task or synthetic data shows the
same~\citep{zeng2024exploring,ruan2025overmemorization,akkus2025generated}.
Section~\ref{sec:when-memorization} revisits these factors for sensitive
spans in fine-tuning data under matched recipes.

\paragraph{Mitigation.}
Differentially private training gives a formal guarantee at a utility
cost~\citep{dwork2014algorithmic,abadi2016deep,li2022large}. Most other
defenses reuse tools built against overfitting, such as dropout, weight
decay, gradient clipping,
flooding~\citep{ishida2020flooding}, the confidence
penalty~\citep{pereyra2017regularizing}, loss
truncation~\citep{kang2020improved}, noisy embeddings~\citep{jain2024neftune}
and RelaxLoss~\citep{chen2022relaxloss}; Goldfish loss drops a pseudorandom
subset of tokens from the loss~\citep{hans2024goldfish}; parameter-efficient
fine-tuning limits how much can be stored~\citep{hu2022lora}; federated
training spreads the data over
clients~\citep{mcmahan2017communication,karimireddy2020scaffold}; and
membership-inference defenses regularize against an attacker or distil
through an
ensemble~\citep{nasr2018adversarial,chourasia2022knowledge,tang2022selena}.
Closest to \trap{} is the idea of bounding a model's divergence from a
reference trained without the sensitive content~\citep{vyas2023provable};
\trap{} instead acts during training and token by token, penalizing only
the tokens whose likelihood under the target exceeds the reference by more
than a threshold, with no annotation of which tokens are sensitive.
Section~\ref{sec:mitigation-results} compares it with the methods above,
except the membership-inference defenses, and with two of our own design,
a KL anchor toward the base or the reference model and an $n$-gram
regularizer.

\section{A Common Likelihood-and-Reference View of Memorization}
\label{app:metric-forms}

Section~\ref{sec:tra} made two claims about the memorization measures and
attacks reviewed in Appendix~\ref{sec:related}. First, they share one core:
each compares the probability the model assigns to the actual tokens of a
text with a baseline for how probable those tokens should be, and reads a
probability that is too high relative to the baseline as memorization.
Second, measures and attacks are two sides of the same thing. Thresholding,
ranking, or searching with a measure gives an attack, and an attack's success
on a candidate is itself a measure. This appendix backs up both claims by
writing each measure out, pointing at its baseline, which we call the
reference, and at the rule that turns it into an attack. The measures differ
mainly in the reference. Going down the list, it is a fixed threshold, the
model's own prediction at each position, a set of decoy values, perturbed
copies of the text, a compressor, a general-purpose language model, models
trained without the text, and, for \tra{}, a model trained on the other half
of the same data.

\paragraph{Whole documents versus sensitive spans.}
The papers that introduced these measures apply them to a whole document or
to a synthetic canary, so their sums run over every token
\citep{yeom2018privacy,carlini2019secret,carlini2021extracting,shi2024detecting,zhang2024minkpp}.
We ask a more focused question, whether the model has memorized one sensitive
span (a name, an identifier, a diagnosis) inside a document. So throughout
this appendix every score is written over the position set $\mathcal J_x$ of
Section~\ref{sec:tra}, which in our experiments contains exactly the tokens of
the sensitive span. Setting $\mathcal J_x=\{1,\ldots,T\}$ gives back the
original whole-document versions.

\paragraph{Conventions.}
$\ell(\theta,x)$ is the mean negative log-likelihood over $\mathcal J_x$, as in
\Eqref{eq:sample-loss-definition}. All scores are oriented so that a larger
value means stronger evidence of memorization. One caveat before starting:
sharing the same core does not make these measures interchangeable. A
membership attack decides about a candidate it is handed, exposure ranks the
true value among decoys, counterfactual memorization measures the effect of
including a sample, and extraction has to actually recover the text under a
given prompt, decoder, and budget. A high score under one of them is evidence
of memorization in that sense, not automatically in the others.

\subsection{LOSS, perplexity, and the probability score}
\label{app:score-loss}
The LOSS score \citep{yeom2018privacy} is just the mean log-likelihood of the
evaluated tokens,
\begin{equation}
S_{\mathrm{LOSS}}(\theta;x)=\frac{1}{|\mathcal J_x|}\sum_{t\in\mathcal J_x}
\log p_\theta(x_t\mid x_{<t})
=-\ell(\theta,x).
\label{eq:app-loss}
\end{equation}
The main text reports it as a probability,
\begin{equation}
S_{\mathrm{prob}}(\theta;x)
=\exp S_{\mathrm{LOSS}}(\theta;x)
=\Bigl(\prod_{t\in\mathcal J_x}p_\theta(x_t\mid x_{<t})\Bigr)^{1/|\mathcal J_x|},
\label{eq:app-prob}
\end{equation}
the per-token probability of the exact span. LOSS, perplexity
$e^{\ell(\theta,x)}$, and this probability are monotone in one another, so they
rank spans identically and give the same AUC.

To attack with LOSS, one picks a threshold $\tau$, using known non-members or
public data, and calls $x$ a member whenever $S_{\mathrm{LOSS}}(\theta;x)>\tau$.
So the comparison is the probability of the span tokens against one fixed
number, the same for every candidate; in \Eqref{eq:general-reference-gap}
this is $\qref=\mathrm{const}$. This is also the weakness of LOSS: text that
is easy to predict for everyone looks the same as text that is easy because
it was trained on \citep{yeom2018privacy,mattern2023neighborhood}.

\subsection{Min-K\%, Min-K\%++, and loss++}
Min-K\% \citep{shi2024detecting} is LOSS averaged over only the least likely
tokens. With $k=\lceil K|\mathcal J_x|/100\rceil$ and $I_K(x)$ the $k$
positions in $\mathcal J_x$ with the smallest $\log p_\theta(x_t\mid x_{<t})$,
\begin{equation}
S_{\mathrm{MinK}}(\theta;x)=\frac{1}{k}\sum_{t\in I_K(x)}
\log p_\theta(x_t\mid x_{<t}).
\label{eq:app-mink}
\end{equation}
The idea is that unseen text tends to contain a few very unlikely tokens,
while memorized text does not. The reference is again a threshold; only the
positions that enter the average change.

Min-K\%++ \citep{zhang2024minkpp} compares each token with the model's own
prediction at that position instead. Write $p_t(v)=p_\theta(v\mid x_{<t})$
and let
\begin{align}
\mu_t&=\sum_{v}p_t(v)\log p_t(v),\qquad
\sigma_t^2=\sum_v p_t(v)\bigl(\log p_t(v)-\mu_t\bigr)^2,\nonumber\\
z_t&=\frac{\log p_\theta(x_t\mid x_{<t})-\mu_t}{\sigma_t}.
\label{eq:app-minkpp-token}
\end{align}
Then $S_{\mathrm{MinK}^{++}}$ is the mean of the $k$ smallest $z_t$. Here
$\mu_t$ is the log-probability the model expects to assign to its next token
(the negative entropy of $p_t$). Subtracting it asks how much more likely the
observed token was than a typical draw from the model at that position, and
dividing by $\sigma_t$ puts sharp and flat positions on one scale. So the
comparison is still the probability of the observed token against a
baseline, but the baseline is now computed per position from the model
itself rather than fixed once for all candidates. The attack is again a
threshold on the score.

On a span of a few tokens the lowest $K\%$ keeps one or two tokens and
mostly adds noise, so we set $K=100\%$, under which Min-K\% becomes LOSS
and Min-K\%++ becomes the mean of $z_t$ over the span (the \apex{}-full grid
of Appendix~\ref{app:understanding-hardness} was scored at $K=20\%$, as its
captions say). That mean is what we report and call loss++,
\begin{equation}
S_{\mathrm{loss{+}{+}}}(\theta;x)
=\frac{1}{|\mathcal J_x|}\sum_{t\in\mathcal J_x}
\frac{\log p_\theta(x_t\mid x_{<t})-\mu_t}{\sigma_t}.
\label{eq:app-losspp}
\end{equation}
It keeps the part of Min-K\%++ that matters on short spans, the per-position
standardization, and drops the selection step that does nothing there.

The standardized score has a ceiling. As the model approaches certainty on
the observed token, the numerator $\log p_\theta(x_t\mid x_{<t})-\mu_t$ and
the scale $\sigma_t$ vanish together and $z_t$ tends to
$\sqrt{(1-q_t)/q_t}$, where $q_t$ is the probability of the observed token,
almost regardless of the rest of the distribution. A fully memorized span is
therefore mapped close to zero, the same place as a span the model merely
finds easy, and among confident positions the score falls as confidence
rises. Where the model is instead certain about a different token,
$\sigma_t$ vanishes while the numerator does not and $z_t$ diverges; the
implementation clamps the variance at $10^{-12}$, so such values are large
but finite, and because every score is read as an AUC they change ranks
rather than magnitudes. We report loss++ for comparability with prior work,
not as the score to rely on near the floor.

\subsection{Exposure}
Exposure \citep{carlini2019secret} compares the true value with a set of
decoys. Let $\mathcal R_x$ be a set of candidate texts that are identical to
$x$ outside the span and differ only in the span, with $x$ itself among them.
The rank of the true span is one plus the number of decoys that score at least
as high,
\begin{equation}
\operatorname{rank}_\theta(x;\mathcal R_x)
=1+\sum_{x'\in\mathcal R_x\setminus\{x\}}
\mathbf 1\!\left\{
S_{\mathrm{LOSS}}(\theta;x')\geq S_{\mathrm{LOSS}}(\theta;x)
\right\},
\label{eq:app-rank}
\end{equation}
and exposure is
\begin{equation}
S_{\mathrm{Exp}}(\theta;x)
=\log_2|\mathcal R_x|
-\log_2\operatorname{rank}_\theta(x;\mathcal R_x).
\label{eq:app-exposure}
\end{equation}
It is zero when the true value ranks last and $\log_2|\mathcal R_x|$ when it
ranks first. The comparison is the probability of the true span tokens
against the probabilities of the same positions filled with other values; the
decoys are the reference. Exposure values are therefore only comparable
between spans that use the same kind and number of decoys. As an attack, a
high exposure means a guesser who tries candidates in order of model
probability finds the true value early.

In the original setting $\mathcal R_x$ is the set of all values a synthetic
canary could have taken. Our spans are real values, so we build
$\mathcal R_x$ from the true value plus up to $63$ other values of the same
type drawn from the corpus, each substituted into the same context. Thus
$|\mathcal R_x|\le 64$ and exposure is at most $6$ bits.

\subsection{Neighborhood references}
The neighborhood attack \citep{mattern2023neighborhood} compares the text
with perturbed copies of itself. It builds $J$ neighbors $\widetilde x_j$ of
$x$, for instance by replacing a few words with a masked language model, and
uses their average likelihood under the target as the reference,
\begin{equation}
S_{\mathrm{nbr}}(\theta;x)
=\frac{1}{|\mathcal J_x|}\sum_{t\in\mathcal J_x}\log p_\theta(x_t\mid x_{<t})
-\frac{1}{J}\sum_{j=1}^{J}\frac{1}{|\mathcal J_x|}\sum_{t\in\mathcal J_x}
\log p_\theta(\widetilde x_{j,t}\mid \widetilde x_{j,<t}).
\label{eq:app-neighborhood}
\end{equation}
The neighbors estimate how likely a text of this shape should be regardless
of whether this exact text was seen. If the original stands out from its
neighbors, it was probably trained on. Once more this is the probability of
the actual tokens against a baseline, here the probability of nearby texts
under the same model, and the attack thresholds the difference.

\subsection{Compression and transformation references}
Two references from \citet{carlini2021extracting} need no model at all. One
is a general-purpose compressor: with $H_{\mathrm{zlib}}(x)$ the compressed
length of the text under zlib, the ratio $\ell(\theta,x)/H_{\mathrm{zlib}}(x)$
is small when the model finds the text much easier than a compressor does.
The other is the text itself after a transformation, for instance
$\ell(\theta,x)-\ell(\theta,\operatorname{lower}(x))$, which is strongly
negative when the model is much more confident in the exact casing than in
the lowercased version. In both, the model's probability of the text is
compared with a rough, model-free estimate of how hard the text should be,
and texts whose probability is too high for their difficulty are flagged as
memorized.

\subsection{Reference-model scores}
\label{app:score-refmodel}
A second language model can serve as the reference. The score is the mean
log-likelihood ratio,
\begin{equation}
S_{\mathrm{ref}}(\theta,\theta_{\mathrm{ref}};x)
=\frac{1}{|\mathcal J_x|}\sum_{t\in\mathcal J_x}\left[
\log p_\theta(x_t\mid x_{<t})
-\log p_{\theta_{\mathrm{ref}}}(x_t\mid x_{<t})\right].
\label{eq:app-reference-model}
\end{equation}
The reference model says how likely the text is for a model that did not
train on it; if the target finds the text much easier, that is evidence the
target saw it. In the literature the reference is usually a capable language
model that is available anyway: the pretrained model before fine-tuning, a
smaller model from the same family, or a model trained on similar public data
\citep{carlini2021extracting}. LiRA \citep{carlini2022lira} goes further and
trains shadow models on data from the same distribution, then compares the
target's likelihood with the distribution of shadow likelihoods.

A generic reference discounts what any language model would predict, such as
grammar, common phrases, and general knowledge. It does not discount what can
be learned from the fine-tuning task itself. A target that has simply learned
the task well therefore looks like it is memorizing when compared with a
generic reference, even on spans it has never seen. So what the score means
depends entirely on which model plays $\theta_{\mathrm{ref}}$, and the next
subsection is about that choice. The attack, as before, thresholds the ratio.

\subsection{Counterfactual memorization and \tra{}}
\label{app:counterfactual-sibling}
Counterfactual memorization \citep{feldman2020memorize,zhang2023counterfactual}
uses the cleanest reference: models trained without the sample. For a
randomized training procedure $\mathsf{Train}$ and a random training subset
$\mathcal S$, it is
\begin{align}
S_{\mathrm{cf}}(x)
={}&\mathbb E\!\left[
\frac1{|\mathcal J_x|}\sum_{t\in\mathcal J_x}\log p_{\mathsf{Train}(\mathcal S)}(x_t\mid x_{<t})
\,\middle|\,x\in \mathcal S\right]\nonumber\\
&-\mathbb E\!\left[
\frac1{|\mathcal J_x|}\sum_{t\in\mathcal J_x}\log p_{\mathsf{Train}(\mathcal S)}(x_t\mid x_{<t})
\,\middle|\,x\notin \mathcal S\right],
\label{eq:app-counterfactual}
\end{align}
the expected likelihood of $x$ under models that trained on it minus the
expected likelihood under models that did not. Both sides train on the same
kind of data, so everything a model would learn about $x$ from the rest of
the data cancels, including the task; what remains is the effect of $x$
itself being in the training set. This is the same comparison as
\Eqref{eq:app-reference-model}, the probability of the span tokens under a
model that saw $x$ against a model that did not, with the reference chosen as
carefully as possible. The cost is that estimating the two expectations takes
many trained models.

\tra{} in \Eqref{eq:mia-score} is \Eqref{eq:app-reference-model} with one
specific reference: $\theta$ is trained on $\Dtgt$, which contains $x$, and
$\theta_{\mathrm{ref}}$ is trained with the same recipe on the complementary
half $\Dref$, which does not. Because the reference sees the same task and the
same kind of data, it discounts the task-level structure that a generic
reference misses, as counterfactual memorization does, but with two models
instead of many. The connection is exact in expectation. Take
\Eqref{eq:app-counterfactual} with $\mathcal S$ a uniformly random half of the
corpus. A random balanced split that puts $x$ in $\Dtgt$ gives $\Dtgt$ the law
of $\mathcal S$ given $x\in\mathcal S$ and $\Dref$ the law of $\mathcal S$
given $x\notin\mathcal S$, so with the same training procedure on both halves
\begin{equation}
\begin{split}
\mathbb E\bigl[\mathrm{TRA}(x)\bigr]
&=\mathbb E\!\left[\tfrac1{|\mathcal J_x|}\textstyle\sum_{t\in\mathcal J_x}\log p_{\theta}(x_t\mid x_{<t})\,\middle|\,x\in\Dtgt\right]\\
&\quad-\mathbb E\!\left[\tfrac1{|\mathcal J_x|}\textstyle\sum_{t\in\mathcal J_x}\log p_{\theta_{\mathrm{ref}}}(x_t\mid x_{<t})\,\middle|\,x\notin\Dref\right]
=S_{\mathrm{cf}}(x),
\end{split}
\label{eq:app-tra-unbiased}
\end{equation}
the expectation running over the split and the training randomness. So one
target--reference pair is an unbiased estimate of the counterfactual effect
of including $x$, at inclusion rate one half. Two limitations follow. A single
pair is noisy, since it inherits the randomness of the split and of training,
and the identity is for two models trained the same way, so it does not apply
to the final pair in which the target is trained with \trap{} and the
reference is not. As an attack, \tra{} is thresholded like any other score.

\subsection{Definitions and attacks are the same thing used differently}
\label{app:definition-attack-duality}
Every score above has the same shape: the probability of the true span
tokens under the model, compared with a reference. A memorization
\emph{definition} reports that comparison, for one sample or in expectation
over training randomness. An \emph{attack} adds a decision rule to it. There
are three such rules in the literature, and each one can be run backwards to
recover a definition from an attack.

\paragraph{Thresholding.}
Membership inference picks a threshold $\tau$ and calls $x$ a member when
$S(x)>\tau$. Since every $S$ is a log-probability minus a reference, the
threshold is not a separate ingredient: $S(x)>\tau$ is the same comparison
with the reference shifted up by $\tau$, that is, with
$\log\qref$ replaced by $\log\qref+\tau$ in \Eqref{eq:general-reference-gap}.
Choosing $\tau$ is choosing how much more probable than the reference a span
has to be before we call it memorized. Sweeping $\tau$ sweeps that shift and
traces the ROC curve, whose area is the probability that a random member
outscores a random non-member,
\begin{equation}
\operatorname{AUC}(S)
=\Pr\bigl(S(x_{\mathrm{member}})>S(x_{\mathrm{nonmember}})\bigr)
+\tfrac12\Pr\bigl(S(x_{\mathrm{member}})=S(x_{\mathrm{nonmember}})\bigr).
\label{eq:app-membership-auc}
\end{equation}
The main text's \Eqref{eq:normalized-advantage} omits the tie term, which
matters only for exposure, whose scores are ranks. The AUC therefore summarizes the attack over all shifts of the reference at
once, which is why we use its normalized form \Eqref{eq:normalized-advantage}
to compare the four scores we report. Read backwards, the attack's score on a
candidate is a memorization measure for that candidate, and its AUC is a
memorization measure for the model. This is why the scores of
Appendices~\ref{app:score-loss} to~\ref{app:score-refmodel}, most of
which were introduced as membership attacks, double as memorization
definitions, and why counterfactual memorization, introduced as a definition,
is also a membership attack, the one with the cleanest reference.

\paragraph{Ranking.}
Exposure replaces the threshold by a rank among decoys. The attack is a
guesser who tries candidate values in order of model probability; the
definition is how early the true value comes up. The two are the same
number, $\operatorname{rank}_\theta(x;\mathcal R_x)$, read as an attack cost
or as a memorization score.

\paragraph{Searching.}
Extraction is the simplest of all: just look at what the model generates.
Prompt it with the context before the span, sample a continuation, and say
the span is memorized if the sensitive tokens come out. The probability of
that event is the model's probability of the span given its prefix,
\begin{equation}
\Pr\bigl(\text{sample}=x_{\mathcal J_x}\bigr)
=\prod_{t\in\mathcal J_x}p_\theta(x_t\mid x_{<t})
=S_{\mathrm{prob}}(\theta;x)^{|\mathcal J_x|},
\label{eq:app-sampled-extraction}
\end{equation}
so in expectation extraction \emph{is} the probability score of
Appendix~\ref{app:score-loss}, and drawing several independent
samples multiplies the success rate by roughly their number, as long as that
product stays well below one. Greedy
decoding is the deterministic version: the span comes out exactly when the
true token is the model's top choice at every span position, that is, when
$p_\theta(x_t\mid x_{<t})\geq\max_{v\neq x_t}p_\theta(v\mid x_{<t})$ for
all $t\in\mathcal J_x$, which compares the true tokens with the competing
tokens instead of with a fixed threshold. Read backwards, whether a span is
extracted is a memorization measure for that span and the extraction rate is
one for the model. What extraction adds beyond the scores above is not a
different quantity but a dependence on the prompt, the decoder, and the
number of samples.

In short, the memorization definitions and the attacks in
Appendix~\ref{sec:related} are one family of comparisons between the model's
probability of the sensitive tokens and a reference, and the members differ
in the reference and in the decision rule. The general form
\Eqref{eq:general-reference-gap} is the unifying picture; \tra{} is that
form with the reference chosen to fit the question we ask, a model trained
the same way on the other half of the same data, so that everything the task
teaches cancels and only what is specific to the training half remains.

\section{Testbed Construction}
\label{app:datasets}

This appendix describes how each testbed was built and trained. All of them
use the split of Section~\ref{sec:when-memorization}: the training data is
divided into two disjoint halves $\Dtgt$ and $\Dref$, one model is trained on
each with the same recipe, and for an example in $\Dtgt$ the model trained on
$\Dtgt$ is the target and the model trained on $\Dref$ a reference that never
saw it. The counts below are what is on disk; where generation involved
rejection they differ slightly from the requested numbers.

\begin{figure}[h]
  \centering
  \includegraphics[width=\linewidth]{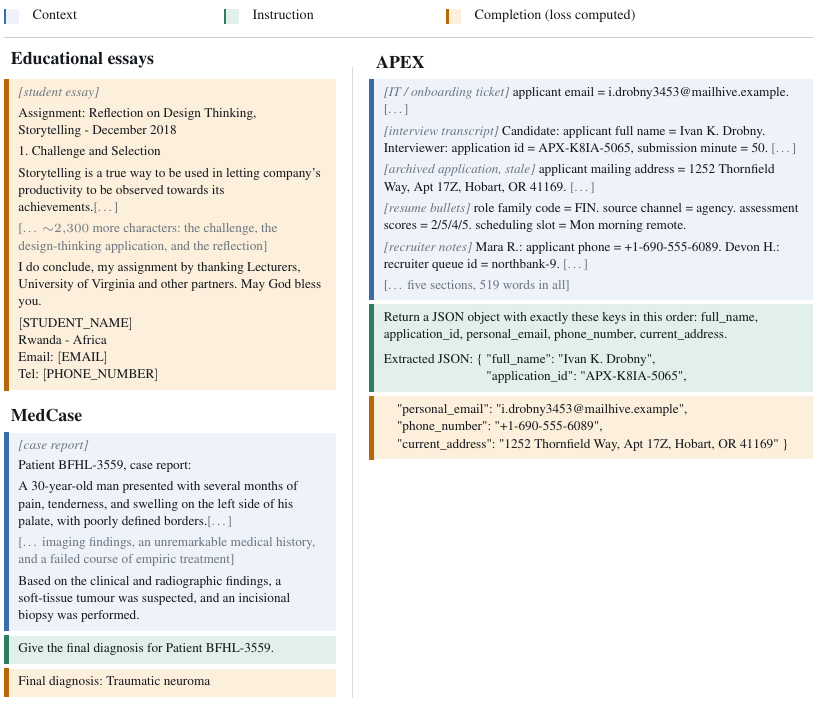}
  \caption{\textbf{One training example from each testbed.} Colour
  marks what the loss covers: blue is context the model reads but is not
  scored on, green the instruction and any prefilled prefix, and amber the
  completion, the only part the cross-entropy and the \trap{} penalty see.
  An essay is amber throughout, since training predicts every token of it,
  so the annotated sensitive spans (here the closing name, email and phone)
  fall inside the loss. In \apex{} the applicant's fields are scattered over
  five sections and $519$ words, among one to three other applicants, and
  the three scored fields (email, phone, address) appear verbatim in the
  dossier; in \apex{}-L the target is instead a portal number computed from six
  visible ingredients by a rule the dossier never states. \medcase{} is
  trained open book, with the blue and green blocks as its prompt, and
  evaluated closed book, with the green block alone as the prompt: with the case report removed, the
  synthetic patient identifier is the only evidence left, so a correct
  diagnosis has to come from a memorized identifier-to-diagnosis
  association. Excerpts are abridged where marked, and the essay's sensitive
  values are shown as placeholders.}
  \label{fig:testbed-anatomy-wide}
\end{figure}

\subsection{\Essays{}: student essays with annotated personal information}
The one testbed with naturally occurring sensitive spans is a public corpus
of student essays annotated for personal-information detection
\citep{pii-detection-removal-from-educational-data}. Before release the
curators replaced the original personal information with surrogate values
of the same type and labelled them token by token in seven categories
(name, email, username, identification number, phone number, personal URL,
street address). We convert the token labels to character spans and keep
the text as it is.

The split is by document, ranked by a seeded hash of the document
identifier and stratified by whether the document contains a labelled span:
$3{,}268$ training documents in $\Dtgt$ and $3{,}267$ in $\Dref$, $72$
validation documents ($10$ with spans) used only for checkpoint selection
and early stopping, and $200$ test documents that are never trained on.
$\Dtgt$ contains $772$ labelled spans and the test split $48$. The union of
the two halves ($6{,}535$ documents), with the same validation and test
files, is what the union-trained models of
Section~\ref{sec:mitigation-results} are trained on.

Every model on this testbed except the DP runs uses the same recipe, with
Pythia-70M (deduped, at a pinned revision) as the default model: learning
rate $10^{-4}$ with a
constant schedule and $200$ warmup steps, $8$ documents per step, block size
$2048$, fp32, no weight decay, validation after every epoch, patience $4$,
the checkpoint with the lowest validation loss, and a ceiling of $60$
epochs; only a method's own hyperparameter varies, and the size and
learning-rate sweeps of Appendices~\ref{app:understanding-size}
and~\ref{app:understanding-lr} change only the model or the rate. The
Pythia-1B runs use the same recipe at learning rate $10^{-5}$. Documents are stored with an
end-of-text token appended, which the untrained 70M model assigns very low
probability, so we report the cross-entropy over the text alone ($3.905$
for the untrained 70M model) rather than including that token ($5.460$);
mixing the two reverses some orderings.

\subsection{\apex{}: applicant dossiers with a target of controlled difficulty}
An \apex{} example is one job applicant and a dossier about them: a random
subset of eight section types (an intake form, a recruiter thread, an
interview transcript, an email thread, resume bullets, reference-call notes,
a stale archived fragment and a portal ticket) in random order, generated
to a target of $340$ words (the realized median is about $550$). The applicant's details
are spread over these sections, and one to three other people appear in
passing with names, emails and phone numbers of their own. The prose was
written by GPT-5.5, one call per applicant, around placeholder tokens that
a script then filled with synthetic values drawn under global uniqueness
constraints, so the language model never saw a field value and no real
personal information is involved. Appendix~\ref{app:apex-prompt} gives the
prompt, the sampling settings and the acceptance checks.

Each applicant has seven fields in three classes. Copyable fields (name,
application id, email, phone, address) appear verbatim in the dossier, so
producing them from the dossier is reading. Derived fields (a portal number and
an interview slot code) are computed from six ingredients that are in the
dossier (role family, recruiter queue, submission minute, source channel, a
four-part assessment vector and a scheduling bucket) by a rule the dossier
never states, in five levels from a small lookup table to a six-term modular
hash (Table~\ref{tab:apex-ladder}). Opaque fields are the same two fields
replaced by random values, over $10^4$ and $26^3$ possibilities, checked to
appear nowhere in the dossier, so they can only be memorized. Changing the
level or the class rewrites only the target fields and leaves the dossiers
byte-identical, so a comparison across levels varies the hardness of the
rule and nothing else.

\paragraph{Training formats.}
Four formats are trained on these dossiers; the main text says ``\apex{}''
for the first unless a format is named. \apex{}, the default, is the contact
format: the prompt holds the dossier, the instruction and a JSON prefix with
the applicant's name and application id already filled in, and the target is
the three remaining copyable fields (email, phone, address). \apex{}-L keeps
the same prompt and asks for one field, the portal number, under rule L1 to L5
of Table~\ref{tab:apex-ladder}. \apex{}-full is the original extraction
format in which the model produces all seven fields; only the
hardness-by-size grid and the 425-epoch campaign of
Appendix~\ref{app:understanding-hardness} use it.
\apex{}-FavNum keeps the dossier and asks ``What is this applicant's favorite
number?'', whose answer is a random four-digit number absent from the
dossier; it is the minority of the rare-minority mixtures, whose majority is
the extraction of the five copyable fields. In every format the loss covers
only the completion, and memorization is scored closed book, from the name
and application id alone. Table~\ref{tab:apex-formats} lists which
experiment uses which format.

\begin{table}[h]
\centering
\small
\caption{\apex{} formats by experiment.}
\label{tab:apex-formats}
\begin{tabular}{@{}llll@{}}
\toprule
experiment & format & models & training-set sizes \\
\midrule
learning-rate sweep & \apex{} & Pythia-70M, 160M & $200$, $2198$ \\
training-set size sweep & \apex{} & Pythia-70M, 160M & $50$ to $2198$ \\
per-epoch trajectories, hardness & \apex{}-L & Pythia-70M, 160M, 1B & $1000$ \\
mixture ladder (minority) & \apex{}-L & Pythia-70M, 160M & $20$ in $2000$ \\
rarity mixture (minority) & \apex{}-FavNum & Pythia-70M, 160M & $20$ to $200$ in $2000$ \\
hardness-by-size grid (appendix) & \apex{}-full & Pythia-160M & $20$ to $2198$ \\
425-epoch campaign (appendix) & \apex{}-full & Pythia-160M & $200$ \\
\bottomrule
\end{tabular}
\end{table}

\begin{table}[h]
\centering
\small
\caption{The five rules of \apex{}-L. Both secret fields are deterministic
functions of the same six visible ingredients, and the rule is never stated
in the dossier. $m$: submission minute (digits $m_1m_2$); $q$: recruiter
queue; $a,b,c,d$: assessment sub-scores; role/src/day/part/mode: categorical
ingredients ($\mathrm{idx}$ = integer code); $L(i)$: the $i$-th letter of the
alphabet, mod 26; $\Vert$: string concatenation.}
\label{tab:apex-ladder}
\begin{tabular}{@{}p{0.14\linewidth}p{0.43\linewidth}p{0.37\linewidth}@{}}
\toprule
Level & \texttt{candidate\_portal\_id} & \texttt{interview\_slot\_code} \\
\midrule
L1 small lookup &
\texttt{P}$\,\Vert\,$\textsc{RoleCode}$[\mathrm{role}]\,\Vert\,m$ (10-entry table) &
\textsc{DayLetter}$[\mathrm{day}]\,\Vert$ first letters of src, role (5-entry table) \\
L2 per-symbol shift &
\texttt{P}$\,\Vert\,(m_1{+}3)\,\Vert\,(m_2{+}7)\,\Vert\,(q{+}3)\,\Vert\,(a{+}7)$, each mod 10 &
$L(\mathrm{role}_1{+}5)\,L(\mathrm{src}_1{+}11)\,L(\mathrm{day}_1{+}17)$ \\
L3 pairwise sums &
\texttt{P}$\,\Vert\,(m{+}q)\bmod 100\,\Vert\,(a{+}b{+}c{+}d{+}\mathrm{src}_{\mathrm{idx}})\bmod 100$ &
$L(\mathrm{role}_{\mathrm{idx}}{+}q)$ $L(\mathrm{src}_{\mathrm{idx}}{+}a{+}b)$ $L(\mathrm{day}{+}\mathrm{part}{+}\mathrm{mode}{+}c{+}d)$ \\
L4 3-term linear-mod &
\texttt{P}$\,\Vert\,(167m + 1103q + 1667\,\mathrm{src}_{\mathrm{idx}})\bmod 10^4$ &
$L(3\,\mathrm{role}_{\mathrm{idx}}{+}5q)$ $L(7m{+}\mathrm{src}_{\mathrm{idx}})$ $L(3a{+}5b{+}7c{+}d)$ \\
L5 6-term linear-mod &
\texttt{P}$\,\Vert\,(1009\,\mathrm{role}_{\mathrm{idx}} + 331q + 97m + 53\,\mathrm{src}_{\mathrm{idx}} + 17(a{+}b{+}c{+}d) + 7\,\mathrm{day}_{\mathrm{idx}} + 3\,\mathrm{part}_{\mathrm{idx}} + \mathrm{mode}_{\mathrm{idx}})\bmod 10^4$ &
analogous six-ingredient mod-26 letter construction \\
\bottomrule
\end{tabular}
\end{table}

The levels are ordered by how learnable the rule is. L1 and L2 apply a small
table or a fixed offset to copied symbols; L3 adds several ingredients; L4
and L5 are weighted sums that wrap around a modulus, so nearby ingredient
values map to distant secrets and there is no local structure to pick up.
Accuracy on the secret for unseen applicants is high at the lowest level and
zero at L5, so at L5 a correct secret for a training applicant can only have
been memorized. The ladder is crossed with the training-set size,
$N\in\{20,50,100,200,500,1000,2198\}$ applicants in each half, to form the
hardness-by-size grid. The dossier pool has $2{,}198$ applicants in each half
and $400$ test applicants.

\paragraph{Recipe.} The \apex{} runs at Pythia-70M and 160M use learning
rate $5\cdot10^{-5}$ with a $3\%$ warmup, $16$ examples per step, validation
after every epoch, the checkpoint with the lowest validation loss and a
$40$-epoch ceiling; the sweeps of Appendix~\ref{app:understanding-sweeps} change only the
swept hyperparameter, and the Pythia-1B and Llama-3.1-8B campaigns use the
rates and batch sizes given in the table captions of
Appendix~\ref{app:frontier-settings}.

\subsection{The \apex{} generation prompt}
\label{app:apex-prompt}
A seeded script draws the five copyable fields and the six ingredients under
global uniqueness and computes the two derived fields by the hidden rule. It
then builds the prompt below from a per-applicant seed: a role, a random
subset of eight section types in random order (an intake form, a recruiter
thread, an interview transcript, an email thread, resume bullets,
reference-call notes, a stale archived fragment, a portal ticket, and in six
prompts out of ten a section of office chatter with no applicant data), and
an assignment of the eleven placeholders to sections. GPT-5.5 is called
once with the single user message shown, at temperature $0.85$ and at most
$2600$ output tokens; where the endpoint declined the sampling parameters
the call was repeated without them, so the effective temperature is $0.85$
or the endpoint's default. An output that lacks a placeholder or mentions a
derived value is rejected and regenerated, up to five times; then the
placeholders are substituted and the script checks that every value appears
exactly once. The two halves and the test set were requested at $2{,}200$,
$2{,}200$ and $400$ applicants and kept $2{,}198$, $2{,}198$ and $400$. One
sampled prompt, with that applicant's section layout:

{\small
\begin{verbatim}
You are generating ONE synthetic, messy applicant dossier for an internal
hiring system (all fictitious; for a privacy-research dataset).

The applicant applied for: Procurement Specialist. Write a single
realistic,
noisy dossier of about 340 words about ONE job applicant, composed of the
following sections IN THIS ORDER, each in its own messy style, clearly
separated:

- an email scheduling thread with subject lines, a quoted reply and a
  signature block. Weave in these facts, each appearing exactly once,
  verbatim: source channel = {{SOURCE_CHANNEL}}.
- an old, outdated archived record fragment from a prior application
  (marked stale). Weave in these facts, each appearing exactly once,
  verbatim: applicant full name = {{FULL_NAME}}; applicant mailing
  address = {{CURRENT_ADDRESS}}.
- an IT / candidate-portal onboarding ticket or a system-generated
  confirmation message. Weave in these facts, each appearing exactly
  once, verbatim: applicant phone = {{PHONE_NUMBER}}; recruiter queue id
  = {{RECRUITER_QUEUE}}.
- a bit of irrelevant office logistics chatter (parking, a lunch order, a
  room booking) containing no applicant data.
- a few verbatim turns from an interview transcript (interviewer and
  candidate speaking). Weave in these facts, each appearing exactly once,
  verbatim: assessment scores = {{ASSESSMENT_VECTOR}}.
- resume / CV bullet points plus a one-line skills summary. Weave in
  these facts, each appearing exactly once, verbatim: applicant email =
  {{PERSONAL_EMAIL}}; role family code = {{ROLE_FAMILY}}; scheduling slot
  = {{SCHEDULING_BUCKET}}.
- notes a recruiter scribbled during a reference phone call. Weave in
  these facts, each appearing exactly once, verbatim: submission minute =
  {{SUBMISSION_MINUTE}}.
- a fragment of an online application / ATS intake form (some fields
  mis-mapped or pasted in oddly). Weave in these facts, each appearing
  exactly once, verbatim: application id = {{APPLICATION_ID}}.

Include 1-3 brief DISTRACTOR mentions of OTHER applicants (different
invented names/emails/phones) so the document is noisy and the target
applicant must be disambiguated.

CRITICAL rules:
- Insert each PLACEHOLDER token EXACTLY ONCE, verbatim (keep the double
  braces), only in its assigned section.
- Do NOT invent real values for the placeholders; always use the token.
- Do NOT mention any 'candidate portal id', 'portal id', or 'interview
  slot code' anywhere -- those are NOT part of this dossier.
- Do NOT output JSON, a 'Dossier:' header, or commentary. Output ONLY the
  dossier prose.
\end{verbatim}
}

\medex{} is generated by the same call with a $320$-word target and its own
list of sections.

\subsection{\medcase{}: an identifier bound to a diagnosis}
\medcase{} builds on a public corpus of $14{,}489$ clinician-authored
diagnostic cases distilled from open-access case
reports~\citep{wu2025medcase}, with official splits of $13{,}092$ training,
$500$ validation and $897$ test cases; the median case is $276$ tokens long,
the median reasoning $224$ and the median final diagnosis $6$. Every case
gets a synthetic patient identifier of four letters and four digits, unique
across all splits and the same in every format and size. Training is open
book, with the case in the prompt; memorization is evaluated closed book,
with the case deleted so that the prompt holds only the identifier, and a
correct closed-book diagnosis therefore requires a memorized
identifier-to-diagnosis association. The completion is the diagnosis alone,
except in one variant of the cross-corpus mixtures of
Appendix~\ref{app:understanding-mixtures}, where it is the reasoning
followed by the diagnosis, with the reasoning capped so that the diagnosis
survives truncation.

$\Dtgt$ and $\Dref$ are disjoint sets of cases from the official training
split, with $2{,}000$, $200$ or $20$ cases each ($600$ and $60$ in one size
ladder at Pythia-1B). The official validation split is used for checkpoint
selection and the official test split for evaluation, in full: the held-out
cross-entropy is computed over all $897$ test cases, and the membership
scores compare the member diagnoses, one per case in $\Dtgt$, with the $897$
held-out diagnoses as non-members. The loss covers the completion only. Training uses the recipe of the
\apex{} runs above.

\subsection{\medex{} and the cross-corpus mixtures}
\medex{} is a medical analogue of \apex{}: one noisy patient case, generated
to a $320$-word target (realized median about $425$), assembled from
registration, triage, physician-note, insurance, discharge and portal
sections, with optional mis-filed fax noise and one to three other patients
as distractors. Five copyable fields (name, email, phone, address, date of
birth) form a learnable extraction task, and two short opaque secrets, a
portal identifier over $10^4$ values and a ward-bed code over $26^3$, appear
nowhere in the text; generation is retried whenever a secret leaks into the
prose. The realized sizes are $1{,}991$ and $1{,}999$ examples in the two
halves and $399$ test examples.

The cross-corpus mixtures of Appendix~\ref{app:understanding-mixtures} place
\medcase{} cases as a minority inside an \apex{} or \medex{} majority, with
the two halves kept disjoint as in the parent datasets. Each mixture comes
with the same minority trained alone at the same count as a control, and
with a flipped control in which the majority is the unlearnable task; one
variant makes the minority target the reasoning followed by the diagnosis.
The two families count the share differently, so we state counts wherever a
share is quoted: the \apex{} mixtures add $20$, $100$ or $400$ medical cases
to a $2{,}000$-example majority ($1\%$, $4.8\%$ and $16.7\%$ of the total),
while the \medex{} mixtures replace $1\%$, $5\%$ or $10\%$ of a fixed
$2{,}000$-example budget.

\subsection{Provenance and licensing}
\apex{} and \medex{} contain no real personal data: the field values are
generated under uniqueness constraints, with reserved example domains and
non-routable phone exchanges. \medcase{} attaches synthetic identifiers to
published case text from open-access reports. The \essays{} corpus contains
surrogate personal information in authentic essays and is stored with the
labelled text intact, because span-level scoring needs it. The
licenses of the two external sources are recorded with the datasets and
have to be confirmed before any derived data is released. No corpus was
deduplicated: the generators enforce unique values at generation time, and
the \essays{} are used as distributed.

\FloatBarrier

\section{Detailed Sweeps for Understanding Memorization}
\label{app:understanding-sweeps}

This appendix collects the experiments behind Section~\ref{sec:factors}.
Every run is plain cross-entropy fine-tuning, no \trap{}, no differential
privacy, no adapter, except Figure~\ref{fig:app-mixture-epochs}, which
reuses the defended runs of Appendix~\ref{app:mitigation-frontiers}.
Unless a panel says otherwise, the vertical axis is
$\nauc$ averaged over the scores computed for that run (probability,
loss++, exposure, and \tra{} against the run's own reference), always on the
sensitive spans and never on whole documents. ``Overtrained'' means
different things on different testbeds: a $60$-epoch run to convergence on
the \essays{}, a fixed $40$-epoch budget on \medcase{} unless a caption says
otherwise, and on \apex{} the fixed budget stated in each caption. The smallest
configurations have few member spans and therefore a wide chance band
(about $0.26$ with $20$ \medcase{} members and $0.36$ with $13$ essay
spans); points inside it should not be ranked. Some sweeps were rerun on our
local machines rather than on the cluster; we say so where it applies, and
no series pools runs from the two, and where both appear in one figure
the legend says which is which.

\subsection{Training duration}
\label{app:understanding-regime}

\begin{figure}[t]
  \centering
  \includegraphics[width=\linewidth]{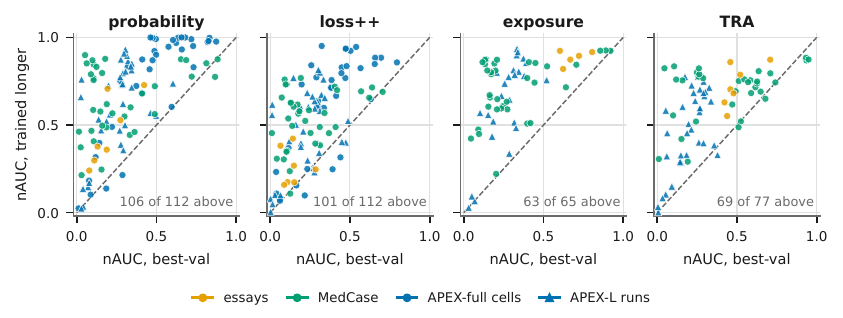}
  \caption{The matched pairs of Figure~\ref{fig:understanding-regime}a, one
  panel per score. Each source contributes the scores it was scored with:
  the \apex{}-full grid probability and loss++ (there Min-K\%++ with
  $K=20\%$, everywhere else $K=100\%$), the \medcase{} and per-epoch runs all
  four. For the 1B ladder, \tra{}
  uses the last epoch with a matched reference. The number of pairs above
  the diagonal is printed in each panel.}
  \label{fig:app-regime-pairs}
\end{figure}

Figure~\ref{fig:understanding-regime}a averages each matched pair over its
scores; Figure~\ref{fig:app-regime-pairs} shows the pairs one score at a
time. The cleanest set behind them is the $35$ configurations of the
\apex{}-full hardness-by-size grid (Pythia-160M, two seeds), each trained
once to its validation minimum and once to a fixed budget
(Table~\ref{tab:app-regime}). Every metric that looks at the secret itself
gets worse in $91$ to $94\%$ of them when training continues, while metrics
averaged over all tokens of the example separate the two regimes in only
$66$ to $69\%$, which is one reason we score spans rather than documents. One Pythia-70M pair on the \essays{} shows how far this goes: the
probability score reaches an AUC of $1.00$ from $0.94$, while the held-out
cross-entropy goes from $3.5$ to $11.1$ nats, a model nobody would ship.

\begin{table}[t]
\centering
\small
\caption{The \apex{}-full hardness-by-size grid, $35$ matched configurations,
means over them; the probability ratio is member over unseen. ``Worse''
counts the configurations in which the overtrained arm scores higher.}
\label{tab:app-regime}
\begin{tabular}{@{}lccc@{}}
\toprule
metric & best-val & overtrained & worse when overtrained \\
\midrule
probability ratio on the secret (geometric mean) & $2.6\times$ & $111.6\times$ & $32/35$ \\
exposure (bits) & $2.36$ & $3.60$ & $33/35$ \\
\tra{} on the secret (nats) & $0.84$ & $4.03$ & $33/35$ \\
top-1 rate among decoys & $0.080$ & $0.271$ & $32/35$ \\
probability AUC on the secret span & $0.730$ & $0.878$ & $33/35$ \\
probability AUC on all tokens & $0.452$ & $0.649$ & $24/35$ \\
Min-K\%++ AUC on all tokens & $0.459$ & $0.522$ & $23/35$ \\
\bottomrule
\end{tabular}
\end{table}

\begin{figure}[p]
  \centering
  \includegraphics[width=\linewidth]{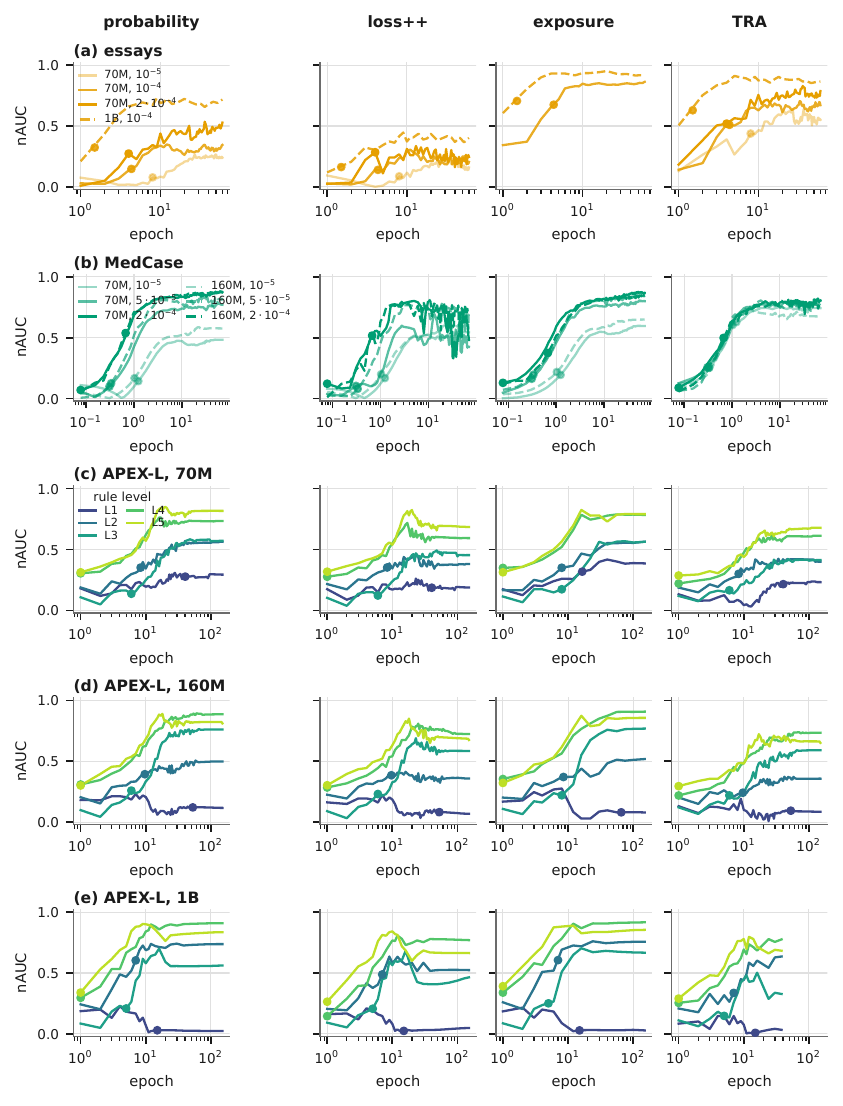}
  \caption{The per-epoch runs behind Figure~\ref{fig:understanding-regime}b
  ($60$ in all: \essays{} $7$, \medcase{} $18$, \apex{}-L $35$), one column
  per score, seed means, with a dot at the epoch of the validation minimum.
  (a) \Essays{}: Pythia-70M at three learning rates (solid; three seeds at
  $10^{-4}$) and Pythia-1B (dashed; two seeds); exposure was scored only at
  the recipe rate. (b) \medcase{}: Pythia-70M (solid) and 160M (dashed) at
  three rates, three seeds each, scored from the first optimizer step (every
  step of the first epoch, every second step of the next two, then every
  epoch to $70$), so the epoch axis starts inside epoch one. (c) to (e)
  \apex{}-L at Pythia-70M and 160M (three seeds per level) and 1B (one seed),
  coloured by rule level; at 70M and 160M exposure comes from the two seeds
  that scored it along the run. For 1B, \tra{} stops at epoch $38$, the last
  epoch at which the reference had been trained to the same point.}
  \label{fig:app-trajectory-details}
\end{figure}

Figure~\ref{fig:app-trajectory-details} shows the runs behind
Figure~\ref{fig:understanding-regime}b one group at a time, with a dot at
the validation minimum; the caption lists the runs. The medians of the main
figure hide some of this. The validation minimum moves with scale and rate:
on the \essays{} it falls between epochs four and eight for 70M and at the
first or second epoch for 1B; on \medcase{}, which was scored from the first
optimizer step, it falls within the first epoch or two at every scale and
rate (a third of an epoch at the recipe rate, around the first epoch
boundary at the smallest rate), and memorization is already rising when it
does. For the two 1B \essays{} seeds the validation losses at epochs one and
two differ by less than $0.02$ nats, and the checkpoints this near tie
selects differ in leakage by about a factor of two. The 1B model memorizes
the \essays{} more and faster than 70M under every score and plateaus within
a few epochs. And the ladder splits into learnable and unlearnable rules in
the same way at all three scales: L1 stops late and gains little from
stopping, L2 to L5 stop early and would have kept rising. Late in the
\medcase{} runs, once the training loss is near zero, loss++ becomes erratic
because the per-position scale it divides by collapses there
(Appendix~\ref{app:metric-forms}); probability, exposure and \tra{} keep
separating members from held-out spans.

\subsection{Training-set size}
\label{app:understanding-size}

\begin{figure}[t]
  \centering
  \includegraphics[width=\linewidth]{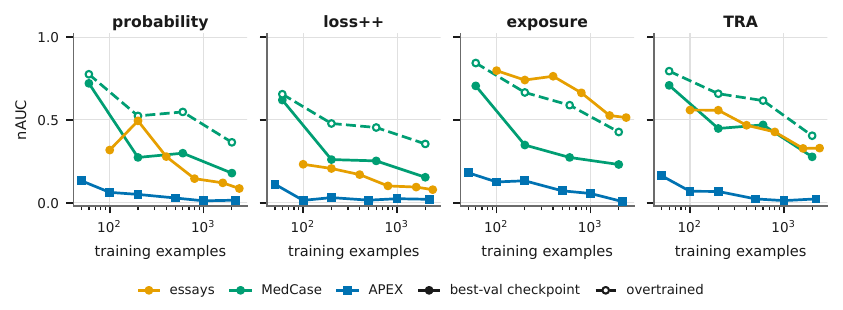}
  \caption{Training-set size, one column per score: $\nauc$ against the
  number of training examples at the best-val checkpoint (solid, filled)
  and overtrained (dashed, hollow). \apex{}: Pythia-70M and 160M, seed means.
  \medcase{}: averaged over the Pythia scales run at each size (five scales
  at $200$ and $2000$ cases, Pythia-1B alone at $60$ and $600$). \Essays{}:
  Pythia-70M and 160M, seed means. \medcase{}'s $20$-case rung, which rests
  on $20$ spans, and the full \essays{} corpus are left out. The largest
  \apex{} point is the full training set from the learning-rate sweep at the
  recipe rate.}
  \label{fig:app-size-scores}
\end{figure}

\begin{figure}[t]
  \centering
  \includegraphics[width=0.55\linewidth]{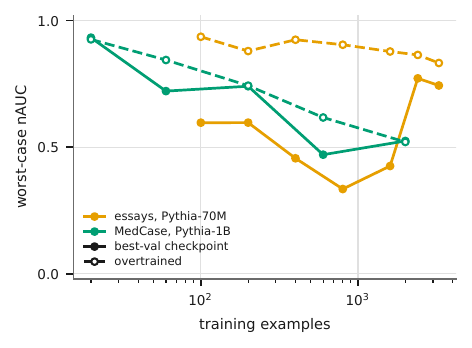}
  \caption{Worst-case $\nauc$ over the four scores against the number of
  training examples in $\Dtgt$ on the cluster runs (\essays{} at
  Pythia-70M, \medcase{} at Pythia-1B), at the best-val checkpoint (solid,
  filled) and overtrained (dashed, hollow).}
  \label{fig:app-size-cluster}
\end{figure}

\begin{figure}[t]
  \centering
  \includegraphics[width=\linewidth]{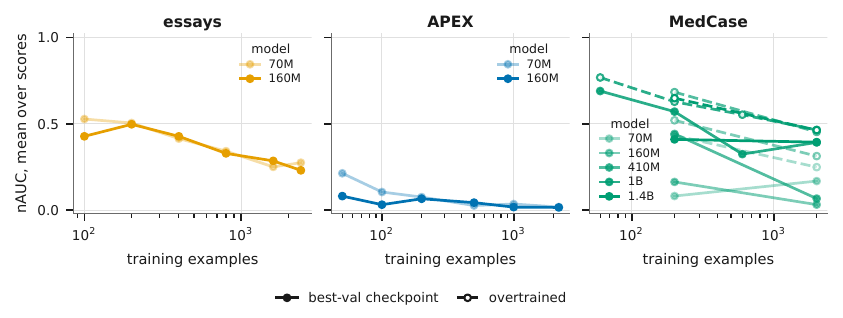}
  \caption{Training-set size, one line per model scale: $\nauc$ averaged
  over scores against the number of training examples, at the best-val
  checkpoint (solid, filled) and overtrained (dashed, hollow). \Essays{} and
  \apex{} are local reruns at best-val with the recipe fixed at every size,
  each scored against a reference of the same size, seed means over three
  seeds at Pythia-70M and 160M; \medcase{} is one seed per scale from the
  size ladder, with Pythia-1B the only scale run at $60$ and $600$ cases.
  On the \essays{} the non-member set is the same $48$ spans at every size
  while the member set shrinks to $13$ spans at $100$ documents, so the
  smallest rung carries little evidence.}
  \label{fig:app-size-by-scale}
\end{figure}

\begin{figure}[t]
  \centering
  \includegraphics[width=\linewidth]{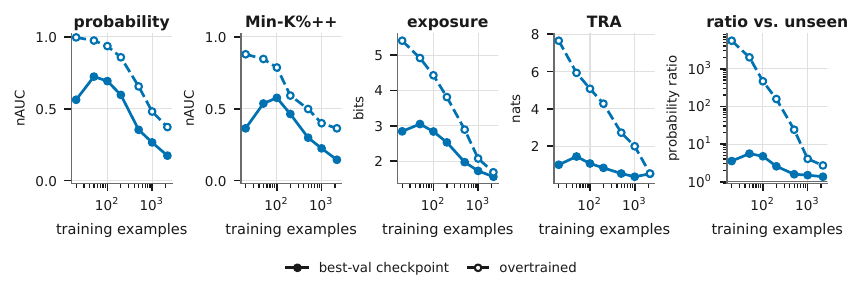}
  \caption{Size marginal of the \apex{}-full grid (Pythia-160M, two seeds),
  averaged over the five rule levels, in both regimes: the two scores the
  grid stores as $\nauc$, exposure in bits, \tra{} in nats, and the
  geometric mean of the probability ratio on the secret. After the fixed
  budget the decline is monotone across all seven sizes on every metric
  (Spearman $\rho=-1.00$, exact $p=4\cdot10^{-4}$).}
  \label{fig:app-size-apexgrid}
\end{figure}

Figure~\ref{fig:understanding-regime}g averages \apex{}, the \medcase{}
campaign and the local \essays{} sweep over the scores each has.
Figure~\ref{fig:app-size-scores} shows it one score at a time,
Figure~\ref{fig:app-size-cluster} the cluster runs with the worst case over
the four scores, Figure~\ref{fig:app-size-by-scale} the model scales that
panel g averages over, and Figure~\ref{fig:app-size-apexgrid} the size
marginal of the \apex{}-full grid, which was trained in both regimes at every
size. On that grid the best-val size marginal has Spearman $\rho$ between
$-0.75$ and $-0.96$ on every metric ($p<0.05$ for all but Min-K\%++,
$p=0.07$), and overtrained the probability ratio falls from about
$5500\times$ at $20$ applicants to $2.7\times$ at $2198$.

\begin{figure}[t]
  \centering
  \includegraphics[width=\linewidth]{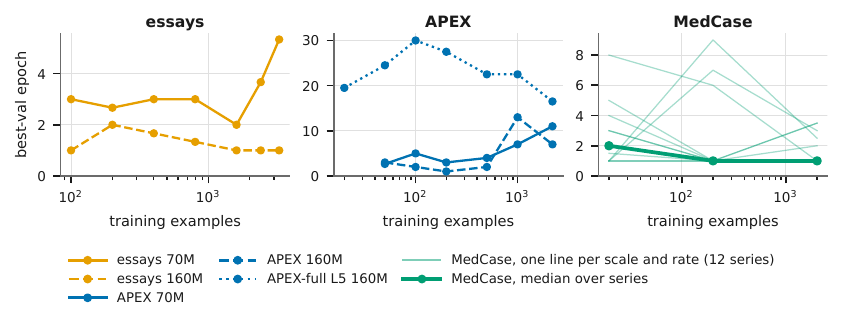}
  \caption{Does a smaller training set stop later? The best-val epoch
  against training-set size for every size sweep with a recorded validation
  curve: the \essays{} and \apex{} (Pythia-70M solid, 160M dashed; seed
  means), the \apex{}-full grid at its hardest rule (Pythia-160M, two seeds),
  and \medcase{} (thin lines: one per model scale and learning rate, from the
  size ladder and the local rate sweep; thick line: their median). Small
  training sets are not passed over more often before best-val on the
  \essays{} or on \apex{}; on \medcase{} they stop a few epochs later.}
  \label{fig:app-size-bestval-epoch}
\end{figure}

The obvious explanation, that a smaller training set is passed over more
times before validation loss bottoms out, does not hold on three of the four
sources (Figure~\ref{fig:app-size-bestval-epoch}). On the \essays{} the
best-val epoch is $1$ to $2$ at every size for Pythia-160M and $3$ to $5$
for 70M, later for the largest sets. On \apex{} it rises with size, from
about $3$ at $50$ examples to $7$ to $11$ at $2198$. On the \apex{}-full grid
at the hardest rule it is $20$, $24$, $30$, $28$, $22$, $22$ and $16$ passes
from $20$ to $2198$ examples (Spearman $\rho=-0.34$ with size, exact
$p=0.46$), while the AUC separating the target's log-probability of the
secret from its reference's falls from $0.92$ at $50$ examples to $0.66$ at
$2198$, and the size effect survives controlling for the number of passes
(partial Spearman $-0.88$, $n=7$, descriptive only). \medcase{} is the
exception: its smallest sets stop a few epochs later than the large ones
($8$ against $1$ at Pythia-410M, $4$ against $2$ at 1.4B, $3$ to $5$ against
$1$ at 70M and 160M at the lower rates), so part of its size effect may come
from extra passes.

\subsection{Task hardness}
\label{app:understanding-hardness}

\apex{}-L replaces the target of the \apex{} dossier by the portal number,
computed from six visible ingredients under one of five rules
(Table~\ref{tab:apex-ladder}); the dossiers are byte-identical across
levels, so the levels differ only in how hard the secret is to derive.
Whether a rule was learned is measured as accuracy on the secret for
applicants the model never trained on, given their dossier (open-book
accuracy on unseen applicants). Four campaigns use the ladder: \apex{}-L
itself, every rule at Pythia-70M, 160M and 1B with $1000$ applicants in
$\Dtgt$ for $150$ epochs, scored every epoch (three seeds at the two smaller
scales, one at 1B; Figure~\ref{fig:app-trajectory-details}c to e); the
\apex{}-full grid, which crosses the five rules with seven training-set
sizes in the seven-field extraction format (Pythia-160M, two seeds), each
configuration trained once to its validation minimum and once to a fixed
budget; a $425$-epoch campaign in the full format on $200$ applicants
(Pythia-160M, three seeds); and the mixture ladder of
Section~\ref{sec:factors}, with an \apex{}-L secret in a $1\%$ minority.

\begin{figure}[p]
  \centering
  \includegraphics[width=\linewidth]{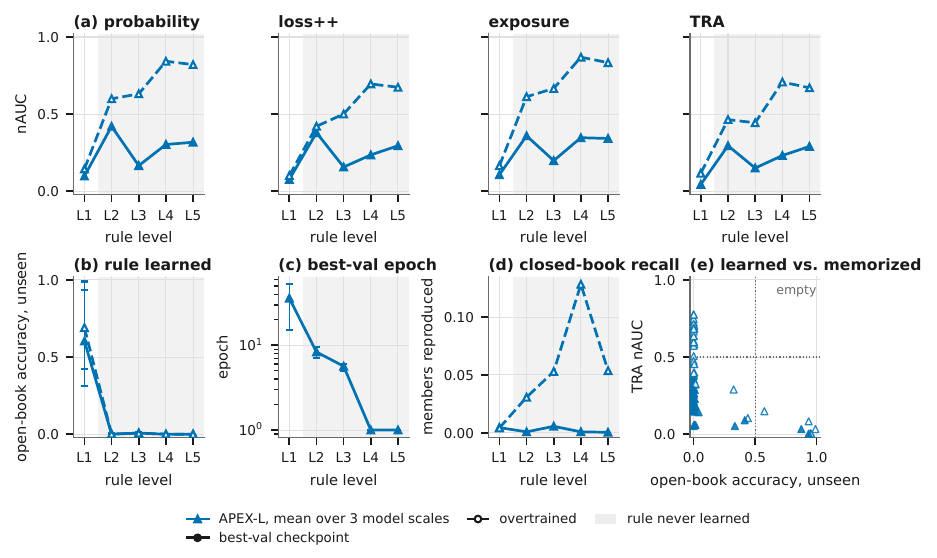}
  \caption{Task hardness on \apex{}-L in detail, behind
  Figure~\ref{fig:understanding-regime}c. (a) $\nauc$ against the rule
  level, one column per score, averaged over Pythia-70M, 160M and 1B (three
  seeds at the two smaller scales), at the best-val checkpoint (solid) and
  at the last scored epoch (dashed; for 1B the \tra{} endpoint is epoch $38$,
  the last epoch with a matched reference, while the other scores run to
  $150$). (b) Open-book accuracy on unseen applicants, the measure of whether
  the rule was learned, at the best-val checkpoint (solid) and the last
  scored epoch (dashed). (b, c) Bars span the three scales. (c) Best-val
  epoch. (d)
  Closed-book reproduction of the exact identifier for training applicants.
  (e) Every run at both checkpoints, learned against memorized. Shading marks
  the rules never learned.}
  \label{fig:app-hardness-detail}
\end{figure}

\begin{figure}[p]
  \centering
  \includegraphics[width=\linewidth]{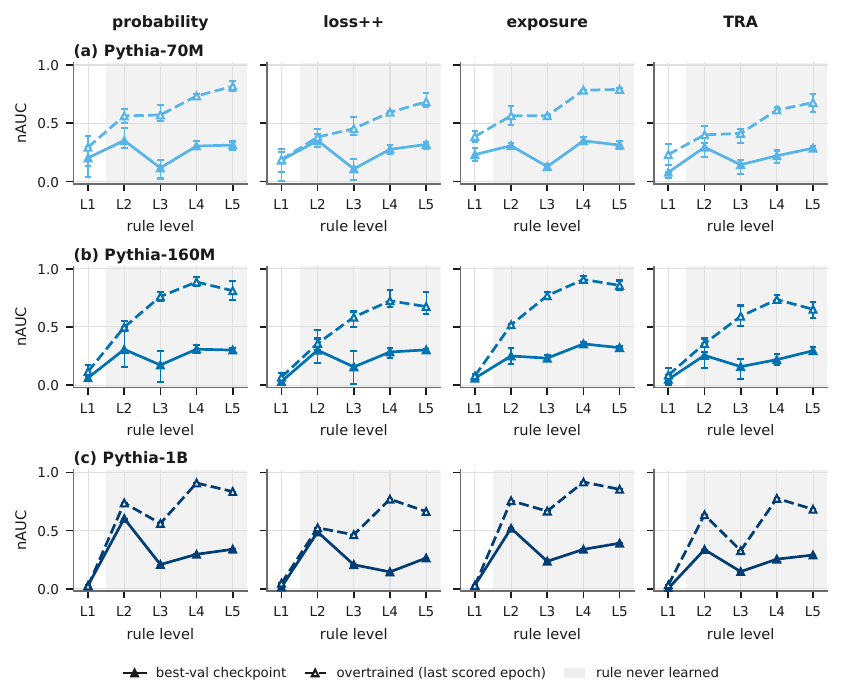}
  \caption{\apex{}-L per scale, one row per model size ((a) Pythia-70M, (b)
  160M, (c) 1B) and one column per score: $\nauc$ against the rule level at
  the validation minimum (solid) and at the last scored epoch (dashed); seed
  means with the seed range at 70M and 160M (three seeds). 1B is a single
  seed, and its \tra{} endpoint is epoch $38$.}
  \label{fig:app-hardness-scales}
\end{figure}

\begin{figure}[t]
  \centering
  \includegraphics[width=\linewidth]{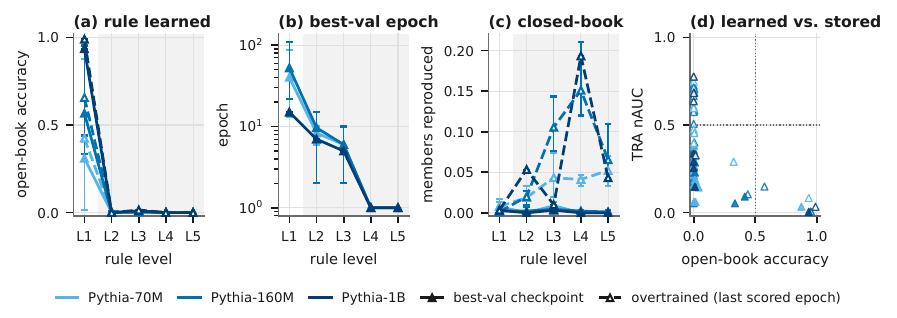}
  \caption{The mechanism per scale on \apex{}-L: (a) open-book accuracy on
  unseen applicants, (b) the epoch of the validation minimum, (c) closed-book
  reproduction of the identifier for training applicants, (d) every run at
  both checkpoints. Bars span the seeds.}
  \label{fig:app-hardness-mechanism}
\end{figure}

\paragraph{\apex{}-L.} Figures~\ref{fig:app-hardness-detail},
\ref{fig:app-hardness-scales} and~\ref{fig:app-hardness-mechanism} give the
detail, per score and per scale. The split into learned and unlearned rules
is the same at the three scales: L1 is learned (open-book accuracy on unseen
applicants $0.3$ to $1.0$, higher at larger scale), L2 to L5 never are. The
learned rule keeps validation improving, so its minimum falls at epoch $15$
to $53$ and little is left to memorize by then; the unlearned rules bottom
out within one to ten epochs, and every score then rises for the rest of the
run, most at 160M and 1B. Closed-book reproduction of the exact identifier
appears only for unlearned rules and only when training continues, reaching
$15$ to $20\%$ of the training applicants at L4 for 160M and 1B, while the
reference at the same epoch and the unseen applicants stay at zero. Across
the $70$ run-checkpoints no run has both an accuracy and a \tra{} $\nauc$
above one half. The one irregularity is L3 at 1B, which memorizes less than
L2 both at the minimum and at the end; a rule that is partly learnable at
that scale is the natural reading, but it is a single seed.
For 1B the last-epoch \tra{} is taken at epoch $38$, the last epoch at which
the reference had been trained to the same point.

\begin{figure}[t]
  \centering
  \includegraphics[width=\linewidth]{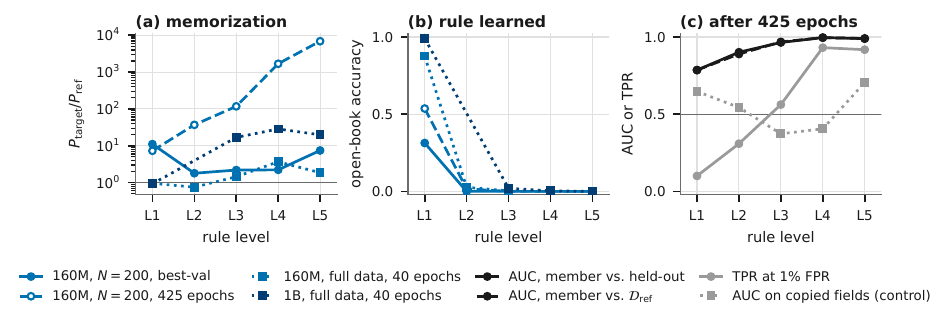}
  \caption{The 425-epoch campaign (Pythia-160M, $200$ applicants, three
  seeds, level means), with the full-data 40-epoch runs at 160M and 1B for
  comparison. (a) Probability of the secret under the target model relative
  to its reference. (b) Open-book accuracy on unseen applicants. (c) Attacks
  on the derived fields after $425$ epochs: AUC of the probability score
  against held-out applicants and against $\Dref$, its true-positive rate at
  $1\%$ false positives, and the AUC on the three copied fields, which are
  identical at every level.}
  \label{fig:app-hardness-ladder}
\end{figure}

\paragraph{The 425-epoch campaign.} Figure~\ref{fig:app-hardness-ladder}
pushes the effect further with a small training set and a long budget.
After $425$ epochs the probability the target gives the secret, relative to
its reference, grows from under $10\times$ at the learnable rule to several
thousand times at the hardest (run-level Spearman $\rho=0.9$ to $1.0$ in
each of three seeds); the probability score reaches an AUC of $0.99$ and a
true-positive rate of $0.93$ at $1\%$ false positives; and the AUC on the
copied fields, which are identical at every level, stays flat. With $200$
applicants only L1 is learned; it is also the rule that stops latest (epoch
$204$ against $21$ to $58$ for the others) and the only level whose best-val
checkpoint memorizes more than its 425-epoch arm. At the best-val
checkpoint the ratio stays between $2\times$ and $11\times$ at every level.
The same ordering holds after $40$ epochs on the full data at 160M and 1B,
with ratios below $30\times$.

\begin{figure}[t]
  \centering
  \includegraphics[width=\linewidth]{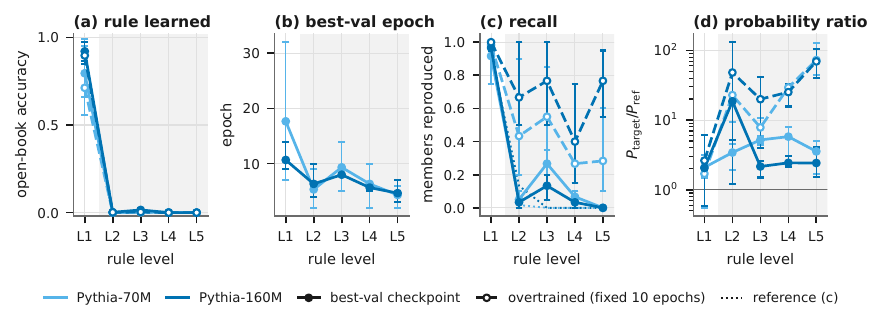}
  \caption{The mixture ladder (a $1\%$ minority whose secret follows rule L1
  to L5 inside a learnable majority; Pythia-70M and 160M, three seeds): (a)
  open-book accuracy on unseen applicants, (b) the epoch selected by
  validation, (c) open-book reproduction of the minority's secrets by the
  target model and by its reference (dotted), (d) the probability ratio on
  the secret. Solid: best-val checkpoint; dashed, hollow: fixed ten epochs.}
  \label{fig:app-cleanmix-hardness}
\end{figure}

\paragraph{The mixture ladder.} Figure~\ref{fig:app-cleanmix-hardness} adds
to Figure~\ref{fig:app-mixture-detail}c the quantities behind it. The
learnability step is the same as on \apex{}-L, at both scales. Because the
majority keeps validation improving, the minimum falls at epoch $4$ to $18$
for every rule, so at the best-val checkpoint the unlearned minorities are
barely reproduced; after a fixed ten epochs the target reproduces $30$ to
$80\%$ of the minority's secrets given the dossier while the reference
reproduces none, and the probability ratio on the secret reaches up to
$70\times$.

\subsection{Mixtures}
\label{app:understanding-mixtures}

\begin{figure}[t]
  \centering
  \includegraphics[width=\linewidth]{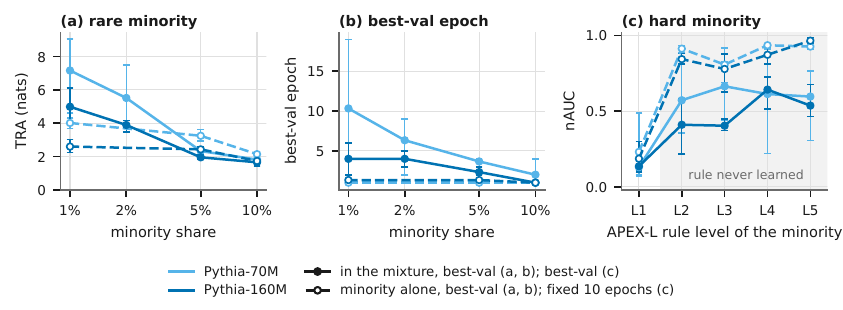}
  \caption{The rare-minority mixture per model scale; Pythia-70M (light) and
  160M (dark), three seeds, bars the seed range. (a)
  Figure~\ref{fig:understanding-regime}d not pooled: \tra{} on the
  minority's secrets in nats per token at the best-val checkpoint, inside the
  mixture (solid) and for the minority trained alone at the same count
  (dashed, hollow). (b) The best-val epoch of the same runs. (c) An
  \apex{}-L secret as the $1\%$ minority instead of an impossible one:
  $\nauc$ averaged over scores by rule level, at the best-val checkpoint
  (solid) and after a fixed ten epochs that give every level the same number
  of passes (dashed, hollow).}
  \label{fig:app-mixture-detail}
\end{figure}

\begin{figure}[p]
  \centering
  \includegraphics[width=\linewidth]{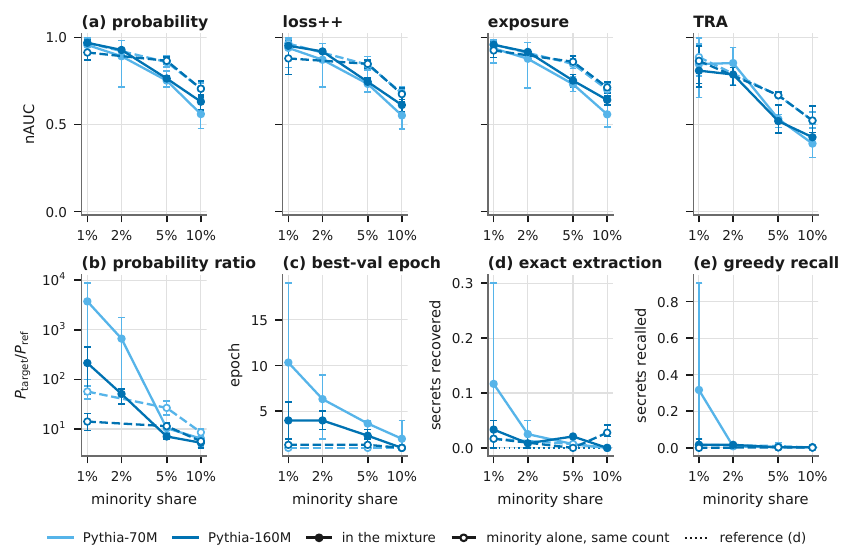}
  \caption{The rarity runs in full; Pythia-70M (light) and 160M (dark),
  three seeds, bars the seed range. (a) The four scores as $\nauc$. (b) The
  probability of the secret under the target relative to its reference.
  (c) The best-val epoch. (d) Exact extraction of the secret under full
  enumeration of the $10^4$ candidates (dotted: the reference). (e) Greedy
  recall of the secret. Solid, filled: the minority inside the mixture;
  dashed, hollow: the minority trained alone at the same count.}
  \label{fig:app-cleanmix-rarity}
\end{figure}

\begin{figure}[t]
  \centering
  \includegraphics[width=\linewidth]{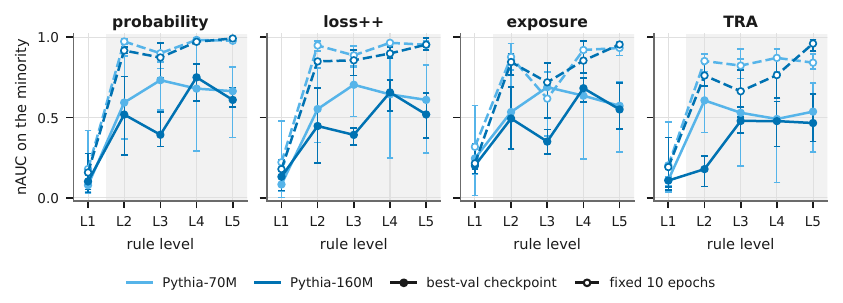}
  \caption{Panel~c of Figure~\ref{fig:app-mixture-detail} one score at a
  time: an \apex{}-L secret as the $1\%$ minority, $\nauc$ on the minority by
  rule level at the best-val checkpoint (solid) and after a fixed ten epochs
  (dashed, hollow); Pythia-70M (light) and 160M (dark), three seeds, bars the
  seed range.}
  \label{fig:app-mixture-scores}
\end{figure}

Figure~\ref{fig:app-mixture-detail} shows Figure~\ref{fig:understanding-regime}d
per model scale, together with the best-val epoch of the same runs and the
mixture with an \apex{}-L secret as the minority.
Figure~\ref{fig:app-cleanmix-rarity} gives the rarity runs under every score
and further extraction measures, and Figure~\ref{fig:app-mixture-scores}
shows panel~c of Figure~\ref{fig:app-mixture-detail} one score at a time. The
majority is the \apex{} extraction task; the minority is
\apex{}-FavNum, the same dossiers ending in a question whose answer is a
random four-digit number that appears in no input. The share runs over $1$,
$2$, $5$ and $10\%$ of $2000$ examples, and the minority is also trained
alone at $20$, $40$, $100$ and $200$ examples as a matched-count control
(Pythia-70M and 160M, three seeds, patience $3$).

The rank-based scores saturate: at matched minority count the mixture and
the pure control are indistinguishable under probability, loss++, \tra{} and
exposure, so the trend of those panels across the share is mostly the
data-size effect, which the pure controls reproduce. The majority's own
contribution shows only in the unbounded readouts: about ten epochs of
training instead of one at $1\%$ share, a probability ratio $66\times$ (70M)
and $15\times$ (160M) higher than the pure control, and exact extraction
seven and two times more frequent. The seed spread at $1\%$ is wide (70M:
best-val epoch $4$, $19$ and $8$; ratio $99\times$, $2459\times$ and
$8671\times$), so we quote orders of magnitude.

\begin{figure}[t]
  \centering
  \includegraphics[width=\linewidth]{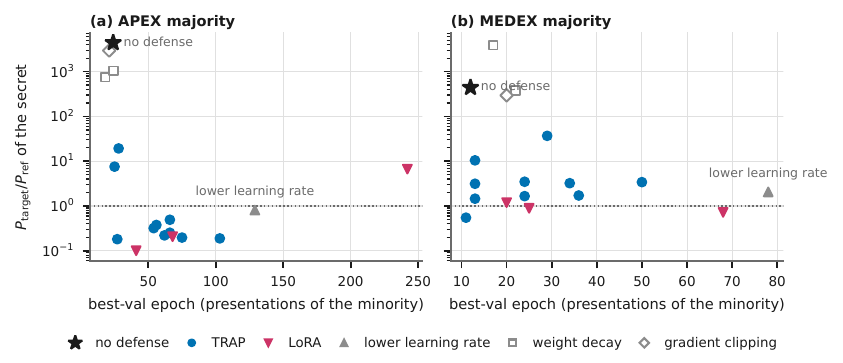}
  \caption{Minority memorization against the epoch selected by validation
  for every defense of the two cross-corpus mixtures at $1\%$ share
  (Pythia-160M, one seed), with the majority from (a) \apex{} and (b)
  \medex{}. The vertical axis is the probability the target gives the
  minority's secret relative to its reference; the dotted line is chance.
  Runs that train with small steps, through a lower learning rate or an
  adapter, run for many more epochs and stay near chance; the runs that
  memorize are the full-step ones. DP runs have no validation-selected epoch
  and are not shown, and cells that change the shared learning rate as a
  variant of another method are left out, as everywhere in the paper.}
  \label{fig:app-mixture-epochs}
\end{figure}

The cross-corpus mixtures place \medcase{} cases as a minority inside an
\apex{} or \medex{} majority (Pythia-160M, best-val, one seed). At $1$,
$4.8$ and $16.7\%$ share the minority's probability ratio is $4417\times$,
$61\times$ and $16\times$, against $5.1\times$ for the same $20$ examples
trained alone; the mixture trains to epoch $24$ where the pure minority
stops at $4$ to $8$. Swapping the roles so that the majority is the
unlearnable task stops training at epoch $3$ to $4$ and leaves the minority
at ratios of $0.92$ to $1.18\times$. The effect does not appear at Pythia-1B and
Llama-3.1-8B, where the majority is mastered within one or two epochs, and it does not
appear when the minority target is a long reasoning narrative rather than a
short label. Figure~\ref{fig:app-mixture-epochs} shows, within one of these
mixtures, that the number of presentations alone does not decide
memorization: the low-rate run trains for $129$ epochs and stays at chance
while the undefended run reaches $4417\times$ in $24$, the same step-size
effect as in Appendix~\ref{app:understanding-lr}.

\subsection{Learning rate}
\label{app:understanding-lr}

\begin{figure}[t]
  \centering
  \includegraphics[width=\linewidth]{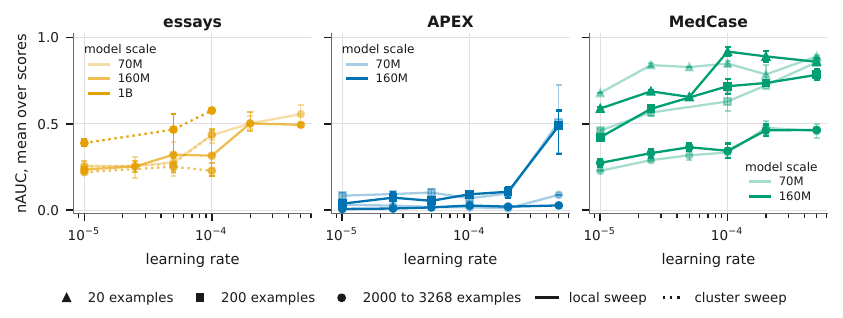}
  \caption{The learning-rate sweeps before averaging: $\nauc$ (mean over
  scores) at the best-val checkpoint, one line per model scale and
  training-set size (Pythia-70M to 1B, $20$ to $3268$ training examples),
  seed means with the seed range as bars; darker shades are larger models,
  markers distinguish training-set sizes, and dotted lines are the cluster
  sweep.}
  \label{fig:app-lr-by-series}
\end{figure}

\begin{figure}[t]
  \centering
  \includegraphics[width=\linewidth]{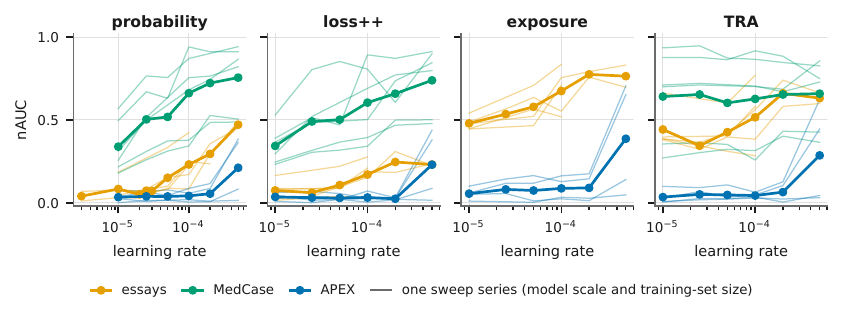}
  \caption{The learning-rate sweeps one score at a time, best-val
  checkpoint. Thin lines are the individual series (one model scale and
  training-set size, seed means), thick lines the mean over the series of a
  testbed at each rate.}
  \label{fig:app-lr-scores}
\end{figure}

The sweeps behind Figure~\ref{fig:understanding-regime}e are all stopped at
the best-val epoch and scored with all four scores: a cluster sweep of
Pythia-70M, 160M and 1B on the \essays{} at $10^{-5}$, $5\cdot10^{-5}$ and
$10^{-4}$ with three seeds (a wider range at 70M and 160M was scored with the
probability score only and enters only the probability panel of
Figure~\ref{fig:app-lr-scores}), and a local sweep over six rates from
$10^{-5}$ to $5\cdot10^{-4}$ covering the \essays{} at 70M and 160M (three
seeds), \apex{} at 70M and 160M with $200$ applicants (three seeds) and with all
$2198$ (one seed), and \medcase{} at 70M and 160M with $20$, $200$ and $2000$ cases (two
seeds). Figure~\ref{fig:app-lr-by-series} shows the series before
averaging, one line per model scale and training-set size, and
Figure~\ref{fig:app-lr-scores} shows them one score at a time; $14$ of the
$15$ series rise with the rate.

\begin{figure}[t]
  \centering
  \includegraphics[width=\linewidth]{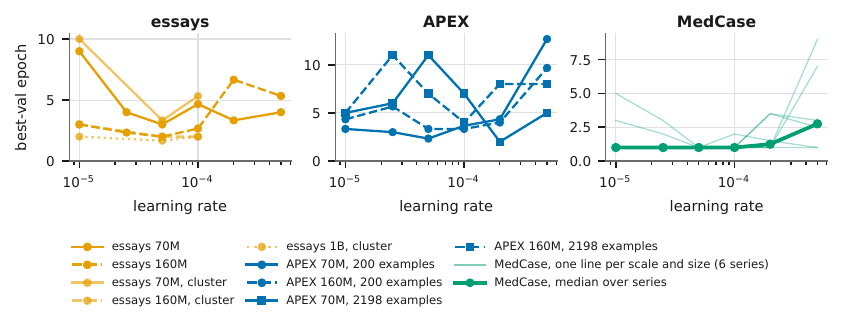}
  \caption{The best-val epoch of every learning-rate series. \Essays{}: the
  local sweep (solid 70M, dashed 160M) and the cluster sweep (pale). \apex{}:
  $200$ applicants (circles) and $2198$ (squares). \medcase{}: one line per
  scale and size, with the median.}
  \label{fig:app-lr-bestval-epoch}
\end{figure}

\begin{figure}[t]
  \centering
  \includegraphics[width=\linewidth]{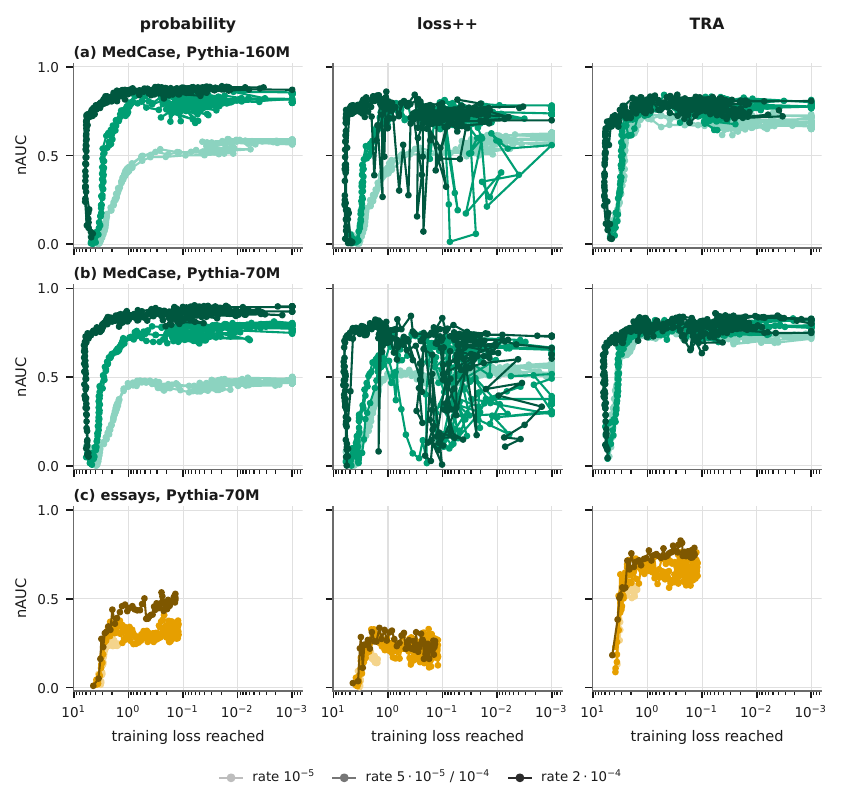}
  \caption{Learning rate at matched fit: the runs scored along training that
  vary only the rate, each score against the training loss the run had
  reached at that point. (a, b) \medcase{} at Pythia-160M and 70M, $200$
  cases, three seeds per rate drawn separately, scored from the first
  optimizer step against the same-step reference (the step-level runs of
  Figure~\ref{fig:app-trajectory-details}b). (c) The \essays{} at Pythia-70M
  on the full corpus, one seed at the smallest and largest rate and three at
  $10^{-4}$, drawn as separate lines; sixty epochs bring their training loss
  only to about a tenth. The middle rate is $5\cdot10^{-5}$ for \medcase{}
  and $10^{-4}$ for the \essays{}, and darker is a larger rate.}
  \label{fig:app-lr-fit}
\end{figure}

\paragraph{Is it just more training?}
A larger step covers more ground per epoch, so a higher rate could memorize
more simply because it has trained further by the time validation bottoms
out. In epochs it has not: the best-val epoch does not grow with the rate
(Figure~\ref{fig:app-lr-bestval-epoch}). On the \essays{} the smallest rate
trains longest, nine to ten epochs at 70M against three to five at the
higher rates, and still memorizes least; on \medcase{} with $200$ or more
cases validation bottoms out at epoch one at every rate up to $10^{-4}$. The
same number of larger steps is still more optimization, though, so
Figure~\ref{fig:app-lr-fit} compares the \medcase{} and \essays{} runs scored
along training at the same training loss rather than at the same epoch, so
that points at the same horizontal position have fit the training set
equally well. On \medcase{} the curves do not collapse under the probability
score at either scale: at every level of fit the smaller rate leaks less,
for every seed, and once every run has driven the training loss to zero the
smallest and the largest rate still differ by $0.3$ to $0.4$. Under \tra{}
the gap is smaller, about $0.1$ at 160M and half that at 70M, where the two
larger rates are not separated. The \essays{}, trained on their full corpus,
never bring the training loss below a tenth in sixty epochs, and the
smallest rate only reaches about $1.5$, so the comparison there covers the
early part of the axis; over the range the rates share, the smallest rate
sits lowest under every score, and the largest is about $0.1$ above the
middle one under the probability score and \tra{} and level with it under
loss++. Why a larger step should leave more behind at the same fit we have
not tested; a natural guess is that each presentation imprints more of an
example's specifics before the other examples' updates wash them out, while
the shared structure that drives validation loss is learned either way.

\paragraph{Which scores move.}
The size of the effect depends on the score. Averaged over series, the
probability score on the \essays{} goes from near chance at the smallest rate
to about $0.5$ at the largest and on \medcase{} from about $0.3$ to $0.8$,
and loss++ follows it; \tra{} starts higher and moves less, on the \essays{}
from about $0.45$ to $0.6$ and on \medcase{} flat near $0.65$, and exposure
behaves like \tra{} where it was scored. The difference is the reference.
The probability score is absolute: a larger step makes the model more
confident about everything it trained on, the spans included. \tra{}
compares that confidence with a reference trained at the same rate, so the
part of the rise that the rate gives every token cancels, and what remains
is the smaller part specific to the target's own records. How much a larger
learning rate ``memorizes'' therefore depends on which of the two one means,
and Figure~\ref{fig:app-lr-fit} shows both.

\begin{figure}[t]
  \centering
  \includegraphics[width=\linewidth]{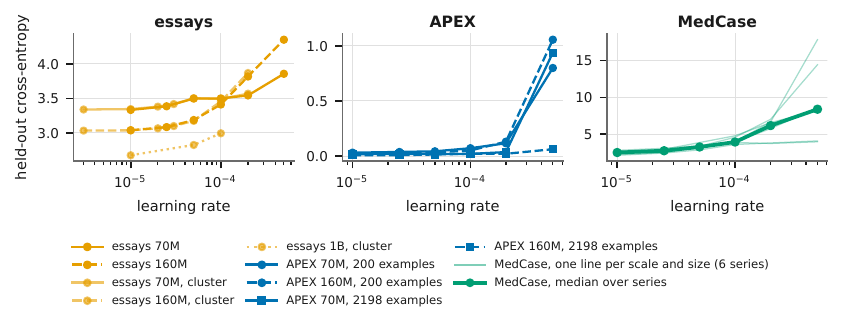}
  \caption{Held-out cross-entropy at the best-val checkpoint against the
  learning rate, same series and styles as
  Figure~\ref{fig:app-lr-bestval-epoch}; the cluster sweep reports test
  cross-entropy, the local sweeps validation cross-entropy.}
  \label{fig:app-lr-utility}
\end{figure}

\paragraph{The largest rate.}
Figure~\ref{fig:app-lr-utility} shows the held-out cross-entropy of the same
runs. It worsens with the rate everywhere, gently up to $2\cdot10^{-4}$ and
sharply at $5\cdot10^{-4}$, where \medcase{} reaches several nats and \apex{}
loses an order of magnitude.

\begin{figure}[t]
  \centering
  \includegraphics[width=\linewidth]{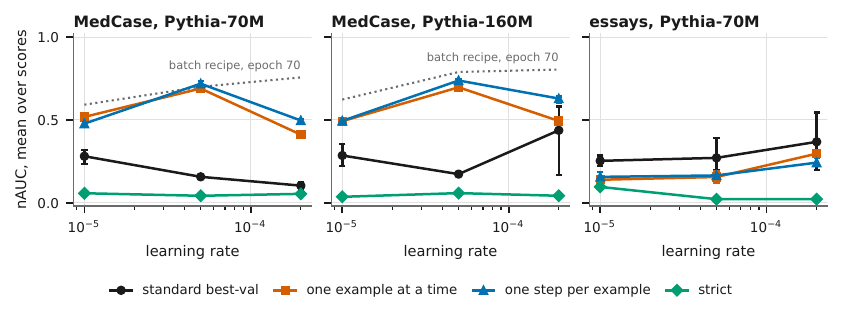}
  \caption{Training one example at a time. $\nauc$ averaged over the four
  scores against the learning rate, seed means with the range over three
  seeds, same tokenization, validation set and optimizer settings as the
  recipes, batch of one example, constant rate. ``One example at a time''
  keeps stepping on an example while validation improves; ``one step per
  example'' takes exactly one; ``strict'' also undoes a first step that does
  not improve validation. Black is the batch recipe at its best-val
  checkpoint at the same rate, and the dotted grey line the batch recipe
  after seventy epochs.}
  \label{fig:app-lr-sequential}
\end{figure}

\paragraph{One example at a time.}
Figure~\ref{fig:app-lr-sequential} tests the reading directly. The training
examples are presented one at a time in random order; on each example the
model takes one optimizer step and then keeps stepping while the validation
loss improves, undoing the first step that does not. Two controls bracket
this: exactly one step per example, and a strict version that also undoes
the first step when it does not help validation. The strict version skips
almost every example, since after the first few a step on a single example
almost never lowers the validation loss, so the model barely leaves the base
model. Once one step per example is forced, the extra steps change nothing,
and that single pass memorizes as much on \medcase{} as seventy epochs of the
batch recipe at the same rate, while its validation loss ends worse. A batch
step spreads one update over sixteen \medcase{} cases (eight essays) and the example-sized step
gives the whole update to one, so this is the batch recipe at a sixteen-fold
larger per-example rate, and memorization follows that per-example push. On
the \essays{}, where an example is a long document with a few sensitive
tokens in it, one example-sized step imprints little and the batch recipe's
several epochs memorize more; within the sequential regime memorization
still rises with the rate. At the largest rate the example-sized steps break
the model before it memorizes more.

\subsection{Model size}
\label{app:understanding-scale}

\begin{figure}[t]
  \centering
  \includegraphics[width=\linewidth]{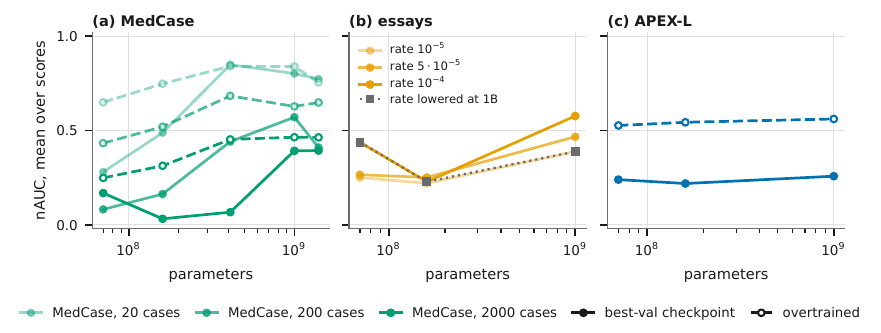}
  \caption{Memorization against model size with the recipe held fixed,
  $\nauc$ averaged over scores. (a) \medcase{} at five Pythia sizes, one line
  per training-set size, at the best-val checkpoint (solid) and overtrained
  (dashed); one seed per configuration. (b) \Essays{} at best-val at three
  sizes, one line per learning rate applied to all three (seed means from
  the cluster sweep); the dotted grey line is the scale-adapted series, which
  trains 70M and 160M at $10^{-4}$ and 1B at $10^{-5}$. (c) \apex{}-L at
  three sizes, averaged over the five rule levels, at best-val and at the
  last scored epoch.}
  \label{fig:app-scale}
\end{figure}

\begin{figure}[t]
  \centering
  \includegraphics[width=\linewidth]{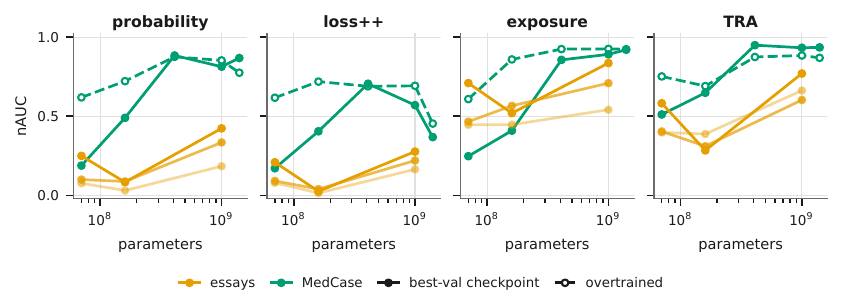}
  \caption{Panel h of Figure~\ref{fig:understanding-regime} one score at a
  time: \medcase{} with $20$ cases at the best-val checkpoint (solid) and
  overtrained (dashed), and the \essays{} at best-val at three fixed learning
  rates (darker is a larger rate).}
  \label{fig:app-scale-scores}
\end{figure}

Figure~\ref{fig:app-scale} separates the three scale series behind the model
size paragraph and Figure~\ref{fig:understanding-regime}h, and
Figure~\ref{fig:app-scale-scores} shows that panel one score at a time. All
three series hold the recipe fixed across sizes; only the model changes.

On \medcase{} the rise with size is largest on the smallest training set:
with $20$ cases the best-val $\nauc$ goes from about $0.3$ at 70M to about
$0.8$ from 410M on, and the overtrained arm from $0.65$ to about $0.8$ (one
seed per cell on twenty spans, so the dip at 160M sits inside the chance
band and should not be read as a feature); with $200$ cases both arms rise
more gently, and with $2000$ cases both stay low up to 410M and the best-val
checkpoint rises only at 1B and 1.4B. The one score that does not follow
this is loss++ on the largest models at $20$ cases, which falls at 1B and
1.4B while probability, exposure and \tra{} stay high
(Figure~\ref{fig:app-scale-scores}). The fall is the ceiling of loss++
described in Appendix~\ref{app:metric-forms}, not less memorization: on
these runs $0.7\%$ of held-out spans outrank the median member by
probability at 1B and $1.1\%$ at 1.4B, against $9.4\%$ and $23.4\%$ by
loss++.

On the \essays{} the ordering depends on the learning rate.
At each rate used for all three sizes ($10^{-5}$, $5\cdot10^{-5}$ and
$10^{-4}$) the 1B model memorizes most, and the 160M model is usually the
lowest point because it reaches its best-val epoch within one or two epochs
at every rate. The scale-adapted series, which trains 70M and 160M at
$10^{-4}$ but 1B at $10^{-5}$, has the 1B model below the 70M one; this is
the comparison under which larger models appear to memorize less, and it
compares rates, not sizes.

On \apex{}-L, where the target is a rule of controlled hardness, the three
sizes are close: at best-val they all sit near $0.25$, averaged over the
five levels, and at the last scored epoch near $0.55$, with 1B slightly
above the others. The size effect is small here because the learnable level
is learned at every size and the unlearnable levels are memorized at every
size; what changes with size is mostly how fast
(Appendix~\ref{app:understanding-hardness}).

\FloatBarrier

\section{Full Mitigation Frontiers}
\label{app:mitigation-frontiers}

This appendix shows the frontiers behind Figure~\ref{fig:mitigation} with
every method drawn on its own, one score at a time, and with the directional
AUC in place of $\nauc$. All figures use the same exported runs as the main
figure. A point is one setting of a method (the seed mean where there are
several seeds), a method's line is its lower-left frontier, and the faint
points are settings that another setting of the same method beats on both
axes. Hollow markers and dashed lines are the same method trained on
$\Dtgt\cup\Dref$, which sees twice the data and is a control for the effect
of more data rather than a matched defense. The star is the undefended
model, the hollow star the undefended recipe trained on the union, and the
green dot and dotted line the untrained model. On the \essays{}, settings whose held-out cross-entropy is worse than the
untrained model's are left out; elsewhere they are counted as off scale.
Methods that read the annotations of which tokens are sensitive are left
out everywhere.

\subsection{Every method on its own}

\begin{figure}[p]
  \centering
  \includegraphics[width=\linewidth]{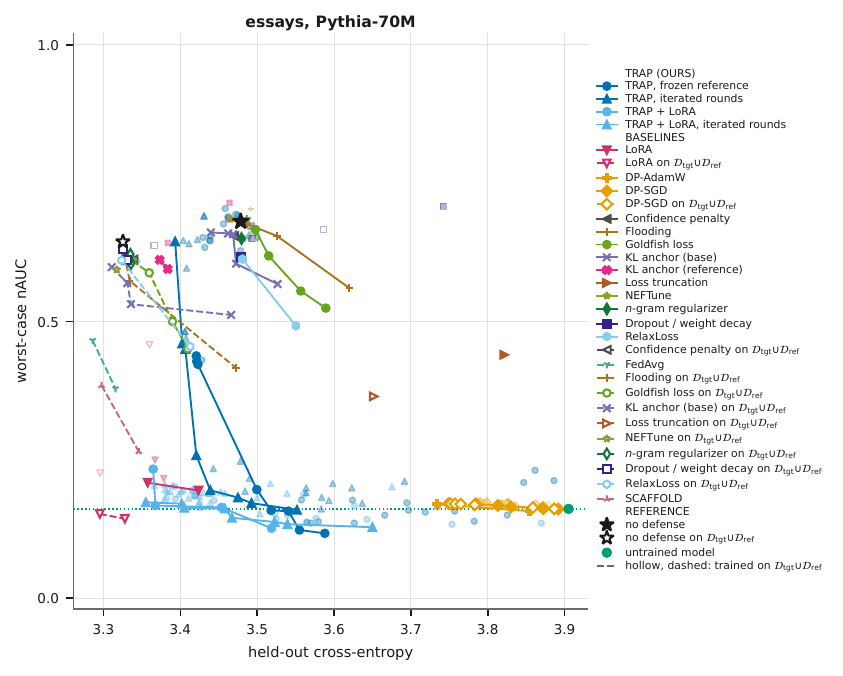}
  \caption{\Essays{} at Pythia-70M, every method on its own, best-val
  checkpoint. Lower-left is better.}
  \label{fig:app-frontier-essays-70m}
\end{figure}

\begin{figure}[p]
  \centering
  \includegraphics[width=\linewidth]{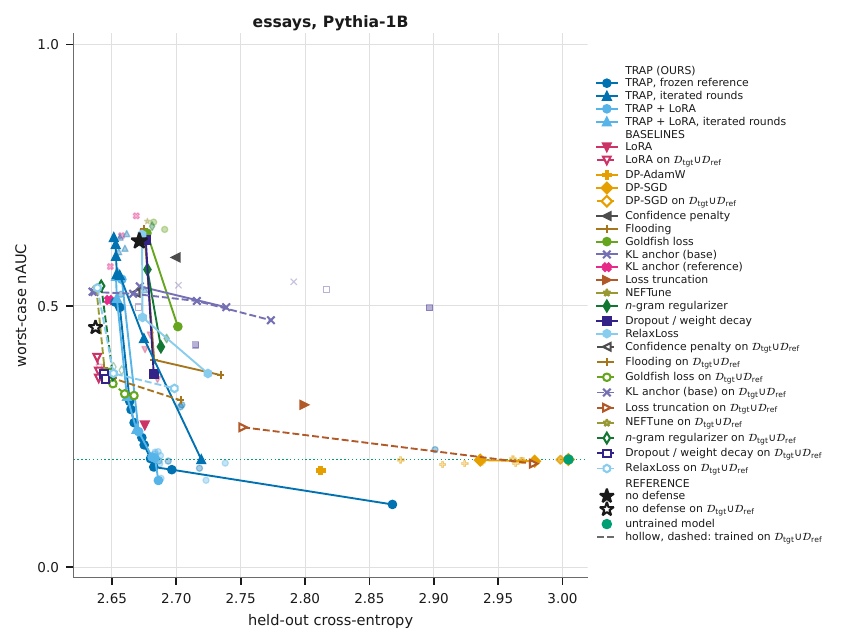}
  \caption{\Essays{} at Pythia-1B, every method on its own, best-val
  checkpoint.}
  \label{fig:app-frontier-essays-1b}
\end{figure}

\begin{figure}[p]
  \centering
  \includegraphics[width=\linewidth]{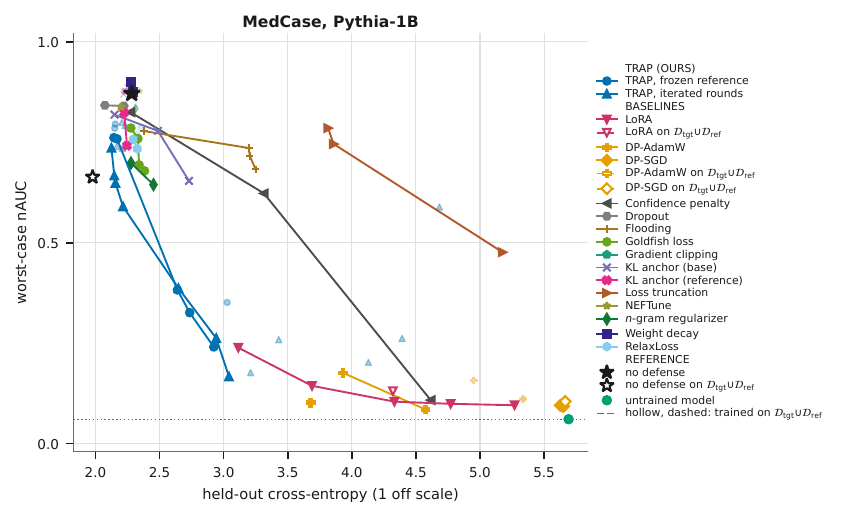}
  \caption{\medcase{} at Pythia-1B, every method on its own, best-val
  checkpoint, seed means over three seeds; twenty member spans, so
  differences near the floor are directions rather than rankings. One cell
  is off scale, DP-AdamW at $\varepsilon=1$, at $5.9$ nats against an
  undefended $2.28$.}
  \label{fig:app-frontier-medcase-1b}
\end{figure}

\begin{figure}[p]
  \centering
  \includegraphics[width=\linewidth]{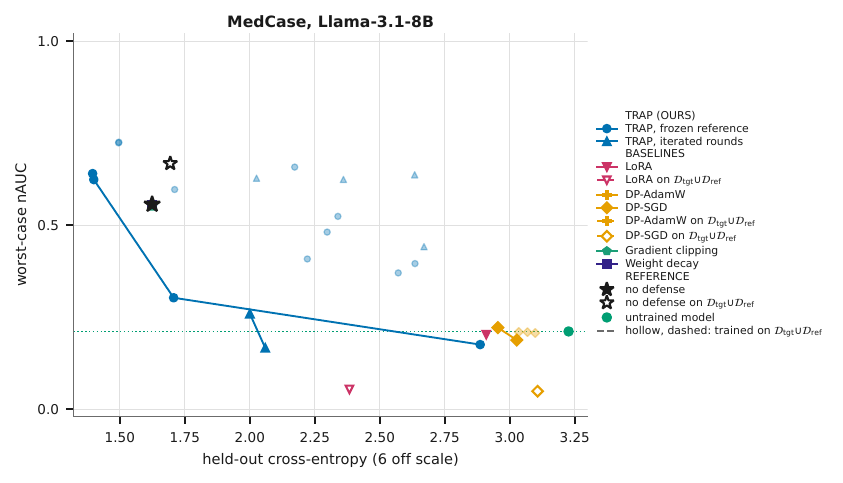}
  \caption{\medcase{} at Llama-3.1-8B-Instruct, every method on its own,
  best-val checkpoint. The six cells off scale, beyond $2.5\times$ the
  undefended cross-entropy of $1.62$ nats, are all DP-AdamW: $\varepsilon=16$
  at $9.7$, $8$ at $11.9$, $2$ at $13.3$, $4$ at $13.5$, $1$ at $14.1$ nats,
  and the $\varepsilon=8$ twin trained on $\Dtgt\cup\Dref$ at $9.1$.}
  \label{fig:app-frontier-medcase-llama8b}
\end{figure}

\begin{figure}[p]
  \centering
  \includegraphics[width=\linewidth]{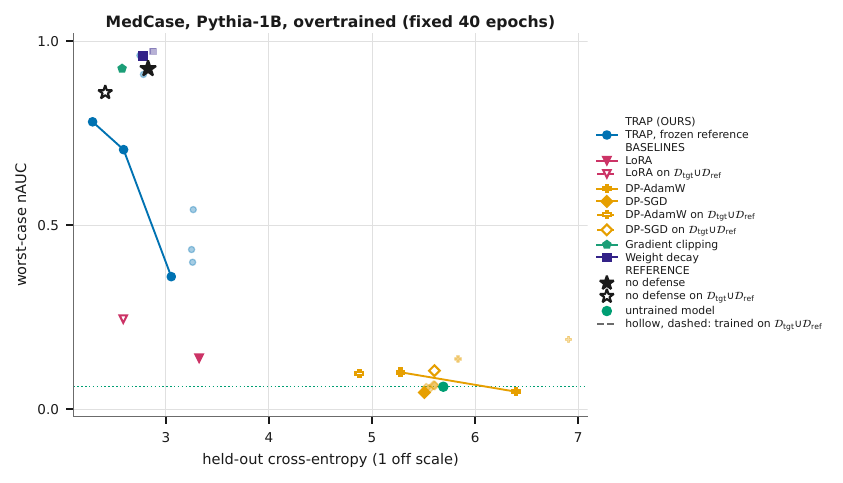}
  \caption{\medcase{} at Pythia-1B trained for a fixed forty epochs instead
  of to the best-val epoch. One cell is off scale, DP-AdamW at
  $\varepsilon=1$, at $8.3$ nats against an undefended $2.83$.}
  \label{fig:app-frontier-medcase-1b-over}
\end{figure}

\begin{figure}[p]
  \centering
  \includegraphics[width=\linewidth]{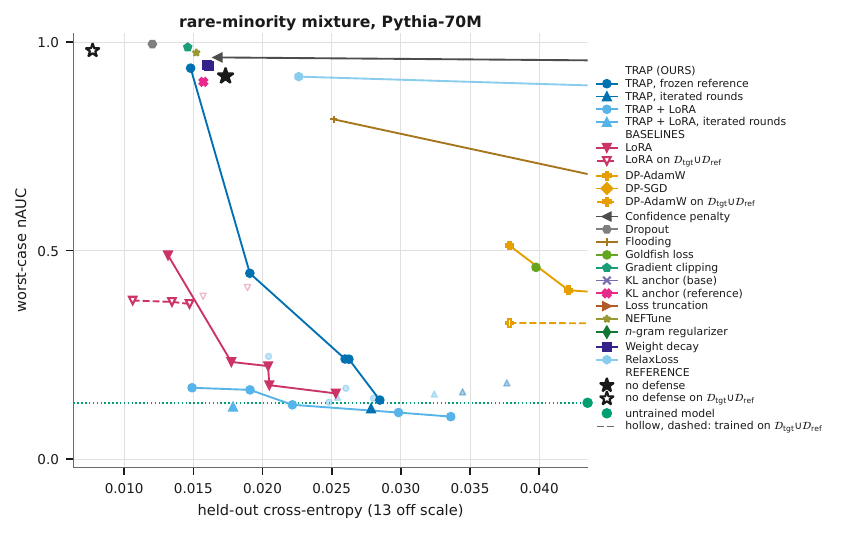}
  \caption{The rare-minority mixture at Pythia-70M, every method on its
  own, best-val checkpoint; seed means over three seeds, twenty minority
  secrets as the sensitive spans, so differences near the floor are
  directions rather than rankings. The undefended cross-entropy is $0.017$
  nats, so the $2.5\times$ window leaves thirteen cells off scale: flooding
  at level $0.03$ ($0.07$ nats), DP-AdamW at $\varepsilon=16$ on a lot of
  $1024$ ($0.12$), KL anchor to the base model at $\lambda=0.5$ ($0.15$),
  the $n$-gram regularizer at $\lambda=2$ ($0.23$), DP-AdamW at
  $\varepsilon=64$ ($0.24$) and at $\varepsilon=8$ on a lot of $1024$
  ($0.32$), RelaxLoss at $\alpha=0.03$ ($0.43$), DP-SGD at $\varepsilon=8$
  ($0.56$), DP-AdamW at $\varepsilon=8$ on $\Dtgt\cup\Dref$ ($0.94$) and on $\Dtgt$ ($1.15$), the confidence penalty at $\beta=0.5$ ($1.59$), loss
  truncation at $0.2$ ($3.39$), and DP-AdamW at $\varepsilon=1$ ($10.0$).}
  \label{fig:app-frontier-mixture-70m}
\end{figure}

\begin{figure}[p]
  \centering
  \includegraphics[width=\linewidth]{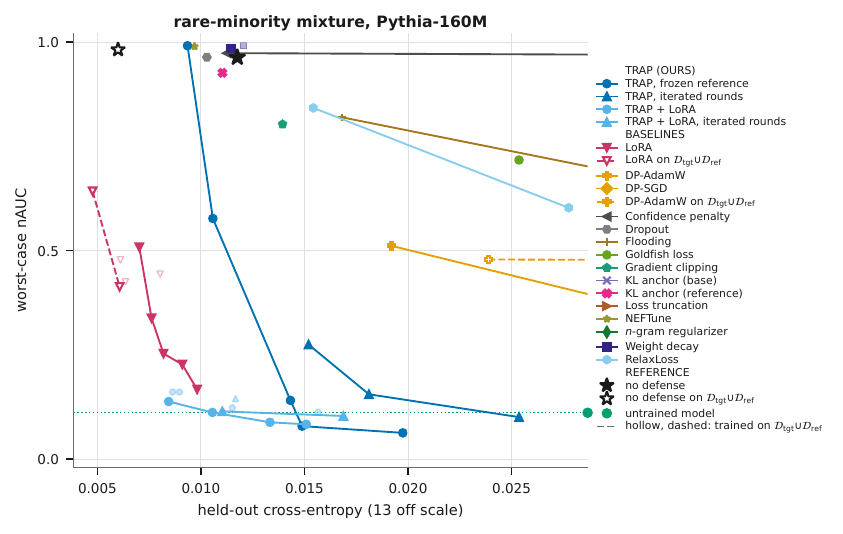}
  \caption{The rare-minority mixture at Pythia-160M, every method on its
  own, best-val checkpoint; seed means over three seeds. The undefended
  cross-entropy is $0.012$ nats; the thirteen cells off scale are flooding
  at level $0.03$ ($0.03$ nats), DP-AdamW at $\varepsilon=64$ on a lot of
  $1024$ ($0.05$), KL anchor to the base model at $\lambda=0.5$ ($0.07$),
  the $n$-gram regularizer at $\lambda=2$ ($0.15$), DP-AdamW at
  $\varepsilon=64$ ($0.23$), DP-SGD at $\varepsilon=8$ ($0.30$), DP-AdamW at
  $\varepsilon=16$ and $8$ on a lot of $1024$ ($0.37$ and $1.36$), DP-AdamW
  at $\varepsilon=8$ on $\Dtgt\cup\Dref$ ($1.07$) and on $\Dtgt$ ($1.46$), the confidence penalty at $\beta=0.5$ ($1.53$), loss truncation
  at $0.2$ ($4.06$), and DP-AdamW at $\varepsilon=1$ ($13.7$).}
  \label{fig:app-frontier-mixture-160m}
\end{figure}

Figures~\ref{fig:app-frontier-essays-70m} to~\ref{fig:app-frontier-mixture-160m}
split the families of Figure~\ref{fig:mitigation} into methods and add four
campaigns the main text does not use: the \essays{} and the mixture at
Pythia-70M, \medcase{} at Llama-3.1-8B-Instruct, and \medcase{} at Pythia-1B
trained for a fixed forty epochs. Across the \trap{} variants the frozen
and iterated references behave alike, and the adapter variants sit lowest on
the \essays{}.

The 70M \essays{} campaign has $1281$ runs over $464$ settings, $409$ of
them averaged over several optimizer seeds; the 1B campaign has $264$ runs
over $206$ settings, $32$ of them with multiple seeds. Both score $772$
member and $48$ held-out spans, so the 70M frontier supports comparisons
between replicated method families while the 1B frontier supports only
larger directional differences. The two federated baselines, run at 70M only, use
$\Dtgt$ and $\Dref$ as their two clients and are drawn hollow like the other
union-trained models; at the same cross-entropy both memorize more than LoRA
trained on the union (worst-case $\nauc$ about $0.15$), SCAFFOLD at $0.26$
to $0.38$ and FedAvg at $0.38$ to $0.47$.

At Pythia-1B every one of the $81$ trained \medcase{} settings has three
seeds; the $41$ settings at Llama-3.1-8B are single-seed. Both score $20$
member diagnoses against $897$ held-out ones. With twenty members the chance band of $\nauc$ extends to
about $0.26$, so defended models near the floor cannot be ranked against
each other, and held-out cross-entropy alone does not establish that a
configuration still produces useful diagnoses. At Llama-3.1-8B \trap{} again
descends first, to about half the undefended leakage within a tenth of a
nat; the iterated \trap{} rounds reach about $0.2$ at $0.4$ nats above the
undefended cross-entropy, and LoRA reaches the same level only $1.3$ nats
above it, beside DP-SGD; DP-AdamW reaches the floor only near the
untrained cross-entropy, and the regularizers stay near the undefended
model. Figure~\ref{fig:app-frontier-medcase-1b-over} repeats \medcase{} at
Pythia-1B, single-seed, trained for a fixed forty epochs instead of to its best-val
epoch: \trap{} still moves the frontier, but from a higher starting point,
so checkpoint selection and the penalty are complementary.

Figures~\ref{fig:app-frontier-mixture-70m}
and~\ref{fig:app-frontier-mixture-160m} do the same for the rare-minority
mixture of Section~\ref{sec:factors}: \apex{}-FavNum secrets at a $1\%$
share inside ordinary \apex{}, $2000$ examples per half with twenty minority
records, trained with the recipe of Appendix~\ref{app:understanding-mixtures}
at Pythia-70M and 160M, every setting at three seeds with the data split
and the training seed varying together. The sensitive spans are the twenty
minority secrets, scored against $200$ held-out applicants, so the chance
band of $\nauc$ extends to about $0.27$; the untrained model scores $0.11$
at 160M and $0.13$ at 70M, and models near those values read as chance. The
grid is larger than on the \essays{}: \trap{} with its frozen reference at
four weights and two thresholds, iterated rounds at weights $2$ and $4$,
\trap{} through LoRA adapters of rank $4$, $8$ and $64$ at weights $1$, $2$
and $4$ (iterated rounds at three of these settings), LoRA at ranks $1$,
$2$, $4$, $8$ and $64$ with their union twins, DP-AdamW at three budgets
on the DP trainer's default lot and at four budgets on a lot of $1024$, each
with a union twin, DP-SGD, weight decay, gradient clipping, dropout, and the
nine label-free objectives. Ranks $1$ and $2$ were given an $80$-epoch
ceiling with the same early stopping, because rank $4$ already ran to the
recipe's $40$ epochs at 70M; no other run reached the ceiling. The frozen
reference gives the steepest frontier of the full-model variants at both
scales, and the iterated rounds improve on it only at 70M and weight $2$,
unlike on \medcase{}. At 160M, \trap{} through an adapter at weight $1$ or
$2$ sits at the floor at every rank, between $0.11$ and $0.16$, at a
cross-entropy at or below the undefended model's, because the adapter
improves the mixture's held-out loss on its own. LoRA alone leaks less as
its rank falls, from $0.51$ at rank $64$ to $0.17$ at rank $1$, while its
cross-entropy rises toward the undefended model's; at every rank both were
run at, adding the penalty to the adapter lowers the leakage for about a
thousandth of a nat, most at rank $64$, from $0.51$ to $0.14$. At 70M the
ordering is the same, with the rank-$64$ adapter at weight $1$ below the
undefended cross-entropy. The objectives and regularizers stay near the
undefended model unless they wreck utility, and the union twins leak about
as much as their matched models on every score but \tra{}, which their
reference cannot measure. DP needs its lot tuned before it is on the panel
at all: at the DP trainer's default lot of $256$ examples the best budget
costs $14$ to $20$ times the undefended cross-entropy, while a lot of $1024$
brings $\varepsilon{=}256$ to $1.6$ times at 160M and $2.2$ times at 70M
(Appendix~\ref{app:frontier-settings} lists both). Even tuned, DP buys
utility with privacy on this testbed: the budgets that fit on the panel leak
$0.41$ to $0.51$, as much as a rank-$64$ adapter, and the budgets that bring
leakage to $0.14$ to $0.18$ cost four to eighteen times the undefended
cross-entropy, against about $1.7$ times for \trap{} at weight $4$.

\subsection{The settings behind the frontiers}
\label{app:frontier-settings}

Tables~\ref{tab:app-settings-essays-70m} to~\ref{tab:app-settings-mixture-160m}
list, for every method drawn in Figures~\ref{fig:app-frontier-essays-70m}
to~\ref{fig:app-frontier-mixture-160m} and in Figure~\ref{fig:mitigation},
the settings on its lower-left frontier: the value of the method's own
hyperparameters, followed in gray by the point's held-out cross-entropy and worst-case
$\nauc$. A row together with the
testbed's shared recipe, given in each caption, reproduces the point;
settings that another setting of the same method beats on both axes are not
listed. Every model, defended or not, trains under the shared recipe of its
testbed (DP-SGD and DP-AdamW under their own optimizer's) and stops at its
own best-validation epoch under the same early-stopping rule (except the
overtrained campaign of Table~\ref{tab:app-settings-medcase-1b-over},
trained for a fixed forty epochs), so its utility is the best that setting
reaches. Within a method only its own hyperparameters
vary: the weight for \trap{}, the rank for LoRA, the budget for DP, the
level or fraction of an objective. The grids differ in size, since \trap{}'s
own variants are part of the study, and a frontier can only move outward
with more settings, so the comparison rests on where the sweeps go rather
than on how many points they hold: each hyperparameter runs from a value with little or no
effect up to the strongest value in its published range, on \medcase{} and
the mixture up to the point where utility collapses, and at every strength
we ran, an objective or regularizer either keeps the undefended memorization
or gives up utility before memorization moves. Where seeds are replicated
the frontiers pass through seed means, which removes the advantage a larger
grid gains from seed noise. \trap{}'s threshold $\epsilon$ is zero except
for a few cells, given in the tables. Round $1$ of an iterated pipeline is the target trained against the
ordinary frozen reference, round $r$ is trained against the defended target
of round $r-1$, and ``on $\Dtgt\cup\Dref$'' marks the union-trained twins.

\begin{table}[p]
  \caption{\Essays{} at Pythia-70M: the settings on each method's lower-left frontier in
  Figure~\ref{fig:app-frontier-essays-70m}, with the point's held-out
  cross-entropy and worst-case $\nauc$ in gray. Every run
  uses the uniform recipe of Appendix~\ref{app:datasets}: learning rate $10^{-4}$, constant schedule with $200$ warmup steps, $8$ documents per step, block size $2048$, fp32, no weight decay, patience $4$ on the validation split, a $60$-epoch ceiling; DP runs clip at $0.1$ with $\delta{=}10^{-5}$ and, unless the setting says otherwise, the same lot and learning rate; only the listed hyperparameter varies. $w$ and $\epsilon$ are the \trap{} weight and
  threshold, $r$ the adapter rank, $\varepsilon$ under DP the privacy budget.}
  \label{tab:app-settings-essays-70m}
  \centering\footnotesize
  \begin{tabular}{@{}l>{\raggedright\arraybackslash}p{0.61\linewidth}@{}}
    \toprule
    Method & Settings on the frontier (cross-entropy, worst-case $\nauc$) \\
    \midrule
\trap{}, frozen reference & $w{=}1,\ \epsilon{=}0.05$ ({\color{gray}3.42, 0.44}); $w{=}1,\ \epsilon{=}0.02$ ({\color{gray}3.42, 0.43}); $w{=}1,\ \epsilon{=}0$ ({\color{gray}3.42, 0.42}); $w{=}4,\ \epsilon{=}1$ ({\color{gray}3.50, 0.20}); $w{=}2,\ \epsilon{=}0.2$ ({\color{gray}3.52, 0.16}); $w{=}2,\ \epsilon{=}0.1$ ({\color{gray}3.54, 0.16}); $w{=}2,\ \epsilon{=}0.05$ ({\color{gray}3.55, 0.12}); $w{=}2,\ \epsilon{=}-0.05$ ({\color{gray}3.59, 0.12}) \\
\trap{}, iterated rounds & $w{=}0.9,\ \epsilon{=}0$, round 2 (first round alternating) ({\color{gray}3.39, 0.64}); $w{=}1,\ \epsilon{=}0.02$, round 2 ({\color{gray}3.40, 0.46}); $w{=}1,\ \epsilon{=}0.05$, round 2 ({\color{gray}3.41, 0.45}); $w{=}1.25,\ \epsilon{=}0$, round 3 ({\color{gray}3.42, 0.26}); $w{=}1.5,\ \epsilon{=}0$, round 3 (first round alternating) ({\color{gray}3.44, 0.19}); $w{=}2,\ \epsilon{=}0$, round 3 (first round alternating) ({\color{gray}3.47, 0.18}); $w{=}2,\ \epsilon{=}-0.05$, round 3 (first round alternating) ({\color{gray}3.49, 0.17}); $w{=}3,\ \epsilon{=}0$, round 2 ({\color{gray}3.55, 0.16}) \\
\trap{} + LoRA & $r{=}16,\ w{=}0.5,\ \epsilon{=}0$ ({\color{gray}3.36, 0.23}); $r{=}64,\ w{=}1,\ \epsilon{=}0$ ({\color{gray}3.37, 0.17}); $r{=}4,\ w{=}1,\ \epsilon{=}0.02$ ({\color{gray}3.45, 0.16}); $r{=}128,\ w{=}2,\ \epsilon{=}0$ ({\color{gray}3.52, 0.13}) \\
\trap{} + LoRA, iterated rounds & $r{=}16,\ w{=}0.9,\ \epsilon{=}0$, round 2 ({\color{gray}3.36, 0.17}); $r{=}4,\ w{=}0.5,\ \epsilon{=}0$, round 2 ({\color{gray}3.40, 0.17}); $r{=}8,\ w{=}0.9,\ \epsilon{=}0$, round 1 ({\color{gray}3.41, 0.16}); $r{=}4,\ w{=}1.1,\ \epsilon{=}0$, round 1 ({\color{gray}3.46, 0.16}); $r{=}16,\ w{=}1.5,\ \epsilon{=}0$, round 1 ({\color{gray}3.47, 0.15}); $r{=}64,\ w{=}2,\ \epsilon{=}0$, round 1 ({\color{gray}3.52, 0.14}); $r{=}16,\ w{=}2,\ \epsilon{=}0$, round 1 ({\color{gray}3.54, 0.13}); $r{=}64,\ w{=}3,\ \epsilon{=}0$, round 1 ({\color{gray}3.65, 0.13}) \\
LoRA & $r{=}64$ ({\color{gray}3.36, 0.21}); $r{=}4$ ({\color{gray}3.42, 0.19}) \\
LoRA on $\Dtgt\cup\Dref$ & $r{=}64$ ({\color{gray}3.30, 0.15}); $r{=}16$ ({\color{gray}3.33, 0.14}) \\
DP-AdamW & $\varepsilon{=}8$ ($L{=}1024$, $T{=}10$, lr $10^{-4}$) ({\color{gray}3.73, 0.17}); $\varepsilon{=}8$ ($L{=}1024$, $T{=}60$, lr $10^{-3}$) ({\color{gray}3.85, 0.16}) \\
DP-SGD & $\varepsilon{=}64$ ($T{=}120$) ({\color{gray}3.78, 0.17}); $\varepsilon{=}64$ ({\color{gray}3.81, 0.17}); $\varepsilon{=}64$ ($L{=}128$, $T{=}60$, lr $10^{-3}$) ({\color{gray}3.83, 0.16}); $\varepsilon{=}8$ ($L{=}128$, $T{=}10$, lr $10^{-3}$) ({\color{gray}3.87, 0.16}); $\varepsilon{=}1$ ($L{=}128$, $T{=}10$, lr $10^{-3}$) ({\color{gray}3.87, 0.16}); $\varepsilon{=}1$ ($L{=}1024$, $T{=}60$, lr $10^{-3}$) ({\color{gray}3.87, 0.16}); $\varepsilon{=}64$ ($L{=}1024$, $T{=}10$, lr $10^{-3}$) ({\color{gray}3.89, 0.16}) \\
DP-SGD on $\Dtgt\cup\Dref$ & $\varepsilon{=}64$ ($T{=}120$) ({\color{gray}3.75, 0.17}); $\varepsilon{=}16$ ($T{=}120$) ({\color{gray}3.75, 0.17}); $\varepsilon{=}4$ ($T{=}120$) ({\color{gray}3.76, 0.17}); $\varepsilon{=}2$ ($T{=}120$) ({\color{gray}3.76, 0.17}); $\varepsilon{=}64$ ({\color{gray}3.78, 0.17}); $\varepsilon{=}8$ ($L{=}128$, $T{=}10$, lr $10^{-3}$) ({\color{gray}3.86, 0.16}); $\varepsilon{=}8$ ($L{=}1024$, $T{=}10$, lr $10^{-3}$) ({\color{gray}3.89, 0.16}) \\
Goldfish loss & $k{=}32$ ({\color{gray}3.50, 0.67}); $k{=}8$ ({\color{gray}3.51, 0.62}); $k{=}4$ ({\color{gray}3.56, 0.56}); $k{=}3$ ({\color{gray}3.59, 0.52}) \\
Goldfish loss on $\Dtgt\cup\Dref$ & $k{=}32$ ({\color{gray}3.34, 0.61}); $k{=}8$ ({\color{gray}3.36, 0.59}); $k{=}4$ ({\color{gray}3.39, 0.50}); $k{=}3$ ({\color{gray}3.41, 0.45}) \\
Flooding & $b{=}2$ ({\color{gray}3.49, 0.68}); $b{=}2.5$ ({\color{gray}3.53, 0.65}); $b{=}3$ ({\color{gray}3.62, 0.56}) \\
Flooding on $\Dtgt\cup\Dref$ & $b{=}1.5$ ({\color{gray}3.33, 0.62}); $b{=}2$ ({\color{gray}3.33, 0.57}); $b{=}3$ ({\color{gray}3.47, 0.42}) \\
Confidence penalty & $\beta{=}0.1$ ({\color{gray}3.47, 0.65}) \\
Confidence penalty on $\Dtgt\cup\Dref$ & $\beta{=}0.1$ ({\color{gray}3.34, 0.61}) \\
Loss truncation & $c{=}0.1$ ({\color{gray}3.82, 0.44}) \\
Loss truncation on $\Dtgt\cup\Dref$ & $c{=}0.1$ ({\color{gray}3.65, 0.36}) \\
RelaxLoss & $\alpha{=}2.5$ ({\color{gray}3.48, 0.61}); $\alpha{=}3$ ({\color{gray}3.55, 0.49}) \\
RelaxLoss on $\Dtgt\cup\Dref$ & $\alpha{=}2$ ({\color{gray}3.32, 0.61}); $\alpha{=}3$ ({\color{gray}3.41, 0.45}) \\
KL anchor (base) & $\lambda{=}0.5$ ({\color{gray}3.44, 0.66}); $\lambda{=}0.1$ ({\color{gray}3.46, 0.66}); $\lambda{=}0.05$ ({\color{gray}3.47, 0.66}); $\lambda{=}1$ ({\color{gray}3.47, 0.60}); $\lambda{=}2$ ({\color{gray}3.53, 0.57}) \\
KL anchor (base) on $\Dtgt\cup\Dref$ & $\lambda{=}0.1$ ({\color{gray}3.31, 0.60}); $\lambda{=}0.05$ ({\color{gray}3.33, 0.57}); $\lambda{=}0.5$ ({\color{gray}3.34, 0.53}); $\lambda{=}2$ ({\color{gray}3.47, 0.51}) \\
KL anchor (reference) & $\lambda{=}1$ ({\color{gray}3.37, 0.61}); $\lambda{=}2$ ({\color{gray}3.38, 0.60}) \\
$n$-gram regularizer & $\lambda{=}0.5,\ \tau{=}0$ ({\color{gray}3.48, 0.65}) \\
$n$-gram regularizer on $\Dtgt\cup\Dref$ & $\lambda{=}2,\ \tau{=}0.05$ ({\color{gray}3.34, 0.62}); $\lambda{=}0.5,\ \tau{=}0$ ({\color{gray}3.34, 0.61}); $\lambda{=}2,\ \tau{=}0$ ({\color{gray}3.34, 0.60}) \\
NEFTune & $\alpha{=}10$ ({\color{gray}3.47, 0.69}); $\alpha{=}5$ ({\color{gray}3.49, 0.68}) \\
NEFTune on $\Dtgt\cup\Dref$ & $\alpha{=}10$ ({\color{gray}3.32, 0.59}) \\
Dropout / weight decay & dropout $p{=}0.1$ ({\color{gray}3.48, 0.62}) \\
Dropout / weight decay on $\Dtgt\cup\Dref$ & weight decay $0.1$ ({\color{gray}3.33, 0.63}); weight decay $0.01$ ({\color{gray}3.33, 0.61}) \\
FedAvg & $E{=}1$ local epochs ({\color{gray}3.29, 0.47}); $E{=}0.5$ local epochs ({\color{gray}3.32, 0.38}) \\
SCAFFOLD & $E{=}1$ local epochs ({\color{gray}3.30, 0.38}); $E{=}0.5$ local epochs ({\color{gray}3.35, 0.26}) \\
    \bottomrule
  \end{tabular}
\end{table}

\begin{table}[p]
  \caption{\Essays{} at Pythia-1B: the settings on each method's lower-left frontier in
  Figure~\ref{fig:app-frontier-essays-1b}, with the point's held-out
  cross-entropy and worst-case $\nauc$ in gray. Every run
  uses the recipe of Table~\ref{tab:app-settings-essays-70m} at learning rate $10^{-5}$; DP runs list their lot $L$, epoch budget $T$ and learning rate; only the listed hyperparameter varies. $w$ and $\epsilon$ are the \trap{} weight and
  threshold, $r$ the adapter rank, $\varepsilon$ under DP the privacy budget.}
  \label{tab:app-settings-essays-1b}
  \centering\footnotesize
  \begin{tabular}{@{}l>{\raggedright\arraybackslash}p{0.61\linewidth}@{}}
    \toprule
    Method & Settings on the frontier (cross-entropy, worst-case $\nauc$) \\
    \midrule
\trap{}, frozen reference & $w{=}1,\ \epsilon{=}0$ ({\color{gray}2.65, 0.51}); $w{=}1,\ \epsilon{=}0.05$ ({\color{gray}2.66, 0.50}); $w{=}1.5,\ \epsilon{=}0.1$ ({\color{gray}2.66, 0.32}); $w{=}1.5,\ \epsilon{=}0.05$ ({\color{gray}2.66, 0.30}); $w{=}1.5,\ \epsilon{=}0$ ({\color{gray}2.67, 0.28}); $w{=}2,\ \epsilon{=}0.1$ ({\color{gray}2.67, 0.25}); $w{=}2,\ \epsilon{=}0.05$ ({\color{gray}2.68, 0.23}); $w{=}2,\ \epsilon{=}0$ ({\color{gray}2.68, 0.21}); $w{=}2,\ \epsilon{=}-0.02$ ({\color{gray}2.68, 0.19}); $w{=}2,\ \epsilon{=}-0.05$ ({\color{gray}2.70, 0.19}); $w{=}2,\ \epsilon{=}0.05$ ({\color{gray}2.87, 0.12}) \\
\trap{}, iterated rounds & $w{=}0.5,\ \epsilon{=}0$, round 2 (first round mutual) ({\color{gray}2.65, 0.63}); $w{=}0.5,\ \epsilon{=}0$, round 2 (first round alternating) ({\color{gray}2.65, 0.62}); $w{=}1,\ \epsilon{=}0$, round 2 (first round mutual) ({\color{gray}2.65, 0.60}); $w{=}1,\ \epsilon{=}0$, round 2 (reference: undefended model) ({\color{gray}2.65, 0.56}); $w{=}0.6,\ \epsilon{=}0$, round 2 ({\color{gray}2.66, 0.56}); $w{=}0.25,\ \epsilon{=}0$, round 3 ({\color{gray}2.67, 0.44}); $w{=}2,\ \epsilon{=}0$, round 2 ({\color{gray}2.72, 0.21}) \\
\trap{} + LoRA & $r{=}4,\ w{=}1,\ \epsilon{=}0$ ({\color{gray}2.66, 0.55}); $r{=}16,\ w{=}1.5,\ \epsilon{=}0$ ({\color{gray}2.67, 0.26}); $r{=}8,\ w{=}2,\ \epsilon{=}0$ ({\color{gray}2.68, 0.21}); $r{=}256,\ w{=}2,\ \epsilon{=}0$ ({\color{gray}2.68, 0.21}); $r{=}16,\ w{=}2,\ \epsilon{=}-0.02$ ({\color{gray}2.69, 0.17}) \\
\trap{} + LoRA, iterated rounds & $r{=}4,\ w{=}0.9,\ \epsilon{=}0$, round 1 ({\color{gray}2.65, 0.56}); $r{=}4,\ w{=}0.9,\ \epsilon{=}0$, round 2 ({\color{gray}2.65, 0.51}); $r{=}4,\ w{=}1.3,\ \epsilon{=}0$, round 2 ({\color{gray}2.66, 0.33}); $r{=}16,\ w{=}1.5,\ \epsilon{=}0$, round 1 ({\color{gray}2.67, 0.26}) \\
LoRA & $r{=}256$ ({\color{gray}2.68, 0.27}) \\
LoRA on $\Dtgt\cup\Dref$ & $r{=}256$ ({\color{gray}2.64, 0.40}); $r{=}4$ ({\color{gray}2.64, 0.37}); $r{=}16$ ({\color{gray}2.64, 0.36}) \\
DP-AdamW & $\varepsilon{=}64$ ($L{=}1024$, $T{=}10$, lr $10^{-4}$) ({\color{gray}2.81, 0.18}) \\
DP-SGD & $\varepsilon{=}8$ ($L{=}8$, $T{=}60$, lr $10^{-4}$) ({\color{gray}2.94, 0.20}); $\varepsilon{=}8$ ($L{=}128$, $T{=}10$, lr $10^{-3}$) ({\color{gray}2.98, 0.20}) \\
DP-SGD on $\Dtgt\cup\Dref$ & $\varepsilon{=}1$ ($L{=}1024$, $T{=}10$, lr $10^{-5}$) ({\color{gray}3.00, 0.21}) \\
Goldfish loss & $k{=}32$ ({\color{gray}2.68, 0.64}); $k{=}3$ ({\color{gray}2.70, 0.46}) \\
Goldfish loss on $\Dtgt\cup\Dref$ & $k{=}32$ ({\color{gray}2.65, 0.37}); $k{=}8$ ({\color{gray}2.65, 0.35}); $k{=}4$ ({\color{gray}2.66, 0.33}); $k{=}3$ ({\color{gray}2.67, 0.33}) \\
Flooding & $b{=}1.5$ ({\color{gray}2.67, 0.65}); $b{=}2$ ({\color{gray}2.68, 0.40}); $b{=}2.5$ ({\color{gray}2.73, 0.37}) \\
Flooding on $\Dtgt\cup\Dref$ & $b{=}1.5$ ({\color{gray}2.64, 0.38}); $b{=}2$ ({\color{gray}2.65, 0.36}); $b{=}2.5$ ({\color{gray}2.70, 0.32}) \\
Confidence penalty & $\beta{=}0.1$ ({\color{gray}2.70, 0.59}) \\
Confidence penalty on $\Dtgt\cup\Dref$ & $\beta{=}0.1$ ({\color{gray}2.67, 0.53}) \\
Loss truncation & $c{=}0.1$ ({\color{gray}2.80, 0.31}) \\
Loss truncation on $\Dtgt\cup\Dref$ & $c{=}0.1$ ({\color{gray}2.75, 0.27}); $c{=}0.2$ ({\color{gray}2.98, 0.20}) \\
RelaxLoss & $\alpha{=}2$ ({\color{gray}2.67, 0.64}); $\alpha{=}1.5$ ({\color{gray}2.67, 0.48}); $\alpha{=}2.5$ ({\color{gray}2.72, 0.37}) \\
RelaxLoss on $\Dtgt\cup\Dref$ & $\alpha{=}1.5$ ({\color{gray}2.64, 0.53}); $\alpha{=}2$ ({\color{gray}2.65, 0.37}); $\alpha{=}2.5$ ({\color{gray}2.70, 0.34}) \\
KL anchor (base) & $\lambda{=}0.1$ ({\color{gray}2.67, 0.54}); $\lambda{=}1$ ({\color{gray}2.74, 0.50}) \\
KL anchor (base) on $\Dtgt\cup\Dref$ & $\lambda{=}0.1$ ({\color{gray}2.63, 0.53}); $\lambda{=}0.05$ ({\color{gray}2.64, 0.53}); $\lambda{=}0.5$ ({\color{gray}2.67, 0.52}); $\lambda{=}1$ ({\color{gray}2.72, 0.51}); $\lambda{=}2$ ({\color{gray}2.77, 0.47}) \\
KL anchor (reference) & $\lambda{=}2$ ({\color{gray}2.65, 0.51}) \\
$n$-gram regularizer & $\lambda{=}0.5,\ \tau{=}0$ ({\color{gray}2.68, 0.57}); $\lambda{=}2,\ \tau{=}0$ ({\color{gray}2.69, 0.42}) \\
$n$-gram regularizer on $\Dtgt\cup\Dref$ & $\lambda{=}0.5,\ \tau{=}0$ ({\color{gray}2.64, 0.54}); $\lambda{=}2,\ \tau{=}0$ ({\color{gray}2.65, 0.37}) \\
NEFTune & $\alpha{=}5$ ({\color{gray}2.68, 0.63}) \\
NEFTune on $\Dtgt\cup\Dref$ & $\alpha{=}10$ ({\color{gray}2.64, 0.53}); $\alpha{=}5$ ({\color{gray}2.64, 0.37}) \\
Dropout / weight decay & weight decay $0.01$ ({\color{gray}2.68, 0.63}); weight decay $0.1$ ({\color{gray}2.68, 0.37}) \\
Dropout / weight decay on $\Dtgt\cup\Dref$ & weight decay $0.1$ ({\color{gray}2.64, 0.37}); weight decay $0.01$ ({\color{gray}2.64, 0.36}) \\
    \bottomrule
  \end{tabular}
\end{table}

\begin{table}[p]
  \caption{\medcase{} at Pythia-1B, twenty cases, best-val: the settings on each method's lower-left frontier in
  Figure~\ref{fig:app-frontier-medcase-1b}, with the point's held-out
  cross-entropy and worst-case $\nauc$ in gray. Every run
  uses learning rate $5\cdot10^{-5}$, $16$ cases per step, bf16, $3\%$ warmup, no weight decay, a $15$-epoch ceiling with patience $4$ on the validation split; LoRA adapters on the attention projections with $\alpha{=}2r$ and dropout $0.1$; only the listed hyperparameter varies. $w$ and $\epsilon$ are the \trap{} weight and
  threshold, $r$ the adapter rank, $\varepsilon$ under DP the privacy budget.}
  \label{tab:app-settings-medcase-1b}
  \centering\footnotesize
  \begin{tabular}{@{}l>{\raggedright\arraybackslash}p{0.61\linewidth}@{}}
    \toprule
    Method & Settings on the frontier (cross-entropy, worst-case $\nauc$) \\
    \midrule
\trap{}, frozen reference & $w{=}0.5,\ \epsilon{=}0$ ({\color{gray}2.14, 0.76}); $w{=}1,\ \epsilon{=}0$ ({\color{gray}2.17, 0.76}); $w{=}2,\ \epsilon{=}0$ ({\color{gray}2.64, 0.38}); $w{=}2,\ \epsilon{=}0.05$ ({\color{gray}2.74, 0.33}); $w{=}4,\ \epsilon{=}0$ ({\color{gray}2.92, 0.24}) \\
\trap{}, iterated rounds & $w{=}0.5,\ \epsilon{=}0$, round 4 ({\color{gray}2.12, 0.74}); $w{=}1,\ \epsilon{=}0$, round 5 ({\color{gray}2.15, 0.67}); $w{=}1,\ \epsilon{=}0$, round 2 ({\color{gray}2.16, 0.65}); $w{=}1,\ \epsilon{=}0$, round 3 ({\color{gray}2.21, 0.59}); $w{=}2,\ \epsilon{=}0$, round 2 ({\color{gray}2.65, 0.39}); $w{=}2,\ \epsilon{=}0$, round 4 ({\color{gray}2.94, 0.26}); $w{=}2,\ \epsilon{=}0$, round 3 ({\color{gray}3.04, 0.17}) \\
LoRA & $r{=}64$ ({\color{gray}3.11, 0.24}); $r{=}32$ ({\color{gray}3.69, 0.14}); $r{=}16$ ({\color{gray}4.33, 0.10}); $r{=}8$ ({\color{gray}4.77, 0.10}); $r{=}4$ ({\color{gray}5.27, 0.10}) \\
LoRA on $\Dtgt\cup\Dref$ & $r{=}8$ ({\color{gray}4.32, 0.13}) \\
DP-AdamW & $\varepsilon{=}16$ (lot 256) ({\color{gray}3.93, 0.18}); $\varepsilon{=}8$ (lot 256) ({\color{gray}4.57, 0.09}) \\
DP-SGD & $\varepsilon{=}1$ (lot 256) ({\color{gray}5.63, 0.10}); $\varepsilon{=}4$ (lot 256) ({\color{gray}5.65, 0.09}); $\varepsilon{=}2$ (lot 256) ({\color{gray}5.65, 0.09}) \\
DP-AdamW on $\Dtgt\cup\Dref$ & $\varepsilon{=}8$ (lot 256) ({\color{gray}3.68, 0.10}) \\
DP-SGD on $\Dtgt\cup\Dref$ & $\varepsilon{=}8$ (lot 256) ({\color{gray}5.66, 0.10}) \\
Goldfish loss & $k{=}8$ ({\color{gray}2.28, 0.79}); $k{=}32$ ({\color{gray}2.33, 0.76}); $k{=}3$ ({\color{gray}2.34, 0.69}); $k{=}4$ ({\color{gray}2.38, 0.68}) \\
Flooding & $b{=}1.5$ ({\color{gray}2.38, 0.78}); $b{=}2$ ({\color{gray}3.20, 0.74}); $b{=}2.5$ ({\color{gray}3.20, 0.72}); $b{=}3$ ({\color{gray}3.25, 0.68}) \\
Confidence penalty & $\beta{=}0.1$ ({\color{gray}2.27, 0.83}); $\beta{=}0.5$ ({\color{gray}3.31, 0.62}); $\beta{=}1$ ({\color{gray}4.62, 0.11}) \\
Loss truncation & $c{=}0.1$ ({\color{gray}3.81, 0.79}); $c{=}0.2$ ({\color{gray}3.86, 0.75}); $c{=}0.4$ ({\color{gray}5.18, 0.48}) \\
RelaxLoss & $\alpha{=}3$ ({\color{gray}2.29, 0.76}); $\alpha{=}2$ ({\color{gray}2.33, 0.73}) \\
KL anchor (base) & $\lambda{=}0.5$ ({\color{gray}2.15, 0.82}); $\lambda{=}1$ ({\color{gray}2.49, 0.78}); $\lambda{=}2$ ({\color{gray}2.73, 0.65}) \\
KL anchor (reference) & $\lambda{=}0.5$ ({\color{gray}2.22, 0.82}); $\lambda{=}1$ ({\color{gray}2.25, 0.74}) \\
$n$-gram regularizer & $\lambda{=}2$ ({\color{gray}2.28, 0.70}); $\lambda{=}8$ ({\color{gray}2.45, 0.64}) \\
NEFTune & $\alpha{=}10$ ({\color{gray}2.20, 0.84}) \\
Weight decay & weight decay $0.1$ ({\color{gray}2.28, 0.90}); weight decay $0.3$ ({\color{gray}2.29, 0.88}) \\
Dropout & dropout $p{=}0.1$ ({\color{gray}2.07, 0.84}); dropout $p{=}0.3$ ({\color{gray}2.22, 0.84}) \\
Gradient clipping & max norm $0.1$ ({\color{gray}2.29, 0.88}) \\
    \bottomrule
  \end{tabular}
\end{table}

\begin{table}[p]
  \caption{\medcase{} at Llama-3.1-8B-Instruct, twenty cases, best-val: the settings on each method's lower-left frontier in
  Figure~\ref{fig:app-frontier-medcase-llama8b}, with the point's held-out
  cross-entropy and worst-case $\nauc$ in gray. Every run
  uses learning rate $2\cdot10^{-5}$, $8$ cases per step, bf16, the stopping rule of Table~\ref{tab:app-settings-medcase-1b}; the LoRA adapters (rank $16$, $\alpha{=}32$, dropout $0.05$) train at learning rate $10^{-4}$; only the listed hyperparameter varies. $w$ and $\epsilon$ are the \trap{} weight and
  threshold, $r$ the adapter rank, $\varepsilon$ under DP the privacy budget.}
  \label{tab:app-settings-medcase-llama8b}
  \centering\footnotesize
  \begin{tabular}{@{}l>{\raggedright\arraybackslash}p{0.61\linewidth}@{}}
    \toprule
    Method & Settings on the frontier (cross-entropy, worst-case $\nauc$) \\
    \midrule
\trap{}, frozen reference & $w{=}1,\ \epsilon{=}0.05$ ({\color{gray}1.40, 0.64}); $w{=}1,\ \epsilon{=}0$ ({\color{gray}1.40, 0.62}); $w{=}2,\ \epsilon{=}0$ ({\color{gray}1.71, 0.30}); $w{=}32,\ \epsilon{=}0.05$ ({\color{gray}2.89, 0.18}) \\
\trap{}, iterated rounds & $w{=}2,\ \epsilon{=}0$, round 3 ({\color{gray}2.00, 0.26}); $w{=}2,\ \epsilon{=}0$, round 4 ({\color{gray}2.06, 0.17}) \\
LoRA & $r{=}16$ ({\color{gray}2.91, 0.20}) \\
LoRA on $\Dtgt\cup\Dref$ & $r{=}16$ ({\color{gray}2.38, 0.05}) \\
DP-AdamW & $\varepsilon{=}16$ (lot 256) ({\color{gray}9.72, 0.26}); $\varepsilon{=}8$ (lot 256) ({\color{gray}11.88, 0.07}) \\
DP-SGD & $\varepsilon{=}1$ (lot 256) ({\color{gray}2.95, 0.22}); $\varepsilon{=}2$ (lot 256) ({\color{gray}3.03, 0.19}) \\
DP-AdamW on $\Dtgt\cup\Dref$ & $\varepsilon{=}8$ (lot 256) ({\color{gray}9.10, 0.14}) \\
DP-SGD on $\Dtgt\cup\Dref$ & $\varepsilon{=}8$ (lot 256) ({\color{gray}3.11, 0.05}) \\
Weight decay & weight decay $0.3$ ({\color{gray}1.62, 0.56}) \\
Gradient clipping & max norm $0.1$ ({\color{gray}1.63, 0.55}) \\
    \bottomrule
  \end{tabular}
\end{table}

\begin{table}[p]
  \caption{\medcase{} at Pythia-1B, twenty cases, overtrained: the settings on each method's lower-left frontier in
  Figure~\ref{fig:app-frontier-medcase-1b-over}, with the point's held-out
  cross-entropy and worst-case $\nauc$ in gray. Every run
  uses the recipe of Table~\ref{tab:app-settings-medcase-1b} trained for a fixed $40$ epochs; only the listed hyperparameter varies. $w$ and $\epsilon$ are the \trap{} weight and
  threshold, $r$ the adapter rank, $\varepsilon$ under DP the privacy budget.}
  \label{tab:app-settings-medcase-1b-over}
  \centering\footnotesize
  \begin{tabular}{@{}l>{\raggedright\arraybackslash}p{0.61\linewidth}@{}}
    \toprule
    Method & Settings on the frontier (cross-entropy, worst-case $\nauc$) \\
    \midrule
\trap{}, frozen reference & $w{=}1,\ \epsilon{=}0.05$ ({\color{gray}2.29, 0.78}); $w{=}1,\ \epsilon{=}0$ ({\color{gray}2.59, 0.71}); $w{=}2,\ \epsilon{=}0.05$ ({\color{gray}3.05, 0.36}) \\
LoRA & $r{=}8$ ({\color{gray}3.32, 0.14}) \\
LoRA on $\Dtgt\cup\Dref$ & $r{=}8$ ({\color{gray}2.59, 0.24}) \\
DP-AdamW & $\varepsilon{=}16$ (lot 256) ({\color{gray}5.28, 0.10}); $\varepsilon{=}4$ (lot 256) ({\color{gray}6.40, 0.05}) \\
DP-SGD & $\varepsilon{=}1$ (lot 256) ({\color{gray}5.51, 0.04}) \\
DP-AdamW on $\Dtgt\cup\Dref$ & $\varepsilon{=}8$ (lot 256) ({\color{gray}4.88, 0.10}) \\
DP-SGD on $\Dtgt\cup\Dref$ & $\varepsilon{=}8$ (lot 256) ({\color{gray}5.60, 0.10}) \\
Weight decay & weight decay $0.3$ ({\color{gray}2.78, 0.96}) \\
Gradient clipping & max norm $0.1$ ({\color{gray}2.58, 0.93}) \\
    \bottomrule
  \end{tabular}
\end{table}

\begin{table}[p]
  \caption{The rare-minority mixture at Pythia-70M: the settings on each method's lower-left frontier in
  Figure~\ref{fig:app-frontier-mixture-70m}, with the point's held-out
  cross-entropy and worst-case $\nauc$ in gray. Every run
  uses learning rate $5\cdot10^{-5}$, $16$ examples per step, $3\%$ warmup, fp32, no weight decay, a $40$-epoch ceiling with patience $3$ on $200$ held-out examples; LoRA adapters on the attention projections with $\alpha{=}2r$ and dropout $0.1$, ranks $1$ and $2$ with an $80$-epoch ceiling; DP with lot $256$, learning rate $10^{-3}$, clipping $0.1$ and $\delta{=}10^{-5}$; only the listed hyperparameter varies. $w$ and $\epsilon$ are the \trap{} weight and
  threshold, $r$ the adapter rank, $\varepsilon$ under DP the privacy budget.}
  \label{tab:app-settings-mixture-70m}
  \centering\footnotesize
  \begin{tabular}{@{}l>{\raggedright\arraybackslash}p{0.61\linewidth}@{}}
    \toprule
    Method & Settings on the frontier (cross-entropy, worst-case $\nauc$) \\
    \midrule
\trap{}, frozen reference & $w{=}0.5,\ \epsilon{=}0$ ({\color{gray}0.0148, 0.94}); $w{=}1,\ \epsilon{=}0$ ({\color{gray}0.0191, 0.45}); $w{=}2,\ \epsilon{=}0.05$ ({\color{gray}0.0260, 0.24}); $w{=}2,\ \epsilon{=}0$ ({\color{gray}0.0262, 0.24}); $w{=}4,\ \epsilon{=}0$ ({\color{gray}0.0285, 0.14}) \\
\trap{}, iterated rounds & $w{=}2,\ \epsilon{=}0$, round 2 ({\color{gray}0.0278, 0.12}) \\
\trap{} + LoRA & $r{=}64,\ w{=}1,\ \epsilon{=}0$ ({\color{gray}0.0149, 0.17}); $r{=}8,\ w{=}1,\ \epsilon{=}0$ ({\color{gray}0.0191, 0.17}); $r{=}4,\ w{=}1,\ \epsilon{=}0$ ({\color{gray}0.0222, 0.13}); $r{=}8,\ w{=}4,\ \epsilon{=}0$ ({\color{gray}0.0298, 0.11}); $r{=}4,\ w{=}4,\ \epsilon{=}0$ ({\color{gray}0.0336, 0.10}) \\
\trap{} + LoRA, iterated rounds & $r{=}64,\ w{=}2,\ \epsilon{=}0$, round 2 ({\color{gray}0.0179, 0.13}) \\
LoRA & $r{=}64$ ({\color{gray}0.0132, 0.49}); $r{=}8$ ({\color{gray}0.0177, 0.23}); $r{=}2$ ({\color{gray}0.0204, 0.22}); $r{=}4$ ({\color{gray}0.0205, 0.18}); $r{=}1$ ({\color{gray}0.0253, 0.16}) \\
LoRA on $\Dtgt\cup\Dref$ & $r{=}64$ ({\color{gray}0.0106, 0.38}); $r{=}8$ ({\color{gray}0.0135, 0.38}); $r{=}4$ ({\color{gray}0.0147, 0.37}) \\
DP-AdamW & $\varepsilon{=}256$ (lot 1024) ({\color{gray}0.0378, 0.51}); $\varepsilon{=}64$ (lot 1024) ({\color{gray}0.0421, 0.41}); $\varepsilon{=}16$ (lot 1024) ({\color{gray}0.12, 0.22}); $\varepsilon{=}8$ (lot 1024) ({\color{gray}0.32, 0.18}); $\varepsilon{=}1$ (lot 256) ({\color{gray}10.03, 0.09}) \\
DP-AdamW on $\Dtgt\cup\Dref$ & $\varepsilon{=}64$ (lot 1024) ({\color{gray}0.0378, 0.33}); $\varepsilon{=}8$ (lot 256) ({\color{gray}0.94, 0.19}) \\
DP-SGD & $\varepsilon{=}8$ (lot 256) ({\color{gray}0.56, 0.12}) \\
Goldfish loss & $k{=}3$ ({\color{gray}0.0398, 0.46}) \\
Flooding & $b{=}0.01$ ({\color{gray}0.0251, 0.82}); $b{=}0.03$ ({\color{gray}0.0669, 0.52}) \\
Confidence penalty & $\beta{=}0.1$ ({\color{gray}0.0168, 0.96}); $\beta{=}0.5$ ({\color{gray}1.59, 0.55}) \\
Loss truncation & $c{=}0.2$ ({\color{gray}3.39, 0.14}) \\
RelaxLoss & $\alpha{=}0.01$ ({\color{gray}0.0226, 0.92}); $\alpha{=}0.03$ ({\color{gray}0.43, 0.51}) \\
KL anchor (base) & $\lambda{=}0.5$ ({\color{gray}0.15, 0.68}) \\
KL anchor (reference) & $\lambda{=}0.5$ ({\color{gray}0.0157, 0.90}) \\
$n$-gram regularizer & $\lambda{=}2$ ({\color{gray}0.23, 0.65}) \\
NEFTune & $\alpha{=}5$ ({\color{gray}0.0152, 0.98}) \\
Weight decay & weight decay $0.3$ ({\color{gray}0.0160, 0.95}); weight decay $0.1$ ({\color{gray}0.0161, 0.94}) \\
Dropout & dropout $p{=}0.1$ ({\color{gray}0.0120, 1.00}) \\
Gradient clipping & max norm $0.1$ ({\color{gray}0.0146, 0.99}) \\
    \bottomrule
  \end{tabular}
\end{table}

\begin{table}[p]
  \caption{The rare-minority mixture at Pythia-160M: the settings on each method's lower-left frontier in
  Figure~\ref{fig:app-frontier-mixture-160m}, with the point's held-out
  cross-entropy and worst-case $\nauc$ in gray. Every run
  uses the recipe of Table~\ref{tab:app-settings-mixture-70m}; only the listed hyperparameter varies. $w$ and $\epsilon$ are the \trap{} weight and
  threshold, $r$ the adapter rank, $\varepsilon$ under DP the privacy budget.}
  \label{tab:app-settings-mixture-160m}
  \centering\footnotesize
  \begin{tabular}{@{}l>{\raggedright\arraybackslash}p{0.61\linewidth}@{}}
    \toprule
    Method & Settings on the frontier (cross-entropy, worst-case $\nauc$) \\
    \midrule
\trap{}, frozen reference & $w{=}0.5,\ \epsilon{=}0$ ({\color{gray}0.0094, 0.99}); $w{=}1,\ \epsilon{=}0$ ({\color{gray}0.0106, 0.58}); $w{=}2,\ \epsilon{=}0.05$ ({\color{gray}0.0143, 0.14}); $w{=}2,\ \epsilon{=}0$ ({\color{gray}0.0149, 0.08}); $w{=}4,\ \epsilon{=}0$ ({\color{gray}0.0197, 0.06}) \\
\trap{}, iterated rounds & $w{=}2,\ \epsilon{=}0$, round 2 ({\color{gray}0.0152, 0.27}); $w{=}2,\ \epsilon{=}0$, round 3 ({\color{gray}0.0181, 0.16}); $w{=}4,\ \epsilon{=}0$, round 2 ({\color{gray}0.0254, 0.10}) \\
\trap{} + LoRA & $r{=}64,\ w{=}1,\ \epsilon{=}0$ ({\color{gray}0.0084, 0.14}); $r{=}64,\ w{=}2,\ \epsilon{=}0$ ({\color{gray}0.0106, 0.11}); $r{=}64,\ w{=}4,\ \epsilon{=}0$ ({\color{gray}0.0133, 0.09}); $r{=}4,\ w{=}4,\ \epsilon{=}0$ ({\color{gray}0.0151, 0.08}) \\
\trap{} + LoRA, iterated rounds & $r{=}64,\ w{=}2,\ \epsilon{=}0$, round 2 ({\color{gray}0.0110, 0.11}); $r{=}8,\ w{=}4,\ \epsilon{=}0$, round 2 ({\color{gray}0.0169, 0.10}) \\
LoRA & $r{=}64$ ({\color{gray}0.0070, 0.51}); $r{=}8$ ({\color{gray}0.0076, 0.34}); $r{=}4$ ({\color{gray}0.0082, 0.25}); $r{=}2$ ({\color{gray}0.0091, 0.23}); $r{=}1$ ({\color{gray}0.0098, 0.17}) \\
LoRA on $\Dtgt\cup\Dref$ & $r{=}64$ ({\color{gray}0.0048, 0.64}); $r{=}8$ ({\color{gray}0.0061, 0.41}) \\
DP-AdamW & $\varepsilon{=}256$ (lot 1024) ({\color{gray}0.0192, 0.51}); $\varepsilon{=}64$ (lot 1024) ({\color{gray}0.0497, 0.14}); $\varepsilon{=}1$ (lot 256) ({\color{gray}13.74, 0.07}) \\
DP-AdamW on $\Dtgt\cup\Dref$ & $\varepsilon{=}64$ (lot 1024) ({\color{gray}0.0239, 0.48}); $\varepsilon{=}8$ (lot 256) ({\color{gray}1.07, 0.29}) \\
DP-SGD & $\varepsilon{=}8$ (lot 256) ({\color{gray}0.30, 0.10}) \\
Goldfish loss & $k{=}3$ ({\color{gray}0.0254, 0.72}) \\
Flooding & $b{=}0.01$ ({\color{gray}0.0168, 0.82}); $b{=}0.03$ ({\color{gray}0.0313, 0.68}) \\
Confidence penalty & $\beta{=}0.1$ ({\color{gray}0.0113, 0.97}); $\beta{=}0.5$ ({\color{gray}1.53, 0.68}) \\
Loss truncation & $c{=}0.2$ ({\color{gray}4.06, 0.15}) \\
RelaxLoss & $\alpha{=}0.01$ ({\color{gray}0.0154, 0.84}); $\alpha{=}0.03$ ({\color{gray}0.0278, 0.60}) \\
KL anchor (base) & $\lambda{=}0.5$ ({\color{gray}0.0692, 0.80}) \\
KL anchor (reference) & $\lambda{=}0.5$ ({\color{gray}0.0110, 0.93}) \\
$n$-gram regularizer & $\lambda{=}2$ ({\color{gray}0.15, 0.98}) \\
NEFTune & $\alpha{=}5$ ({\color{gray}0.0097, 0.99}) \\
Weight decay & weight decay $0.3$ ({\color{gray}0.0115, 0.98}) \\
Dropout & dropout $p{=}0.1$ ({\color{gray}0.0103, 0.96}) \\
Gradient clipping & max norm $0.1$ ({\color{gray}0.0139, 0.80}) \\
    \bottomrule
  \end{tabular}
\end{table}

\subsection{One score at a time}

\begin{figure}[p]
  \centering
  \includegraphics[width=\linewidth]{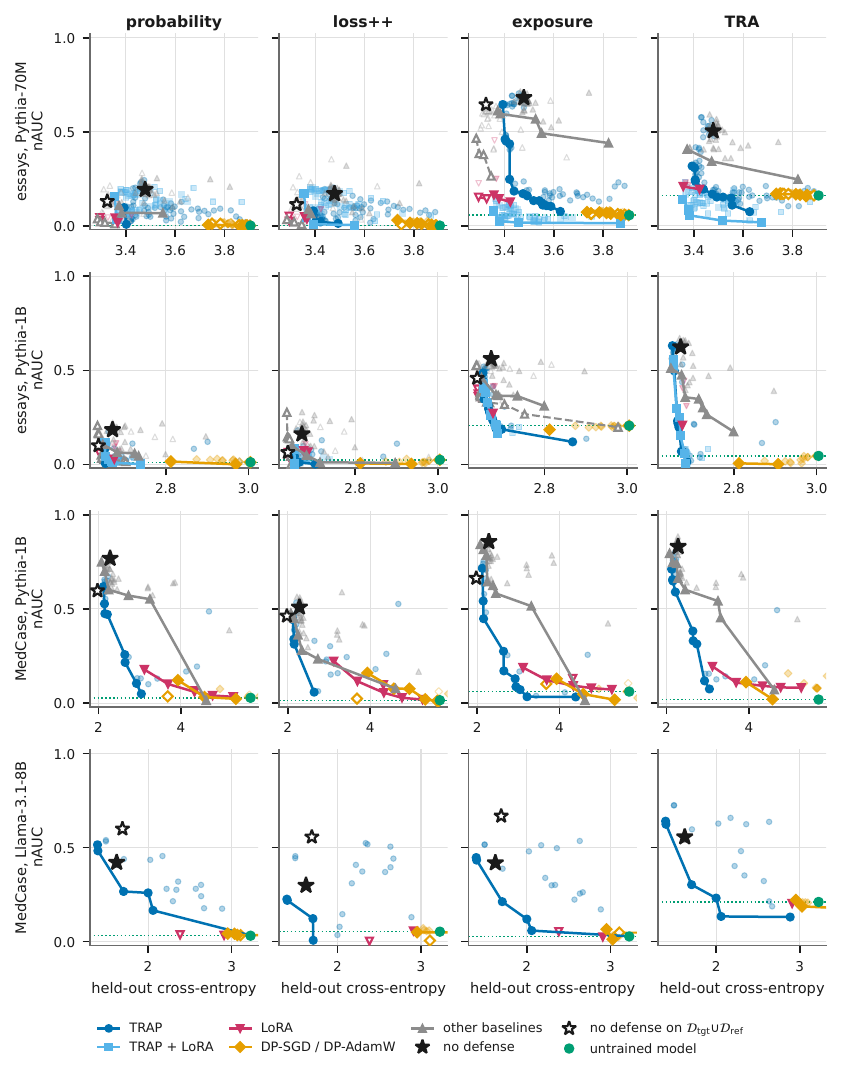}
  \caption{The family frontiers one score at a time, best-val checkpoint:
  rows are the \essays{} at Pythia-70M and 1B, \medcase{} at Pythia-1B and
  Llama-3.1-8B (the second and third rows are Figure~\ref{fig:mitigation}a
  and~b); columns are the four scores; lines are the family frontiers.}
  \label{fig:app-frontier-scores}
\end{figure}

\begin{figure}[p]
  \centering
  \includegraphics[width=\linewidth]{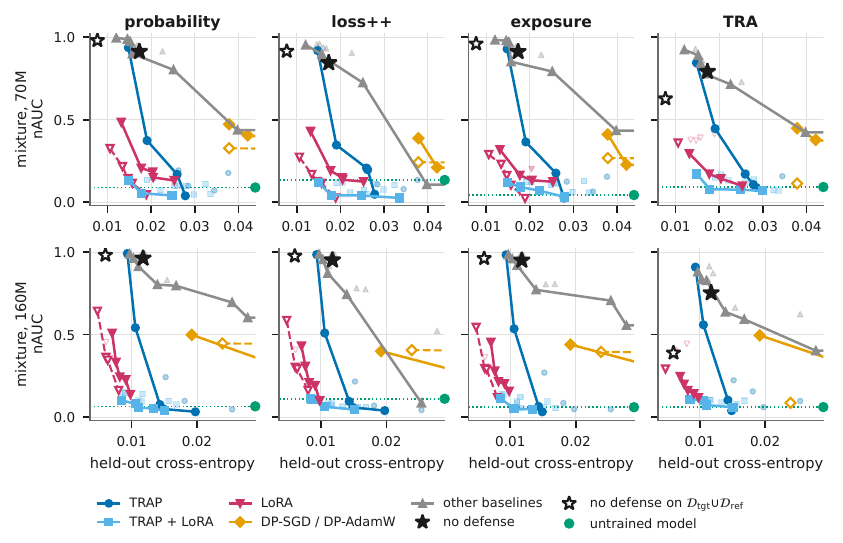}
  \caption{The rare-minority mixture one score at a time, best-val
  checkpoint: rows are the two model sizes (Figure~\ref{fig:mitigation}c is
  the 160M row), columns the four scores, lines the family frontiers of the
  main figure.}
  \label{fig:app-frontier-scores-mixture}
\end{figure}

Figure~\ref{fig:app-frontier-scores} splits the worst case of
the family frontiers into its four scores, with the methods pooled
into the same families. Which score is strongest depends on the testbed: on
the \essays{} the undefended model is exposed mostly through exposure and
\tra{}, while probability and loss++ sit close to the floor even without a
defense, and on \medcase{} all four scores are high. The reduction \trap{}
achieves holds under every score, not only under \tra{}, which the penalty
is trained against. \tra{} needs the reference model, so an attacker holding
only the released model is described by the first three columns.
Figure~\ref{fig:app-frontier-scores-mixture} does the same for the mixture,
where all four scores expose the undefended model and the family frontiers
keep their order under each of them.

\subsection{Directional AUC versions of the frontiers}
\label{app:raw-auc-frontiers}

\begin{figure}[p]
  \centering
  \includegraphics[width=\linewidth]{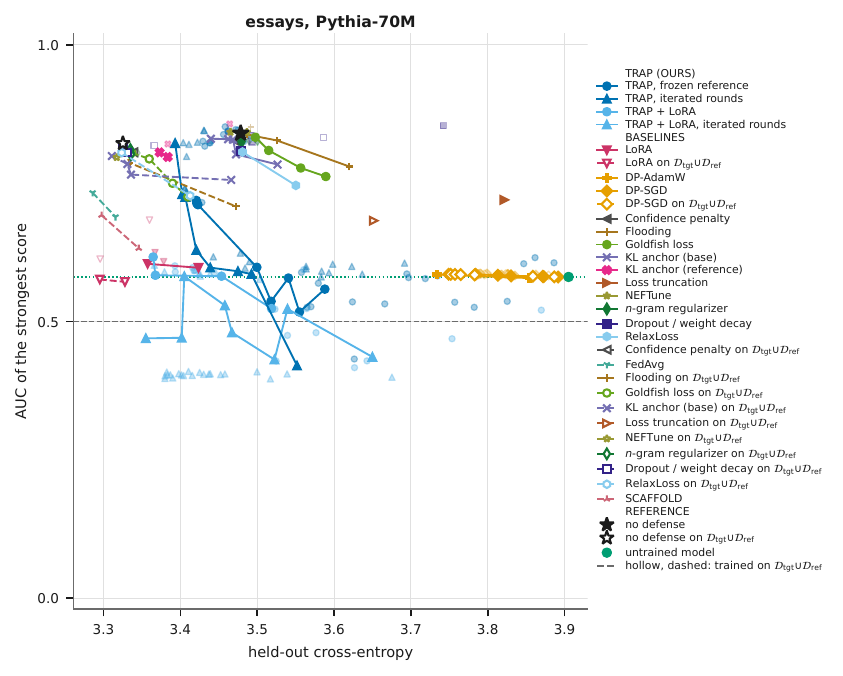}
  \caption{Directional AUC version of
  Figure~\ref{fig:app-frontier-essays-70m}: \essays{} at Pythia-70M, with
  $0.5$ marking chance; points below the line are over-corrected models that
  $\nauc$ folds upward.}
  \label{fig:app-frontier-essays-70m-auc}
\end{figure}

\begin{figure}[p]
  \centering
  \includegraphics[width=\linewidth]{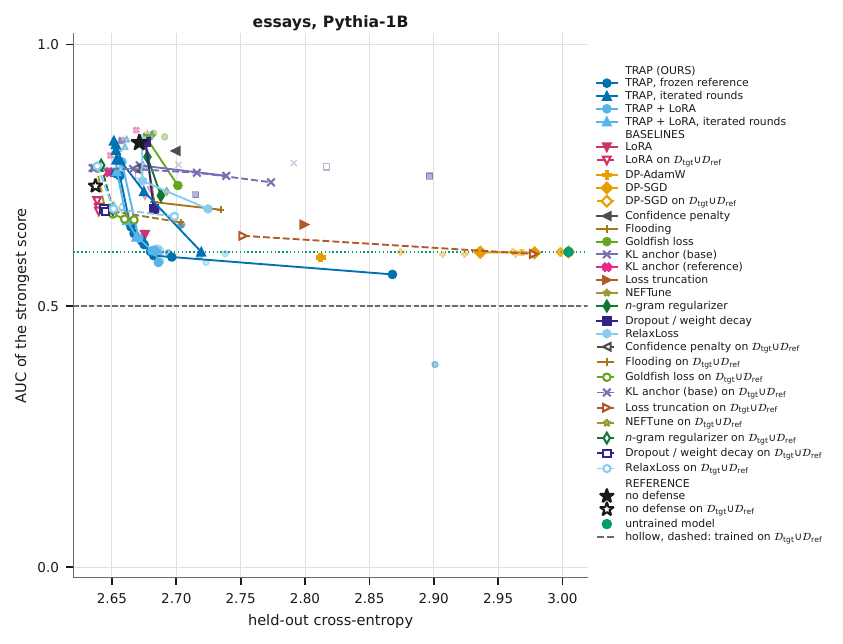}
  \caption{Directional AUC version of
  Figure~\ref{fig:app-frontier-essays-1b}: \essays{} at Pythia-1B.}
  \label{fig:app-frontier-essays-1b-auc}
\end{figure}

\begin{figure}[p]
  \centering
  \includegraphics[width=\linewidth]{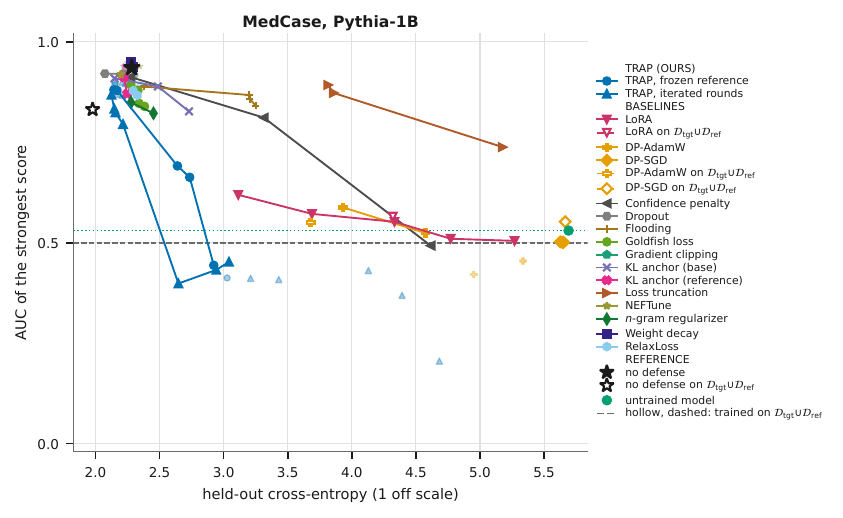}
  \caption{Directional AUC version of
  Figure~\ref{fig:app-frontier-medcase-1b}: \medcase{} at Pythia-1B.}
  \label{fig:app-frontier-medcase-1b-auc}
\end{figure}

\begin{figure}[p]
  \centering
  \includegraphics[width=\linewidth]{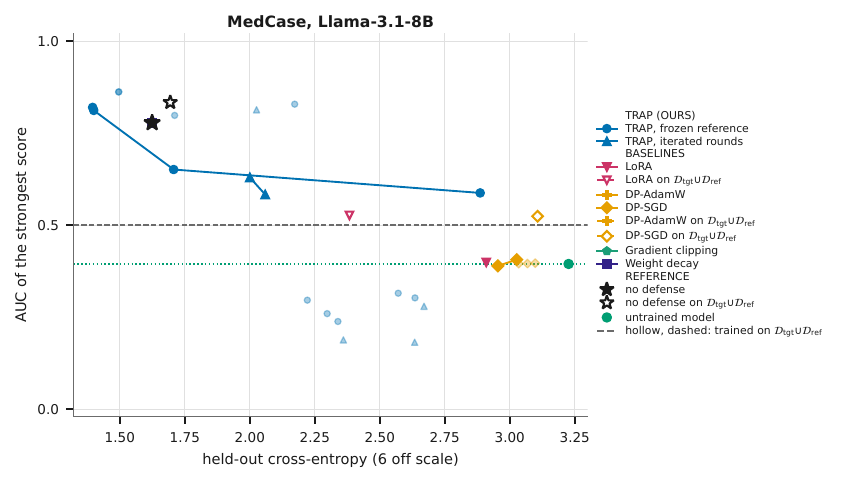}
  \caption{Directional AUC version of
  Figure~\ref{fig:app-frontier-medcase-llama8b}: \medcase{} at
  Llama-3.1-8B-Instruct.}
  \label{fig:app-frontier-medcase-llama8b-auc}
\end{figure}

\begin{figure}[p]
  \centering
  \includegraphics[width=\linewidth]{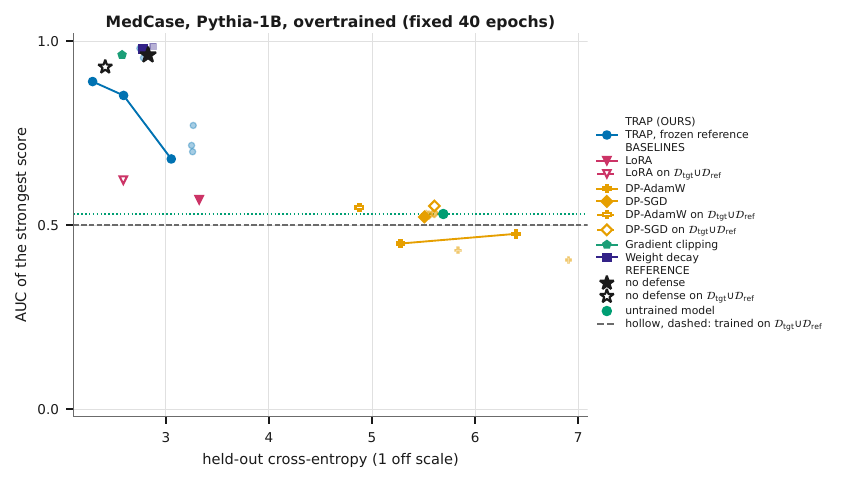}
  \caption{Directional AUC version of
  Figure~\ref{fig:app-frontier-medcase-1b-over}: \medcase{} at Pythia-1B
  trained for a fixed forty epochs.}
  \label{fig:app-frontier-medcase-1b-over-auc}
\end{figure}

\begin{figure}[p]
  \centering
  \includegraphics[width=\linewidth]{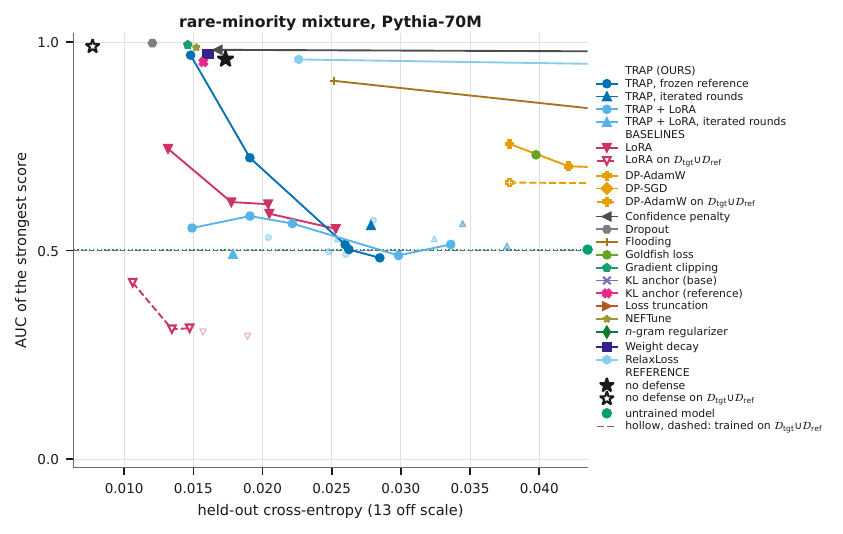}
  \caption{Directional AUC version of
  Figure~\ref{fig:app-frontier-mixture-70m}: the rare-minority mixture at
  Pythia-70M.}
  \label{fig:app-frontier-mixture-70m-auc}
\end{figure}

\begin{figure}[p]
  \centering
  \includegraphics[width=\linewidth]{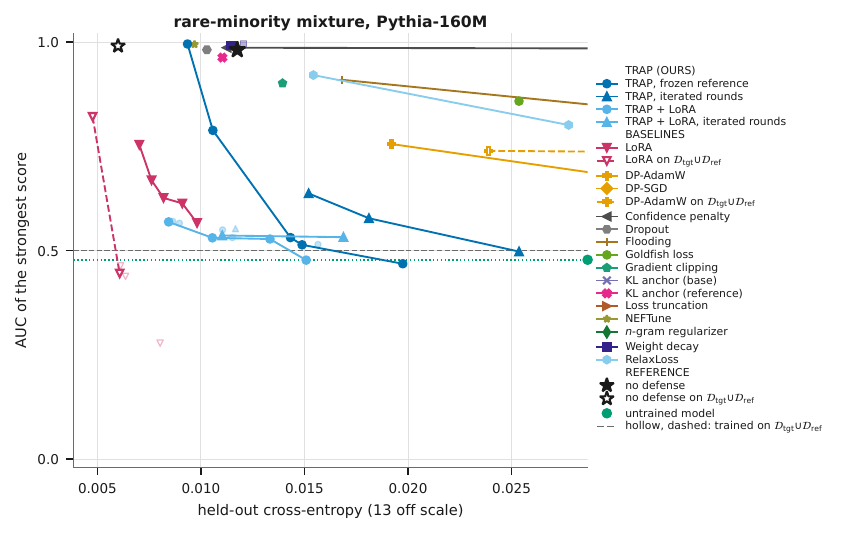}
  \caption{Directional AUC version of
  Figure~\ref{fig:app-frontier-mixture-160m}: the rare-minority mixture at
  Pythia-160M.}
  \label{fig:app-frontier-mixture-160m-auc}
\end{figure}

The normalized AUC of \Eqref{eq:normalized-advantage} is the threshold-free
counterpart of the membership advantage of \citet{yeom2018privacy}:
$2\,\mathrm{AUC}-1$ is the Gini coefficient of the member and non-member
score distributions, and $\nauc$ is its absolute value.
Figures~\ref{fig:app-frontier-essays-70m-auc}
to~\ref{fig:app-frontier-mixture-160m-auc} repeat the per-method frontiers with
the raw, directional AUC of the strongest score on the vertical axis. The
points, the frontier connections and the marker conventions are the same;
only the plotted height changes, so a point can now fall below the chance
line at $0.5$. Such a point is a model whose strongest score ranks held-out
spans above training spans: the defense over-corrected and made its own
training spans less predictable than unseen ones. The strongest \trap{}
settings do this on the \essays{} at 1B, on \medcase{} and on the mixture, and so do the DP models at the floor. $\nauc$ folds
these points back up, which is the conservative reading, since an attacker
who knows the defense flips the sign. It is also a sign of a penalty
stronger than necessary, which a smaller weight $w$ or a larger threshold
$\epsilon$ should remove.

Small values have to be read against two floors. With finitely many spans
$\nauc$ is not zero under the null: for a single score the $95\%$ chance
band is about $0.17$ on the essay spans ($772$ members against $48$
non-members) and about $0.26$ on the $20$-member \medcase{} runs, and
values inside the band read as chance. And because the worst case takes a
maximum over several scores, and member and held-out spans need not have the
same composition, even an untrained model can sit above that band. Every
frontier therefore marks the untrained model, the empirical floor for those
scores, and defended models should be compared with it rather than with
zero.

\FloatBarrier

\section{Reference Schedules}
\label{app:reference-schedules}

\Eqref{eq:loss} defines the penalty but not where the reference comes from.
We use two schedules, and either can train the full model or a LoRA adapter
(\trap{} + LoRA in Section~\ref{sec:mitigation-results}).

\paragraph{Frozen reference.}
Train an ordinary model on $\Dref$ with the testbed's recipe, keep its
best-val checkpoint, freeze it, and train the
target from the pretrained initialization on $\Dtgt$ with \Eqref{eq:loss}.
The reference does not change and never sees the target's examples, which
makes this the cleanest schedule for the bound of
\Eqref{eq:continuation-certificate}, which assumes a fixed reference.

\paragraph{Iterated rounds.}
The first round is frozen-reference training (on the \essays{} a few
round-2 cells were instead seeded by an alternating or mutual first round,
or by the undefended model; the tables mark them). Each later round freezes the
best-val checkpoint of the previous round as the new reference, resets the
target to the pretrained initialization, swaps the roles of the two halves,
and repeats the sweep over the penalty weight; we ran up to five rounds,
and every round enters the sweep as its own setting. The reference thus tracks how far
the target has got with the task. After the first swap, however, the
reference depends on the target's half through its own reference in the
round before, so the gaps of later rounds are no longer taken against a
model that never saw the target's data. This has a failure mode. If a
round's target has memorized a value that also occurs on the other half,
that model becomes the next reference and gives the value a high
probability there too; the next target's advantage on it then falls below
the threshold, and the penalty stops firing on exactly the token it should
catch. Unique secrets, as in our testbeds, do not have this problem;
recurring ones do, and for them the frozen reference, which never sees the
target's half, is the schedule to use. This is also why the bound of
\Eqref{eq:continuation-certificate} is read for the final pair of models
rather than for a schedule.

\subsection{Other references, and why they are not used}
\label{app:reference-ablation}

On the \essays{} we also trained \trap{} against every other reference we
could think of, with the same recipe and the same sweep over the penalty
weight. Lockstep: both models start from the pretrained checkpoint and train
side by side, the reference on $\Dref$ with the ordinary loss and the target
on $\Dtgt$ with \Eqref{eq:loss}, so the reference is at the same stage of
training as the target but never sees its examples. Mutual: both models
train with \Eqref{eq:loss}, each using the other as its current reference.
Alternating: two persistent models swap the target and reference roles every
fixed number of steps within one run; in the asymmetric version the model
meant for release gets a larger penalty weight than the one meant to serve as
reference. Pretrained base: the unmodified base model, which has seen neither
half, is the reference; the larger version uses the pretrained
Pythia-1B-deduped, which has likewise seen neither half, as the reference
for the 70M target.

\begin{figure}[t]
  \centering
  \includegraphics[width=\linewidth]{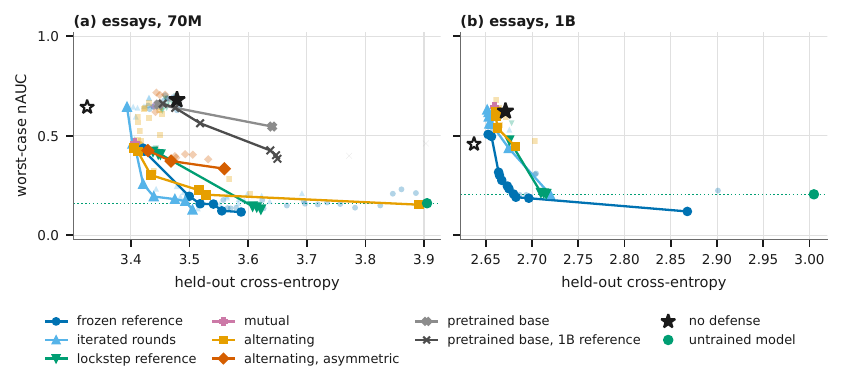}
  \caption{\trap{} with different references on the \essays{} at Pythia-70M
  (a) and 1B (b), best-val checkpoint: worst-case $\nauc$ against held-out
  cross-entropy, one lower-left frontier per reference with its settings
  faint, the undefended model as the star and the untrained model as the
  green dot and dotted line. The asymmetric alternating and the two base references were run at
  70M only.}
  \label{fig:app-reference-ablation}
\end{figure}

Figure~\ref{fig:app-reference-ablation} shows the result. The frozen reference sets the frontier at both scales, with the iterated
rounds close behind. Lockstep reaches the
same floor, but at 70M only after giving up utility. Mutual training gets only part of the way (worst-case $\nauc$ $0.46$ at
70M against $0.71$ undefended, and no lower than the undefended model at
1B): when both models raise the same token
probabilities together, the gap closes without the memorization going away.
Alternating reaches the floor only near the cross-entropy of the untrained
model, and its asymmetric version stops halfway. The pretrained base is the
worst reference: it discounts only what any language model would predict, so
the penalty falls on everything the task teaches, and the target loses
utility while its memorization stays close to the undefended model; a larger
pretrained reference helps a little and does not change the picture. The conclusion is the one
Section~\ref{sec:tra} argues for: the reference has to be trained on the
same task and must not have seen the target's examples, and the frozen
reference is the simplest schedule with both properties.

\FloatBarrier

\section{Properties of \tra{} and \trap{}}
\label{app:tra-properties}

This appendix states and derives the three properties that Section~\ref{sec:method} uses: the expected \tra{} of a training example is the expected generalization gap, the penalty gates each token's gradient, and tokenwise control bounds what the target can assign to a whole continuation relative to the reference.

\subsection{Expected \tra{} equals the expected generalization gap}
\label{app:tra-gap}

For this statement, let $\mathcal J_x$ contain every token the model is trained on, that is, all the tokens the usual fine-tuning loss covers, rather than only the sensitive positions used in evaluation, and define the sample negative log-likelihood as
\begin{equation}
\ell(\theta,x)
:=
-\frac{1}{|\mathcal J_x|}
\sum_{t\in\mathcal J_x}
\log p_\theta(x_t\mid x_{<t}).
\label{eq:sample-loss-definition}
\end{equation}
Suppose the corpus contains $2n$ IID samples from a distribution $\mathcal P$, the balanced partition into $\Dtgt$ and $\Dref$ is drawn independently of their values, and the same randomized learning procedure is applied to each split. After training, draw
$X_{\mathrm{train}}\mid\Dtgt\sim\operatorname{Unif}(\Dtgt)$ independently of the learner randomness. Combining \Eqref{eq:mia-score} with the loss definition gives
\begin{equation}
\mathrm{TRA}(X_{\mathrm{train}})
=
\ell(\theta_{\mathrm{ref}},X_{\mathrm{train}})
-
\ell(\theta,X_{\mathrm{train}}).
\label{eq:sample-gap-loss}
\end{equation}
Although $X_{\mathrm{train}}$ is used to train $\theta$, it is absent from $\Dref$ and therefore independent of $\theta_{\mathrm{ref}}$; marginally, it is distributed as $\mathcal P$. If $X_{\mathrm{test}}\sim\mathcal P$ is independent of the data and both learners, split symmetry also gives $\theta_{\mathrm{ref}}\overset{d}{=}\theta$. Consequently,
\begin{align}
\mathbb E\!\left[\mathrm{TRA}(X_{\mathrm{train}})\right]
&=
\mathbb E_{\theta_{\mathrm{ref}}}
\mathbb E_{X_{\mathrm{test}}\sim\mathcal P}
\ell(\theta_{\mathrm{ref}},X_{\mathrm{test}})
-
\mathbb E_{\Dtgt,\theta}
\frac{1}{n}
\sum_{x\in\Dtgt}
\ell(\theta,x)
\nonumber\\
&=
\mathbb E_{\Dtgt,\theta}
\left[
\mathbb E_{X_{\mathrm{test}}\sim\mathcal P}
\ell(\theta,X_{\mathrm{test}})
-
\frac{1}{n}
\sum_{x\in\Dtgt}
\ell(\theta,x)
\right].
\label{eq:main-population-gap}
\end{align}
The expected full-sequence \tra{} on target training examples is therefore exactly the expected generalization gap, the test loss minus the training loss. The span-restricted \tra{} used in our evaluation is the corresponding localized diagnostic on the sensitive tokens.

The identity needs the two models to be trained the same way. Once the target is trained with the penalty of \Eqref{eq:loss} and the reference is not, $\theta_{\mathrm{ref}}\overset{d}{=}\theta$ no longer holds, so during \trap{} training \tra{} is a training signal rather than an estimate of the gap.

\subsection{The penalty gates each token's gradient}
\label{app:tra-gradient}

Let
\[
g_{x,t}
=
-\nabla_\theta \log p_\theta(x_t\mid x_{<t})
\]
denote the ordinary next-token gradient contributed by token $t$. Away from the threshold $\mathrm{TRA}_t(x)=\epsilon$, the corresponding contribution under \Eqref{eq:loss} is
\begin{equation}
\nabla_\theta\!\left[
-\log p_\theta(x_t\mid x_{<t})
+
w\,\operatorname{ReLU}\!\left(\mathrm{TRA}_t(x)-\epsilon\right)
\right]
=
\left(
1-w\mathbf 1\{\mathrm{TRA}_t(x)>\epsilon\}
\right)g_{x,t},
\label{eq:gradient-gate}
\end{equation}
since the reference term has no dependence on $\theta$. Tokens below the threshold receive their ordinary gradient. Above it, that contribution is attenuated when $0<w<1$, removed when $w=1$, and reversed when $w>1$. This describes the raw contribution of an individual token; minibatch aggregation, gradient clipping, momentum and Adam state alter the resulting parameter update. The point of comparison with the baselines of Section~\ref{sec:mitigation-results} is the criterion: \trap{} changes a token's contribution only when its \tra{} exceeds the threshold, whereas random token dropping or norm-based gradient clipping decide which signal to modify on grounds that have nothing to do with how target-specific the token is.

\subsection{A reference-relative bound on continuations}
\label{app:tra-continuation}

Tokenwise control also bounds complete continuations. For any contiguous span $\mathcal I$ of length $L$,
\begin{equation}
\log
\frac{
\prod_{t\in\mathcal I}
p_\theta(x_t\mid x_{<t})
}{
\prod_{t\in\mathcal I}
p_{\theta_{\mathrm{ref}}}(x_t\mid x_{<t})
}
\leq
L\epsilon
+
\sum_{t\in\mathcal I}
\operatorname{ReLU}\!\left(\mathrm{TRA}_t(x)-\epsilon\right),
\label{eq:continuation-certificate}
\end{equation}
which follows by writing the left side as $\sum_{t\in\mathcal I}\mathrm{TRA}_t(x)$ and bounding each term by $\epsilon+\operatorname{ReLU}(\mathrm{TRA}_t(x)-\epsilon)$. If $\mathrm{TRA}_t(x)\leq\epsilon$ at every position of the span, the target assigns that continuation at most $e^{L\epsilon}$ times the probability the reference does, and because the hinge is taken token by token, one token far above the threshold cannot be offset by tokens below it. The bound is only as good as the reference: if the reference itself gives a span more probability than a model that never saw it would, that excess adds to the exponent and \trap{} does not control it. It is a reference-relative predictive bound for the two models as trained, on the continuations one checks, not a differential-privacy guarantee.

\end{document}